\documentclass{article}

     \PassOptionsToPackage{numbers, compress}{natbib}

 \usepackage[final]{neurips_2024}

\usepackage[utf8]{inputenc} 
\usepackage[T1]{fontenc}    
\usepackage{hyperref}       
\usepackage{url}            
\usepackage{booktabs}       
\usepackage{amsfonts}       
\usepackage{nicefrac}       
\usepackage{microtype}      
\usepackage{xcolor}         

\usepackage{booktabs}       
\usepackage{xcolor} 
\usepackage{graphicx}
\usepackage{amsthm,amsmath,amssymb}
\usepackage{mathrsfs}
\newtheorem{theorem}{Theorem}

\newtheorem{corollary}{Corollary}
\newtheorem{prop}{Proposition}
\usepackage{stfloats} 
\usepackage{miniplot}
\usepackage{bbm}
\usepackage{algorithm}
\usepackage{algorithmic}
\usepackage{multicol}
\usepackage{wrapfig}
\usepackage{subcaption}

\usepackage{braket}
\usepackage{setspace}
\usepackage{lipsum}
\usepackage{multirow}
\usepackage{diagbox}
\usepackage[bottom]{footmisc}
\usepackage{color}
\newcommand{\tabincell}[2]{\begin{tabular}{@{}#1@{}}#2\end{tabular}}
\usepackage{minitoc} 

\usepackage[toc,page,header]{appendix}
\usepackage{minitoc}

\title{Distributional Reinforcement Learning with Regularized Wasserstein Loss}

\author{%
	Ke Sun$^1$, Yingnan Zhao$^{2}$, Wulong  Liu$^3$, Bei Jiang$^{1}$, Linglong Kong$^1$\thanks{Corresponding author}\\
	$^1$University of Alberta, Edmonton, Canada\\
	$^2$ Harbin Engineering University, China\\
	$^3$Huawei Noah’s Ark Lab\\
	\texttt{\{ksun6,bei1,lkong\}@ualberta.ca}\\
	\texttt{zhaoyingnan@hrbeu.edu.cn} \\
	\texttt{liuwulong@huawei.com} \\
}

\begin{document}

	\maketitle

	\begin{abstract}
		The empirical success of distributional reinforcement learning~(RL) highly relies on the choice of distribution divergence equipped with an appropriate distribution representation. In this paper, we propose \textit{Sinkhorn distributional RL~(SinkhornDRL)}, which leverages Sinkhorn divergence—a regularized Wasserstein loss—to minimize the difference between current and target Bellman return distributions. Theoretically, we prove the contraction properties of SinkhornDRL, aligning with the interpolation nature of Sinkhorn divergence between Wasserstein distance and Maximum Mean Discrepancy~(MMD). The introduced SinkhornDRL enriches the family of distributional RL algorithms, contributing to interpreting the algorithm behaviors compared with existing approaches by our investigation into their relationships. Empirically, we show that SinkhornDRL consistently outperforms or matches existing algorithms on the Atari games suite and particularly stands out in the multi-dimensional reward setting. \thanks{Code is available in \url{https://github.com/datake/SinkhornDistRL}.}.
	\end{abstract}
	
	\renewcommand \thepart{}
	\renewcommand \partname{}

	\doparttoc 
	\faketableofcontents 
	
	\section{Introduction}\label{sec:intro}
	
	The design of classical reinforcement learning~(RL) algorithms primarily focuses on the expectation of cumulative rewards that an agent observes while interacting with the environment. Recently, a new class of RL algorithms called \textit{distributional RL} estimates the full distribution of total returns and has exhibited state-of-the-art performance in a wide range of environments, such as C51~\cite{bellemare2017distributional}, Quantile-Regression DQN~(QR-DQN)~\cite{dabney2017distributional}, EDRL~\cite{rowland2019statistics}, Implicit Quantile Networks~(IQN)~\cite{dabney2018implicit}, Fully Parameterized Quantile Function~(FQF)~\cite{yang2019fully}, Non-Crossing QR-DQN~\cite{zhou2020non}, Maximum Mean Discrepancy~(MMD-DQN)~\cite{nguyen2020distributional}, Spline~(SPL-DQN)~\cite{luo2021distributional},  and Sketch-DQN~\cite{wenliang2023distributional}. Beyond the performance advantage, distributional RL has also possessed benefits in risk-sensitive control~\cite{dabney2018implicit,chen2024provable}, exploration~\cite{mavrin2019distributional,cho2024pitfall,sun2021interpreting}, offline setting~\cite{ma2021conservative,wu2023distributional}, statistical value estimation~\cite{rowland2023statistical}, robustness~\cite{sun2021exploring} and optimization~\cite{sun2024does, wang2023benefits,rowland2023analysis,wang2024more}.

	\noindent \textbf{Limitations of Typical Distributional RL Algorithms. } Despite the gradual introduction of numerous algorithms, quantile regression-based algorithms~\cite{dabney2017distributional,dabney2018implicit,yang2019fully,rowland2019statistics,luo2021distributional,rowland2023analysis,rowland2023statistical} dominate attention and research in the realm of distributional RL. These algorithms utilize quantile regression to approximate the one-dimensional Wasserstein distance to compare two return distributions. Nevertheless, two major limitations hinder their performance improvement and wider practical deployment.  \textit{\underline{1) Inaccuracy in Capturing Return Distribution Characteristics.}} The way of directly generating quantiles of return distributions via neural networks often suffers from the non-crossing issue~\cite{zhou2020non}, where the learned quantile curves fail to guarantee a non-decreasing property. This leads to abnormal distribution estimates and reduced model interpretability. The inaccurate distribution estimate is fundamentally attributed to the use of pre-specified statistics~\cite{rowland2019statistics}, while unrestricted statistics based on deterministic samples can be potentially more effective in complex environments~\cite{nguyen2020distributional}. \underline{\textit{2) Difficulties in Extension to Multi-dimensional Rewards.}} Many RL tasks involve multiple sources of rewards~\cite{lizotte2010efficient,dann2023reinforcement}, hybrid reward architecture~\cite{van2017hybrid,lin2020rd}, or sub-reward structures after reward decomposition~\cite{lin2019distributional,zhang2021distributional}, which require learning multi-dimensional return distributions to reduce the intrinsic uncertainty of the environments. However, it remains elusive how to use quantile regressions to approximate a multi-dimensional Wasserstein distance, while circumventing the computational intractability issue in the related multi-dimensional output space.

	\noindent \textbf{Motivation of Sinkhorn Divergence: a Regularized Wasserstein loss.} Sinkhorn divergence~\cite{sinkhorn1967diagonal}  has emerged as a theoretically principled and computationally efficient alternative for approximating Wasserstein distance. It has gained increasing attention in the field of optimal transport~\cite{arjovsky2017wasserstein,genevay2019sample,feydy2019interpolating,peyre2019computational} and has been successfully applied in various areas of machine learning~\cite{patrini2020sinkhorn, genevay2018learning, wong2019wasserstein, fatras2021unbalanced, cao2021don}. By introducing entropic regularization, Sinkhorn divergence can efficiently approximate a multi-dimensional Wasserstein distance using computationally efficient matrix scaling algorithms~\cite{sinkhorn1967diagonal,peyre2019computational}. This makes it feasible to apply optimal transport distances to RL tasks with multi-dimensional rewards~(see experiments in Section~\ref{sec:experiment_multi}). Moreover, Sinkhorn divergence enables the leverage of samples to approximate return distributions instead of relying on pre-specified statistics, e.g., quantiles, thereby increasing the accuracy in capturing the full data complexity behind return distributions and naturally avoiding the non-crossing issues in distributional RL. Beyond addressing the two main limitations mentioned above, the well-controlled regularization introduced in Sinkrhorn divergence helps to find a ``smoother'' transport plan relative to Wasserstein distance, making it less sensitive to noises or small perturbations when comparing two return distributions~(see Appendix~\ref{appendix:transportplan} for the visualization). The term "smoother" refers to the effect of regularization in Sinkhorn divergence to encourage a more uniformly distributed transport plan. This regularization also aligns with the maximum-entropy principle~\cite{jaynes1957information, darroch1972generalized}, which aims to maximize entropy while keeping the transportation cost constrained. Furthermore, the resulting strongly convex loss function~\cite{alaya2019screening} and the induced smoothness by regularization facilitate faster and more stable convergence in the deep RL setting~(see more details in Sections~\ref{sec:method} and~\ref{sec:experiments}).

	\noindent \textbf{Contributions.}  In this work, we propose a new family of distributional RL algorithms based on Sinkhorn divergence, a regularized Wasserstein loss, to address the limitations of quantile regression-based algorithms while promoting more stable training.  As Sinkhorn divergence interpolates between Wasserstein distance and MMD~\cite{gretton2006kernel,feydy2019interpolating,peyre2019computational},  we
	probe this relationship in the RL context, characterizing the convergence properties of dynamic programming under Sinkhorn divergence and revealing the connections of different distances. Our study enriches the class of distributional RL algorithms, making them more effective for a broader range of scenarios and potentially inspiring advancement in other related areas of distribution learning. Our key contributions are summarized as follows:
	
	\textbf{(1) Algorithm.} We introduce a Sinkhorn distributional RL algorithm, called SinkhornDRL, which overcomes the primary shortcomings of predominantly utilized quantile regression-based algorithms.  SinkhornDRL can be seamlessly integrated into existing model architectures and easily implemented.
	
	\textbf{(2) Theory.} We establish the properties of Sinkhorn divergence within distributional RL and derive the relevant convergence results for (multi-dimensional) distributional dynamic programming.
	
	\textbf{(3) Experiments.} We conduct an extensive comparison of SinkhornDRL with typical distributional RL algorithms across 55 Atari games, performing rigorous sensitivity analyses and computation cost assessments. We also verify the efficacy of SinkhornDRL in the multi-dimensional reward setting.

	\section{Preliminary Knowledge}\label{sec:preliminary}
	
	\subsection{Distributional Reinforcement Learning}
	In classical RL, an agent interacts with an environment via a Markov decision process~(MDP), a 5-tuple ($\mathcal{S}, \mathcal{A}, R, P, \gamma$), where $\mathcal{S}$ and $\mathcal{A}$ are the state and action spaces, $P$ is the environment transition dynamics, $R$ is the reward function, and $\gamma \in (0,1)$ is the discount factor. 
	
	Given a policy $\pi$, the discounted sum of future rewards $Z^{\pi}$ is a random variable with $Z^{\pi}(s, a)=\sum_{t=0}^{\infty} \gamma^t R(s_t, a_t)$, where $s_0=s$, $a_0=a$, $s_{t+1}\sim P(\cdot|s_t, a_t)$, and $a_t \sim \pi(\cdot|s_t)$. In expectation-based RL, the action-value function $Q^{\pi}$ is defined as $Q^{\pi}(s, a)=\mathbb{E}\left[Z^{\pi}(s, a)\right]$, which is iteratively updated via Bellman operator $\mathcal{T}^{\pi}$ through $\mathcal{T}^{\pi} Q(s, a)=\mathbb{E}[R(s, a)]+\gamma \mathbb{E}_{s^{\prime} \sim p, \pi}\left[Q\left(s^{\prime}, a^{\prime}\right)\right]$,  where $s^{\prime} \sim P(\cdot | s, a)$ and $a^{\prime} \sim \pi\left(\cdot | s^{\prime}\right)$. In contrast, distributional RL focuses on the action-return distribution, the full distribution of $Z^{\pi}(s, a)$. The return distribution is iteratively updated by applying the distributional Bellman operator $\mathfrak{T}^{\pi}$ through $\mathfrak{T}^{\pi} Z(s, a) :\stackrel{D}{=} R(s, a)+\gamma Z\left(s^{\prime}, a^{\prime}\right)$, where $D$ denotes the distribution and the equality implies random variables of both sides are equal in distribution. The distributional Bellman operator $\mathfrak{T}^{\pi}$ is contractive under certain distribution divergence metrics~\cite{elie2020dynamic}.

	\subsection{Divergences between Measures}

	\noindent \textbf{Optimal Transport~(OT) and Wasserstein / Earth Mover's Distance.} The optimal transport~(OT) metric $W_c$ defines a powerful geometry to compare two probability measures $(\mu, \nu)$, i.e., $W_c = \inf _{\Pi \in \mathbf{\Pi}(\mu, \nu)} \int c(x, y) \mathrm{d} \Pi(x, y)$, where $c$ is the cost function, $\Pi$ is the joint distribution with marginals $(\mu, \nu)$, and the minimizer $\Pi^*$ is called the \textit{optimal transport plan} or \textit{optimal coupling}. The $p$-Wasserstein distance $W_p =(\inf _{\Pi \in \mathbf{\Pi}(\mu, \nu)} \int \Vert x - y \Vert^p \mathrm{d} \Pi(x, y))^{1/p} $ is a special case of optimal transport with the Euclidean norm as the cost function. 	Relative to conventional divergences, including Hellinger, total variation or Kullback-Leibler divergences, the formulation of OT and Wasserstein distance inherently integrates the spatial or geometric relationships between data points and allows them to recover the full support of measures. This theoretical advantage comes, however, with a heavy computational price tag, especially in the high-dimensional space. Specifically, finding the optimal transport plan amounts to solving a linear program and the cost scales at least in $\mathcal{O}(d^3 \log(d))$ when comparing two histograms of dimension $d$~\cite{cuturi2013sinkhorn}.

	\noindent \textbf{Maximum Mean Discrepancy~\cite{gretton2006kernel}.} Define two random variables $X$ and $Y$. The squared Maximum Mean Discrepancy~(MMD) $\text{MMD}^2_k$ with the kernel $k$ is formulated as $\text{MMD}^2_k =\mathbb{E}\left[k\left(X, X^{\prime}\right)\right] + \mathbb{E}\left[k\left(Y, Y^{\prime}\right)\right]-2 \mathbb{E}\left[k(X, Y)\right]$, where $k(\cdot, \cdot)$ is a continuous kernel and $X^\prime$ (resp. $Y^\prime$) is a random variable independent of $X$ (resp. $Y$). Mathematically, the ``flat'' geometry that MMD induces on the space of probability measures does not faithfully lift the ground distance~\cite{feydy2019interpolating}, potentially inferior to OT when comparing two complicated distributions. However, MMD is cheaper to compute than OT with a smaller \textit{sample complexity}, i.e., the number of samples for measures to approximate the true distance~\cite{genevay2019sample}. We provide more details of various distribution divergences as well as their existing contraction properties in distributional RL in Appendix~\ref{appendix:distance}.
	
	\noindent \textbf{Notations.} We constantly use the \textit{unrectified kernel} $k_\alpha=-\Vert x-y \Vert^\alpha$ in our algorithm analysis. With a slight abuse of notation, we also use $Z_\theta$ to denote $\theta$ parameterized return distribution.

	\section{Related Work}\label{sec:relatedwork}

	Based on the choice of distribution divergences and the distribution representation, distributional RL algorithms can be classified into three categories. 
	
	\noindent \textbf{{(1) Categorical Distributional RL.}} As the first successful class, categorical distributional RL~\cite{bellemare2017distributional}, e.g., C51, represents the return distribution using a categorical distribution with discrete fixed supports within a predefined interval. 
	
	\noindent \textbf{{(2) Quantile Regression~(Wasserstein Distance) Distributional RL.}} QR-DQN~\cite{dabney2017distributional} employs quantile regression to approximate the one-dimensional Wasserstein distance. It learns the quantile values for a series of fixed quantiles, offering greater flexibility in the support compared with categorical distributional RL. IQN~\cite{dabney2018implicit} enhances this approach by utilizing an implicit model to produce more expressive quantile values, instead of fixed ones in QR-DQN, while FQF~\cite{yang2019fully} further advances IQN by introducing a more expressive quantile network.  However, as mentioned in Section~\ref{sec:intro}, quantile regression distributional RL struggles with accurately capturing return distribution characteristics and handling multi-dimensional reward settings. SinkhornDRL, with the assistance of an entropy regularization, offers an alternative approach that addresses the two limitations simultaneously. 
	
	\noindent  \textbf{{(3) MMD Distributional RL.}} Rooted in kernel methods~\cite{gretton2006kernel, wenliang2023distributional}, MMD-DQN~\cite{nguyen2020distributional} learns unrestricted statistics, i.e., samples,  to represent the return distribution and optimizes under MMD, which can manage multi-dimensional rewards. However, the data geometry captured by MMD with a specific kernel may be limited, as it is highly sensitive to the characteristics of kernels and the induced Reproducing Kernel Hilbert space~(RKHS)~\cite{genevay2018learning,gretton2006kernel,fukumizu2009kernel}. In contrast, SinkhornDRL is fundamentally based on OT, inherently capturing the spatial and geometric layout of return distributions. This enables SinkhornDRL to potentially surpass MMD-DQN by leveraging a richer representation of data geometry. In Section~\ref{sec:experiments}, we present extensive experiments to demonstrate the advantage of SinkhornDRL over MMD-DQN, particularly in the multi-dimensional reward scenario in Section~\ref{sec:experiment_multi}.

	\section{Sinkhorn Distributional RL~(SinkhornDRL)}\label{sec:method}

	The algorithmic evolution of distributional RL can be primarily viewed along two dimensions~\cite{nguyen2020distributional}. (1) Introducing new distributional RL families beyond the three established ones, leveraging alternative distribution divergences combined with suitable density estimation techniques. (2) Enhancing existing algorithms within a particular family by increasing their model capacity, e.g., IQN and FQF. Concretely, SinkhornDRL falls into the first dimension, aiming to expand the range of distributional RL algorithm families.
	
	\subsection{Sinkhorn Divergence and New Convergence Properties in Distributional RL}

	Sinkhorn divergence~\cite{sinkhorn1967diagonal} efficiently approximates the optimal transport problem by introducing an entropic regularization. It aims at finding a sweet trade-off that simultaneously leverages the geometry property of Wasserstein distance (optimal transport distances) and the favorable sample complexity advantage and unbiased gradient estimates of MMD~\cite{genevay2018learning, feydy2019interpolating}. For two probability measures $\mu$ and $\nu$, the entropic regularized Wasserstein distance $\mathcal{W}_{c, \varepsilon}(\mu, \nu)$ is formulated as
	\begin{equation}\begin{aligned}\label{eq:sinkhorn_regularization}
			\mathcal{W}_{c, \varepsilon}(\mu, \nu) = \min _{\Pi \in \mathbf{\Pi}(\mu, \nu)} \int c(x, y) \mathrm{d} \Pi(x, y) + \varepsilon \text{KL}(\Pi|\mu \otimes \nu),
	\end{aligned}\end{equation}
	where the entropic regularization $\text{KL}(\Pi|\mu \otimes \nu) = \int \log \left(\frac{\Pi(x, y)}{\mathrm{d} \mu(x) \mathrm{d} \nu(y)}\right) \mathrm{d} \Pi(x, y)$, also known as \textit{mutual information}, makes the optimization strongly convex and differential~\cite{alaya2019screening, feydy2019interpolating}, allowing for efficient matrix scaling algorithms for approximation, such as Sinkhorn Iterations~\cite{sinkhorn1967diagonal}. 
	In statistical physics, $\mathcal{W}_{c, \varepsilon}(\mu, \nu)$ can be re-factored as a projection problem:
	\begin{equation}\begin{aligned}\label{eq:sinkhorn_gibbs}
			\mathcal{W}_{c, \varepsilon}(\mu, \nu) := \min _{\Pi \in \mathbf{\Pi}(\mu, \nu)} \text{KL}\left(\Pi|\mathcal{K}\right),
	\end{aligned}\end{equation}
	where $\mathcal{K}$ is the Gibbs distribution and its density function satisfies $d\mathcal{K}(x, y)=e^{-c(x, y) / \varepsilon}d\mu(x)d\nu(y)$. This problem is often referred to as the ``static  Schrödinger problem''~\cite{leonard2013survey,ruschendorf1998closedness} as it was initially considered in statistical physics. Formally, the Sinkhorn divergence is defined as
	\begin{equation}\begin{aligned}\label{eq:sinkhorn_loss}
			\overline{\mathcal{W}}_{c, \varepsilon}(\mu, \nu)=2 \mathcal{W}_{c, \varepsilon}(\mu, \nu)-\mathcal{W}_{c, \varepsilon}(\mu, \mu)-\mathcal{W}_{c, \varepsilon}(\nu, \nu),
	\end{aligned}\end{equation}
	which is smooth, positive definite, and metricizes the convergence in law~\cite{feydy2019interpolating}. This definition subtracts two self-distance terms to ensure non-negativity and metric properties. 
	
	\noindent \textbf{Properties for Convergence.} The contraction analysis of distributional Bellman operator  $\mathfrak{T}^{\pi}$ under a distribution divergence $d_p$ depends on its \textit{scale sensitive}~\textbf{(S)} and \textit{sum invariant}~\textbf{(I)} properties~\cite{bellemare2017cramer, bellemare2017distributional}. We say $d_p$ is scale sensitive (of order $\tau$) if there exists a $\tau > 0$, such that for all random variables $X, Y$ and a real value $a>0$, $d_p(a X, a Y) \leq |a|^\tau d_p(X, Y)$. $d_p$ has the sum invariant property if whenever a random variable $A$ is independent from $X, Y$, we have $d_p(A+X, A+Y)\leq d_p(X, Y)$. Based on these properties, \cite{bellemare2017distributional} shows that $\mathfrak{T}^{\pi}$ is $\gamma$-contractive under the supremal form of Wasserstein distance $W_p$, which is regarding the first term of $\mathcal{W}_{c, \varepsilon}$ or directly letting $\varepsilon=0$  in Eq.~\ref{eq:sinkhorn_regularization}. When examining the regularized loss form of $\mathcal{W}_{c, \varepsilon}$, a natural question arises: \textit{What is the influence of the incorporated regularization term on the contraction of $\mathfrak{T}^{\pi}$?} We begin to address this question in Proposition~\ref{prop:regularization}, focusing on the separate regularization term. Here, we define mutual information as $\text{MI}_\Pi(\mu(s, a), \nu(s, a)) = \text{KL}(\Pi | \mu(s, a) \otimes  \nu(s, a))$ and its supremal form $\text{MI}_\Pi^\infty(\mu, \nu) = \sup_{(s, a)\in \mathcal{S}\times \mathcal{A}} \text{KL}(\Pi | \mu(s, a) \otimes  \nu(s, a))$ given a joint distribution $\Pi$. 
	
	\begin{prop}\label{prop:regularization}  $\mathfrak{T}^{\pi}$ is non-expansive under $\text{MI}_\Pi^\infty$ for any non-trivial joint distribution $\Pi$.
	\end{prop}
	Please refer to Appendix~\ref{appendix:prop_regularization} for the proof, where we investigate both \textbf{(S)} and \textbf{(I)} properties. The non-trivial $\Pi$ rules out the independence case of $\mu$ and $\nu$, where $\text{KL}(\Pi|\mu \otimes \nu)$ would degenerate to zero. Although the non-expansive nature of the introduced regularization term, as shown in Proposition~\ref{prop:regularization}, may potentially slow the convergence in Sinkhorn divergence compared with $W_p$ without the regularization, we will demonstrate that  $\mathfrak{T}^{\pi}$ is still contractive under the full Sinkhorn divergence in Theorem~\ref{theorem:sinkhorn}.  Before introducing Theorem~\ref{theorem:sinkhorn}, we first present the sum-invariant and a new variant of scale-sensitive properties in Proposition~\ref{prop:scalesum}, which acts as the foundation for Theorem~\ref{theorem:sinkhorn}.
	
	\begin{prop}\label{prop:scalesum}  Considering ${\mathcal{W}}_{c, \varepsilon}$ with the unrectified kernel $k_\alpha:=-\|x-y\|^{\alpha}$ as $-c$ ($\alpha>0$) and a scaling factor $a \in (0, 1)$, ${\mathcal{W}}_{c, \varepsilon}$ is sum-invariant \textbf{(I)} and satisfies ${\mathcal{W}}_{c, \varepsilon}(a\mu, a\nu) 
		\leq {\Delta}_\varepsilon(a, \alpha) 	{\mathcal{W}}_{c, \varepsilon}(\mu, \nu)$ \textbf{(S)}  with a scaling constant ${\Delta}_\varepsilon(a, \alpha) \in (|a|^\alpha, 1)$ for any $\mu$ and $\nu$ in a finite set of probability measures. 
		
	\end{prop}
	
	\noindent \textit{Proof Sketch.} The detailed proof is provided in Appendix~\ref{appendix:prop_scalesum}. Let $\Pi^*$ be the optimal coupling of ${\mathcal{W}}_{c, \varepsilon}$, we define a ratio $\lambda_\varepsilon(\mu, \nu)$ that satisfies $\lambda_\varepsilon(\mu, \nu) = \frac{\varepsilon \text{KL}(\Pi^*|\mu \otimes \nu)}{\mathcal{W}_{c, \varepsilon}}\in (0, 1)$ for a generally non-zero ${\mathcal{W}}_{c, \varepsilon}$. The ratio $\lambda_\varepsilon(\mu, \nu)$ measures the proportion of the entropic regularization term over the whole loss term $\mathcal{W}_{c, \varepsilon}$. Therefore, the contraction factor $\Delta_\varepsilon(a, \alpha)$ is defined as $\Delta_\varepsilon(a, \alpha) =|a|^\alpha (1 - \sup_{\mu, \nu} \lambda_\varepsilon(\mu, \nu))  +  \sup_{U, V} \lambda_\varepsilon(\mu, \nu)) \in (|a|^\alpha, 1)$ with $\sup_{\mu, \nu} \lambda_\varepsilon(\mu, \nu) < 1$, which is determined by the scale factor $a$, the order $\alpha$, the hyperparameter $\varepsilon$, and the set of interested probability measures.

	\noindent \textbf{Contraction Guarantee and Interpolation Relationship.} Proposition~\ref{prop:scalesum} reveals that ${\mathcal{W}}_{c, \varepsilon}$ with an unrectified kernel satisfies  \textbf{(I)} and a variant of  \textbf{(S)} properties.  While the scaling constant ${\Delta}_\varepsilon(a, \alpha) $ in \textbf{(S)} has a complicated form, it remains strictly less than one, even considering a non-expansive nature of the entropic regularization as shown in Proposition~\ref{prop:regularization}. We denote the supremal form of Sinkhorn divergence as $\overline{\mathcal{W}}_{c, \varepsilon}^\infty(\mu, \nu): \overline{\mathcal{W}}_{c, \varepsilon}^\infty(\mu, \nu) = \sup_{(s, a)\in \mathcal{S}\times \mathcal{A}} \overline{\mathcal{W}}_{c, \varepsilon}(\mu(s, a), \nu(s, a)).$  In Theorem~\ref{theorem:sinkhorn}, we will integrate all these properties to demonstrate the contraction property of distributional dynamic programming under $\overline{\mathcal{W}}_{c, \varepsilon}$, specifically highlighting the interpolation property of Sinkhorn divergence between MMD and Wasserstein distance in the context of distributional RL. 
	
	\begin{theorem}\label{theorem:sinkhorn} Considering $\overline{\mathcal{W}}_{c, \varepsilon}(\mu, \nu)$ with an unrectified kernel $k_\alpha:=-\|x-y\|^{\alpha}$ as $-c$ ($\alpha>0$), where $\mu, \nu \in$ the distribution set of $ \{Z^\pi(s, a)\}$ for $s\in \mathcal{S}$, $a \in \mathcal{A}$ in a finite MDP. We define the ratio $\overline{\lambda}_\varepsilon(\mu, \nu) $ as  $\overline{\lambda}_\varepsilon(\mu, \nu) = \frac{ \varepsilon \text{KL}(\Pi^*|\mu \otimes \nu)}{\overline{\mathcal{W}}_{c, \epsilon} (\mu, \nu)} \in (0, 1) $ with $ \sup_{\mu, \nu} \overline{\lambda}_\varepsilon(\mu, \nu)<1$. Then, we have:
		
		(1) \textbf{($\varepsilon \rightarrow 0$)} $\overline{\mathcal{W}}_{c, \varepsilon}(\mu, \nu) \rightarrow 2 W^\alpha_{\alpha}(\mu, \nu)$. When $\varepsilon=0$, $\mathfrak{T}^\pi$ is  $\gamma^\alpha$-contractive under $\overline{\mathcal{W}}_{c, \varepsilon}^\infty$.
		
		(2) \textbf{($\varepsilon \rightarrow +\infty$)} $\overline{\mathcal{W}}_{c, \varepsilon}(\mu, \nu) \rightarrow \text{MMD}_{k_\alpha}^2(\mu, \nu)$. When $\varepsilon=+\infty$, $\mathfrak{T}^\pi$ is $\gamma^{\alpha}$-contractive under $\overline{\mathcal{W}}_{c, \varepsilon}^\infty$.

		(3) \textbf{($\varepsilon \in (0, +\infty)$)}, $\mathfrak{T}^\pi$ is at least \textbf{$\overline{\Delta}_\varepsilon(\gamma, \alpha)$-contractive } under $\overline{\mathcal{W}}_{c, \varepsilon}^\infty$, where $\overline{\Delta}_\varepsilon(\gamma, \alpha)$ is an MDP-dependent constant defined as $\overline{\Delta}_\varepsilon(\gamma, \alpha)= \gamma^\alpha (1 - \sup_{\mu, \nu} \overline{\lambda}_\varepsilon(\mu, \nu)) + \sup_{\mu, \nu} \overline{\lambda}_\varepsilon(\mu, \nu)) \in (\gamma^\alpha, 1)$.
		
	\end{theorem}

	\noindent \textit{Proof Sketch.}  The detailed proof of Theorem~\ref{theorem:sinkhorn} can be found in Appendix~\ref{appendix:sinkhorn}. Theorem~\ref{theorem:sinkhorn} (1) and (2) are follow-up conclusions in terms of the convergence behavior of $\mathfrak{T}^\pi$ based on the interpolation relationship between Sinkhorn divergence with Wasserstein distance and MMD~\cite{genevay2018learning}. We also provide a rigorous analysis within the context of distributional RL for completeness. Our critical theoretical contribution is the part (3) for the general $\varepsilon \in (0, \infty)$, where we show that $\mathfrak{T}^\pi$ is at least a $\overline{\Delta}_\varepsilon(\gamma, \alpha)$-contractive operator. The contraction factor $\overline{\Delta}_\varepsilon(\gamma, \alpha) \in (\gamma^\alpha, 1)$ depends on the return distribution set $\{Z^\pi(s, a)\}$ of the considered MDP, and it is also a function of $\gamma, \varepsilon$ and $\alpha$. Due to the influence of the regularization term in Sinkhorn loss, $\overline{\Delta}_\varepsilon(\gamma, \alpha)$ is larger than $|\gamma|^\alpha$, the contraction factor for Wasserstein distance without the regularization.  Thus, $\overline{\Delta}_\varepsilon(\gamma, \alpha)$ can  be seen as an interpolation between $\gamma^\alpha$ and 1,  with the coefficient $\sup_{\mu, \nu} \overline{\lambda}_\varepsilon(\mu, \nu) \in (0, 1)$ defined in Theorem~\ref{theorem:sinkhorn}. The ratio $\overline{\lambda}_\varepsilon(\mu, \nu)$ measures the proportion of the KL regularization term relative to $\overline{\mathcal{W}}_{c, \varepsilon}$. As $\varepsilon \rightarrow 0$ or $+\infty$, $\sup_{\mu, \nu}\overline{\lambda}_\varepsilon(\mu, \nu) \rightarrow 0$, leading to $\gamma^\alpha$-contraction. This aligns with parts (1) and (2).

	\noindent \textbf{Consistency with Existing Contraction Conclusions.} As Sinkhorn divergence interpolates between Wasserstein distance and MMD, its contraction property for $\varepsilon \in [0, \infty]$ also aligns well with the existing distributional RL algorithms when $c=-k_\alpha$. It is worth noting that using Gaussian kernels in the cost function does not yield concise or consistent contraction results like those in Theorem~\ref{theorem:sinkhorn} (3). This conclusion is consistent with MMD-DQN~\cite{nguyen2020distributional}~($\varepsilon \rightarrow +\infty$), where $\mathfrak{T}^\pi$ is generally not a contraction operator under MMD with Gaussian kernels, as counterexamples exist (Theorem 2) in \cite{nguyen2020distributional}. Guided by our theoretical results, we employ the rectified kernel $k_\alpha$ as the cost function and set $\alpha=2$ in our experiments, ensuring that $\mathfrak{T}^\pi$ retains the contraction property guaranteed by Theorem~\ref{theorem:sinkhorn} (3). In Table~\ref{table:comparison}, we also summarize the main properties of distribution divergences in typical distributional RL algorithms, including the convergence rate of $\mathfrak{T}^\pi$ and sample complexity, i.e., the convergence rate of a given metric between a measure and its empirical counterpart as a function of the number of samples $n$.

		\begin{table*}[t!]
		\centering
		\scalebox{0.75}{
			\begin{tabular}{cccccc}
				\toprule[1pt]
				\textbf{Algorithm} & \textbf{$d_p$ Distribution Divergence} & \textbf{Representation $Z_\theta$} & \textbf{Convergence Rate of $\mathfrak{T}^\pi$} & \textbf{Sample Complexity of $d_p$}\\
				\hline
				C51&Cramér distance&Categorical Distribution&$\sqrt{\gamma}$& \diagbox{}{} \\
				QR-DQN-1&Wasserstein distance&Quantiles&$\gamma$&$\mathcal{O}(n^{-\frac{1}{d}})$\\
				MMD-DQN&MMD&Samples&$\gamma^{\alpha/2}$ ($k_\alpha$) &$\mathcal{O}(n^{-1})$\\
				\hline
				\tabincell{c}{SinkhornDRL \\ (ours)}&\tabincell{c}{Sinkhorn divergence \\ ($c = -k_\alpha$)} &Samples&\tabincell{c}{$\gamma$ ($\varepsilon \rightarrow 0$)\\ $\gamma^{\alpha/2}$ ($\varepsilon \rightarrow \infty$)}&\tabincell{c}{$\mathcal{O}(n^{\frac{e^{\frac{\kappa}{\varepsilon}}}{\varepsilon^{\lfloor d / 2\rfloor \sqrt{n}}}})$ ($\varepsilon \rightarrow 0$)\\$\mathcal{O}(n^{-\frac{1}{2}})$ ($\varepsilon \rightarrow \infty$)} \\
				\hline
				\bottomrule[1pt]
			\end{tabular}
		} 
		\caption{Properties of different distribution divergences in typical distributional RL algorithms. $d$ is the sample dimension and $\kappa=2 \beta d+\|c\|_{\infty}$, where the cost function $c$ is $\beta$-Lipschitz~\cite{genevay2019sample}. Sample complexity is improved to $\mathcal{O}(1 / n)$  using the kernel herding technique~\cite{chen2012super} in MMD.} 
		\label{table:comparison}
	\end{table*}
	
	\subsection{Extension to Multi-dimensional Return Distributions} 
	
	As the ability to extend to the multi-dimensional reward setting is one of the major advantages of SinkhornDRL over quantile regression-based algorithms, we next demonstrate that the joint distributional Bellman operator in the multi-dimensional reward case is contractive under Sinkhorn divergence $\overline{\mathcal{W}}_{c, \varepsilon}^\infty$. First, we define a $d$-dimensional reward function as $\mathbf{R}: \mathcal{S} \times \mathcal{A} \rightarrow P(\mathbb{R}^d)$, where $d$ represents the number of reward sources. Consequently, we have joint return distributions of the $d$-dimensional return vector $\mathbf{Z}^{\pi}(s, a) = \sum_{t=0}^{\infty} \gamma^t \mathbf{R}(s_t, a_t)$, where $\mathbf{Z}^{\pi}(s, a) = (Z_1^\pi(s, a), \cdots, Z_d^\pi(s, a))^\top$. The joint distributional Bellman operator $\mathfrak{T}_d^\pi$ applied on the joint distribution of the random vector $\mathbf{Z}(s, a)$ is defined as  $\mathfrak{T}_d^\pi \mathbf{Z}(s, a): \stackrel{D}{=} \mathbf{R}(s, a)+\gamma \mathbf{Z}\left(s^{\prime}, a^{\prime}\right)$, where $s^\prime \sim P(\cdot | s, a)$, $a^\prime \sim \pi(\cdot | s^\prime)$.
	\begin{corollary}\label{corollary:multi}  
		For two joint distributions $\mathbf{Z}_1$ and $\mathbf{Z}_2$,  $\mathfrak{T}_d^\pi$ is  $\overline{\Delta}_\varepsilon(\gamma, \alpha)$-contractive under $\overline{\mathcal{W}}_{c, \varepsilon}^\infty$, i.e.,
		\begin{equation}
			\begin{aligned}
				\overline{\mathcal{W}}_{c, \varepsilon}^\infty(\mathfrak{T}^\pi \mathbf{Z}_1, \mathfrak{T}^\pi \mathbf{Z}_2)  \leq  \overline{\Delta}_\varepsilon(\gamma, \alpha) \overline{\mathcal{W}}_{c, \varepsilon}^\infty(\mathbf{Z}_1, \mathbf{Z}_2).
			\end{aligned}
		\end{equation}
	\end{corollary}

	\begin{algorithm}[b!]
		\caption{Generic Sinkhorn distributional RL Update}
		\textbf{Require}: Number of generated samples $N$,  the cost function $c$, hyperparameter $\varepsilon$ and the target network $Z_{\theta^*}$.\\
		\textbf{Input}: Sample transition $(s, a, r^\prime, s^\prime)$
		\begin{algorithmic}[1] 
			\STATE \textbf{Policy evaluation}: $a^* \sim \pi(\cdot|s^\prime)$ or \textbf{Control}: $a^{*} \leftarrow \arg \max _{a^{\prime} \in \mathcal{A}} \frac{1}{N} \sum_{i=1}^{N} Z_{\theta}\left(s^{\prime}, a^{\prime}\right)_{i}$
			\STATE $\mathfrak{T} Z_{i} \leftarrow r+\gamma Z_{\theta^*}\left(s^{\prime}, a^{*}\right)_{i}, \forall 1 \leq i \leq N$
		\end{algorithmic}
		\textbf{Output}: $\overline{\mathcal{W}}_{c, \varepsilon}\left(\left\{Z_{\theta}(s, a)_{i}\right\}_{i=1}^{N},\left\{\mathfrak{T}  Z_{j}\right\}_{j=1}^{N}\right)$ 
		\label{alg:sinkhorn}
	\end{algorithm}
	
	Please refer to Appendix~\ref{appendix:corollary_multi} for the proof. The contraction guarantee of Sinkhorn divergence enables us to effectively deploy our SinkhornDRL algorithm in various RL tasks that involve multiple sources of rewards~\cite{lizotte2010efficient,dann2023reinforcement}, hybrid reward architecture~\cite{van2017hybrid,lin2020rd}, or sub-reward structures after reward decomposition~\cite{lin2019distributional,zhang2021distributional}. We compare SinkhornDRL with MMD-DQN in multiple reward sources setting in Section~\ref{sec:experiment_multi}, where SinkhornDRL significantly outperforms MMD-DQN by leveraging its ability to capture richer data geometry, a key advantage of optimal transport distances.
	
	\subsection{SinkhornDRL Algorithm and Approximation}

	\noindent \textbf{Equipping Sinkhorn Divergence and Particle Representation.} The key to applying Sinkhorn divergence in distributional RL is to leverage the Sinkhorn loss $\overline{\mathcal{W}}_{c, \varepsilon}$ to measure the distance between the current action-return distribution $Z_\theta(s, a)$ and the target distribution $\mathfrak{T}^\pi Z_\theta(s, a)$. This yields $\overline{\mathcal{W}}_{c, \varepsilon}(Z_\theta(s, a), \mathfrak{T}^\pi Z_\theta(s,a))$ for each $s, a$ pair. For the representation of $Z_\theta(s, a)$, we employ the unrestricted statistics, i.e., deterministic samples, akin to MMD-DQN, instead of predefined statistic functionals like quantiles in QR-DQN or categorical distributions in C51. More concretely,  we use neural networks to generate samples to approximate the return distributions, expressed as $Z_\theta(s, a):=\{Z_\theta(s, a)_i\}_{i=1}^N$, where $N$ is the number of generated samples. We refer to these samples $\{Z_\theta(s, a)_i\}_{i=1}^N$ as \textit{particles}. We then use the Dirac mixture $\frac{1}{N}\sum_{i=1}^{N} \delta_{Z_\theta(s, a)_i}$ to approximate the true density function of $Z^\pi(s, a)$, thus minimizing the Sinkhorn divergence between the approximate distribution and its distributional Bellman target. A generic Sinkhorn distributional RL algorithm with particle representation is provided in Algorithm~\ref{alg:sinkhorn}.

	\noindent \textbf{Efficient Approximation via Sinkhorn Iterations with Guarantee.} By introducing an entropy regularization, Sinkhorn divergence renders optimal transport computationally feasible, especially in the high-dimensional space, via efficient algorithms, e.g., Sinkhorn Iterations~\cite{sinkhorn1967diagonal,genevay2018learning}. Notably, Sinkhorn iteration with $L$ steps yields a differentiable and solvable efficient loss function as the main burden is the matrix-vector multiplication, which streams well on the GPU by simply adding extra differentiable layers on the typical deep neural network, such as a DQN architecture.   \textit{It has been proven that Sinkhorn iterations asymptotically converge to the true loss in a linear rate}~\cite{genevay2018learning,franklin1989scaling,cuturi2013sinkhorn,altschuler2017near}. We provide a detailed description of Sinkhorn iterations in Algorithm~\ref{alg:sinkhorn_iterations} and a full version in Algorithm~\ref{algorithm:full} of Appendix~\ref{appendix:algorithm}. In practice, selecting proper values of $L$ and $\varepsilon$ is crucial. To this end, we conduct a rigorous sensitivity analysis, detailed in Section~\ref{sec:experiments}.
	
	\noindent  \textbf{Remark: Relationship with IQN and FQF.}  In the realm of distributional RL algorithms, it is important to highlight that QR-DQN and MMD-DQN are direct counterparts to SinkhornDRL within the first dimension of algorithmic evolution. In contrast, IQN and FQF enhance QR-DQN and position them in the second modeling dimension, which are orthogonal to our work. As discussed in \cite{nguyen2020distributional}, the techniques from IQN and FQF can naturally extend both MMD-DQN and SinkhornDRL. For instance, we can implicitly generate $\{Z_\theta(s, a)_i\}_{i=1}^N$ by applying a neural network to $N$ samples of a base sampling distribution, as in IQN. We can also use a proposal network to learn the weights of each generated sample as in FQF. We leave these modeling extensions as 
	future works and our current study focuses on rigorously investigating the simplest modeling choice via Sinkhorn divergence.

	\section{Experiments}\label{sec:experiments}
	
	We substantiate the effectiveness of SinkhornDRL as described in Algorithm~\ref{alg:sinkhorn} on the entire 55 Atari 2600 games. Without increasing the model capacity for a fair comparison, we leverage the same architecture as QR-DQN and MMD-DQN,  and replace the quantiles output in QR-DQN with $N$ particles~(samples). In contrast to MMD-DQN, SinkhornDRL only changes the distribution divergence from MMD to Sinkhorn divergence. As such, the potential performance improvement of our algorithm is directly attributed to the theoretical advantages of Sinkhorn divergence over MMD.

	\noindent \textbf{Baseline Implementation.} We choose DQN~\cite{mnih2015human} and three typical distributional RL algorithms as classic baselines, including C51~\cite{bellemare2017distributional}, QR-DQN~\cite{dabney2017distributional} and MMD-DQN~\cite{nguyen2020distributional}. For a fair comparison, we build SinkhornDRL and all baselines based on a well-accepted PyTorch implementation\footnote{\url{https://github.com/ShangtongZhang/DeepRL}} of distributional RL algorithms. We re-implement MMD-DQN based on its original TensorFlow implementation\footnote{\url{https://github.com/thanhnguyentang/mmdrl}}, and keep the same setting. For example, our MMD-DQN still employs Gaussian kernels $k_h(x, y)=\exp(-(x-y)^2/h)$ with the same kernel mixture trick covering a range of bandwidths $h$ as adopted in MMD-DQN~\cite{nguyen2020distributional}. 
	
	\noindent \textbf{SinkhornDRL Implementation and Hyperparameter Settings.} For a fair comparison with QR-DQN, C51, and MMD-DQN, we use the same hyperparameters: the number of generated samples $N=200$, Adam optimizer with $\text{lr}=0.00005, \epsilon_{\text{Adam}}=0.01/32$. In SinkhornDRL, we choose the number of Sinkhorn iterations $L=10$ and smoothing hyperparameter $\varepsilon=10.0$ in Section~\ref{sec:experiments_performance} after conducting sensitivity analysis in Section~\ref{sec:experiments_sensitivity}. Guided by the contraction guarantee analyzed in Theorem~\ref{theorem:sinkhorn}, we use \textit{the unrectified kernel}, specifically setting $-c=k_\alpha$ and choosing $\alpha=2$. This choice ensures \textit{ our implementation is consistent with the theoretical results regarding the contraction guarantee in Theorem~\ref{theorem:sinkhorn} (3)}. We evaluate all algorithms on 55 Atari games, averaging results over three seeds. The shade in the learning curves of each game represents the standard deviation.
	
	
	
	\subsection{Performance of SinkhornDRL}\label{sec:experiments_performance}
	
	\begin{figure*}[t!]
		\centering
		\begin{subfigure}[t]{0.32\textwidth}
			\centering
			\includegraphics[width=\textwidth,trim=0 10 0 10,clip]{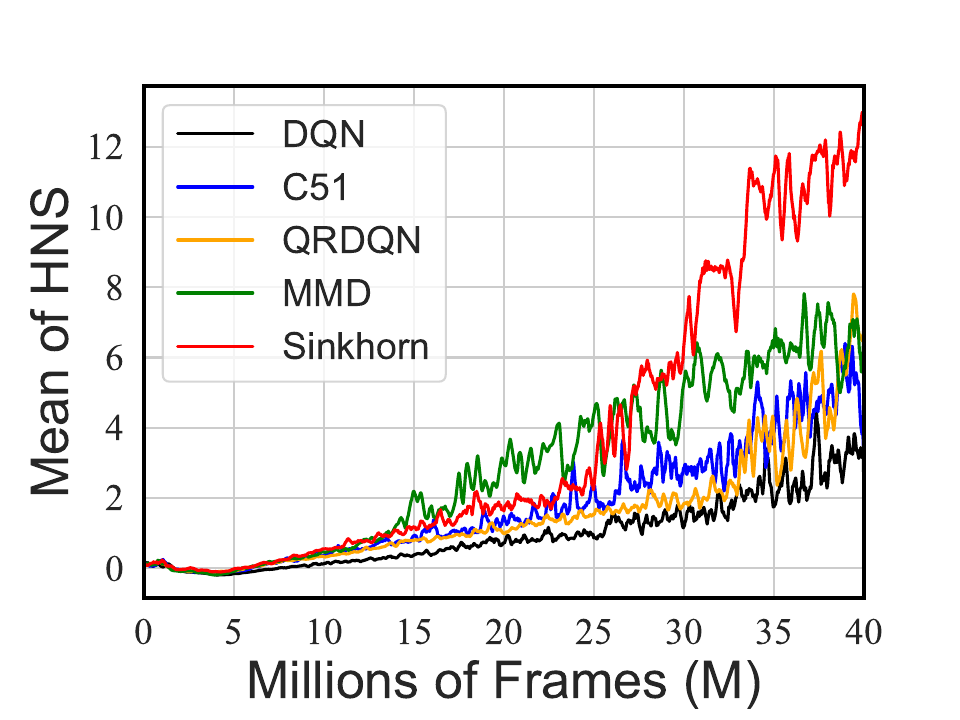}
		\end{subfigure}
		\begin{subfigure}[t]{0.32\textwidth}
			\centering
			\includegraphics[width=\textwidth,trim=0 10 0 10,clip]{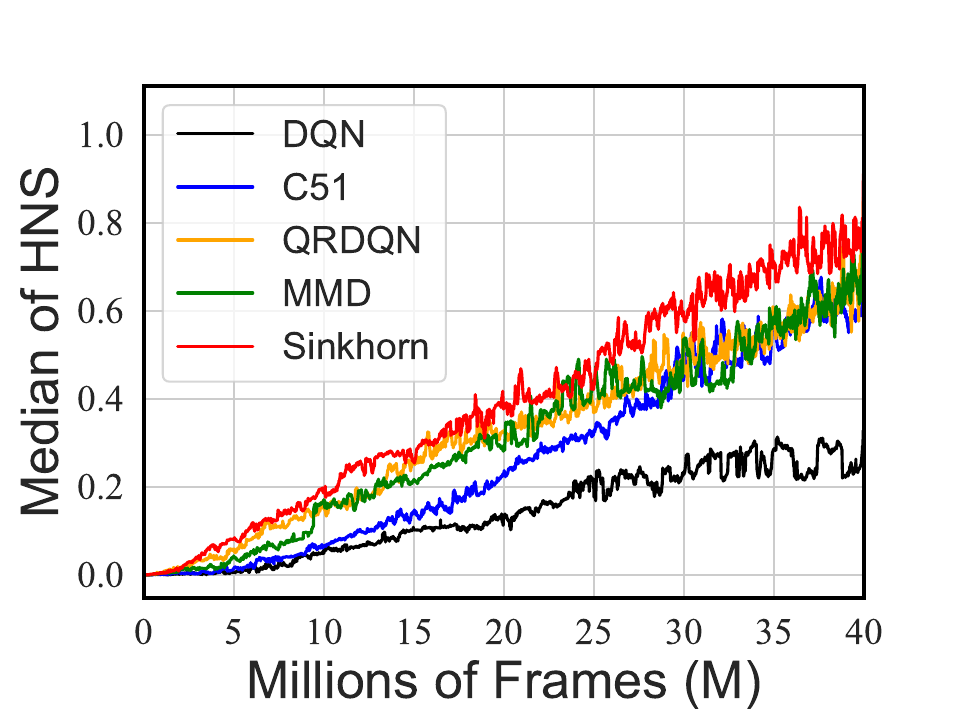}
		\end{subfigure}
		\begin{subfigure}[t]{0.32\textwidth}
			\centering
			\includegraphics[width=\textwidth,trim=0 10 0 10,clip]{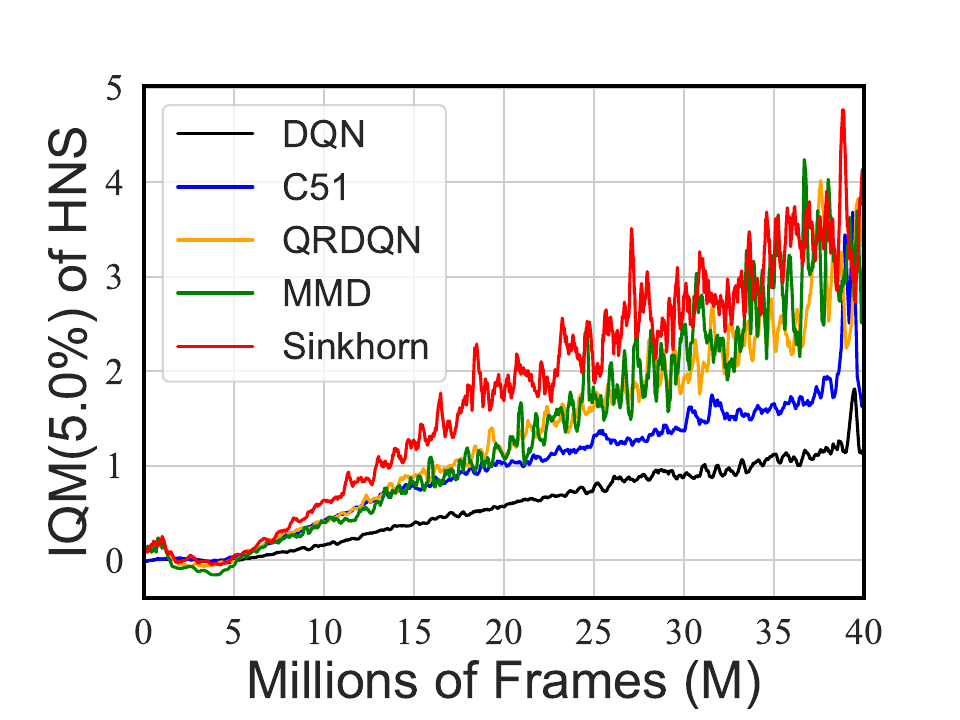}
		\end{subfigure}
		\caption{Mean~(left), Median~(middle), and IQM~(5$\%$)~(right) of Human-Normalized Scores~(HNS) summarized over 55 Atari games. We run 3 seeds for each algorithm.}
		\label{fig:learningcurves_summary}
	\end{figure*}

	\noindent \textbf{Learning Curves of Human Normalized Scores (HNS).} We compare the learning curves of the Mean, Median, and Interquartile Mean~(IQM)~\cite{agarwal2021deep} across all considered distributional RL algorithms in  Figure~\ref{fig:learningcurves_summary} summarized over 55 Atari games. The IQM~(x$\%$) computes the mean from the $x \%$ to $ (1-x) \%$ range of HNS, providing a robust alternative to the Mean that mitigates the impact of extremely high scores on specific games and is more statistically efficient than the Median. For computational feasibility, we evaluate the algorithms over 40M training frames. Our findings reveal that SinkhornDRL achieves state-of-the-art performance in terms of mean, median, and IQM~(5$\%$) of HNS across most training phases. Notably, SinkhornDRL exhibits slower convergence during the early training phase, as indicated by the Mean of HNS~(left panel of Figure~\ref{fig:learningcurves_summary}). This slower initial convergence can be explained by the slower contraction factor $\overline{\Delta}_\varepsilon(\gamma, \alpha) > \gamma^\alpha$ in Theorem~\ref{theorem:sinkhorn}, as opposed to MMD-DQN. 
	To ensure the reliability of our results, we also provide the learning curves for each Atari game in Figure~\ref{fig:allgames} in Appendix~\ref{appendix:experiment_games}. Furthermore, a table summarizing all raw scores is available in Table~\ref{table:allresults} in Appendix~\ref{appendix:allresults}. This table highlights that SinkhornDRL achieves the highest numbers of best and second-best performance of all games among all baseline algorithms. A summary table of  Mean,  IQM, and Median HNS is  also given in Table~\ref{table:allresults_summary}  of  Appendix~\ref{appendix:table_HNS}. Overall, we conclude that SinkhornDRL generally outperforms existing distributional RL algorithms.

	\begin{figure*}[b!]
	\centering
	\begin{subfigure}[t]{0.48\textwidth}
		\centering
		\includegraphics[width=\textwidth,trim=20 0 60 10,clip]{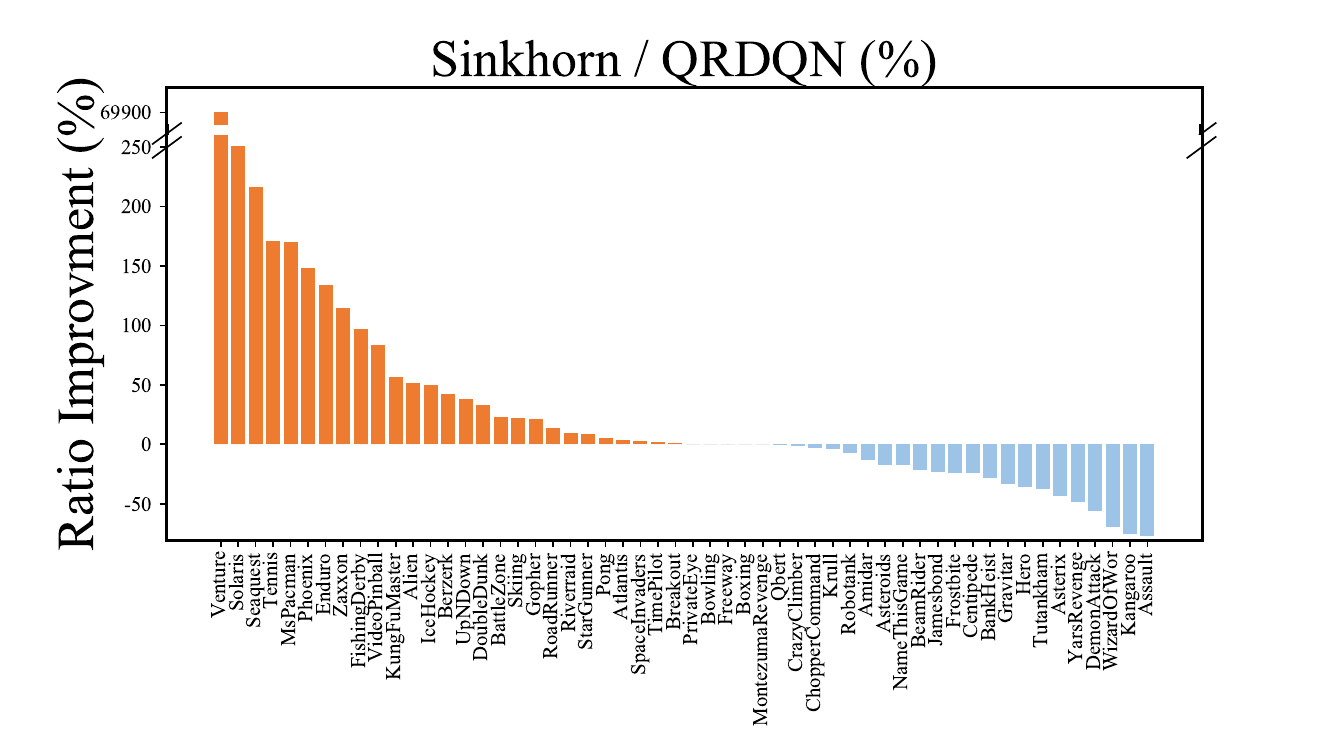}
		\caption{SinkhornDRL vs QR-DQN}
	\end{subfigure}
	\begin{subfigure}[t]{0.48\textwidth}
		\centering
		\includegraphics[width=\textwidth,trim=20 0 60 10,clip]{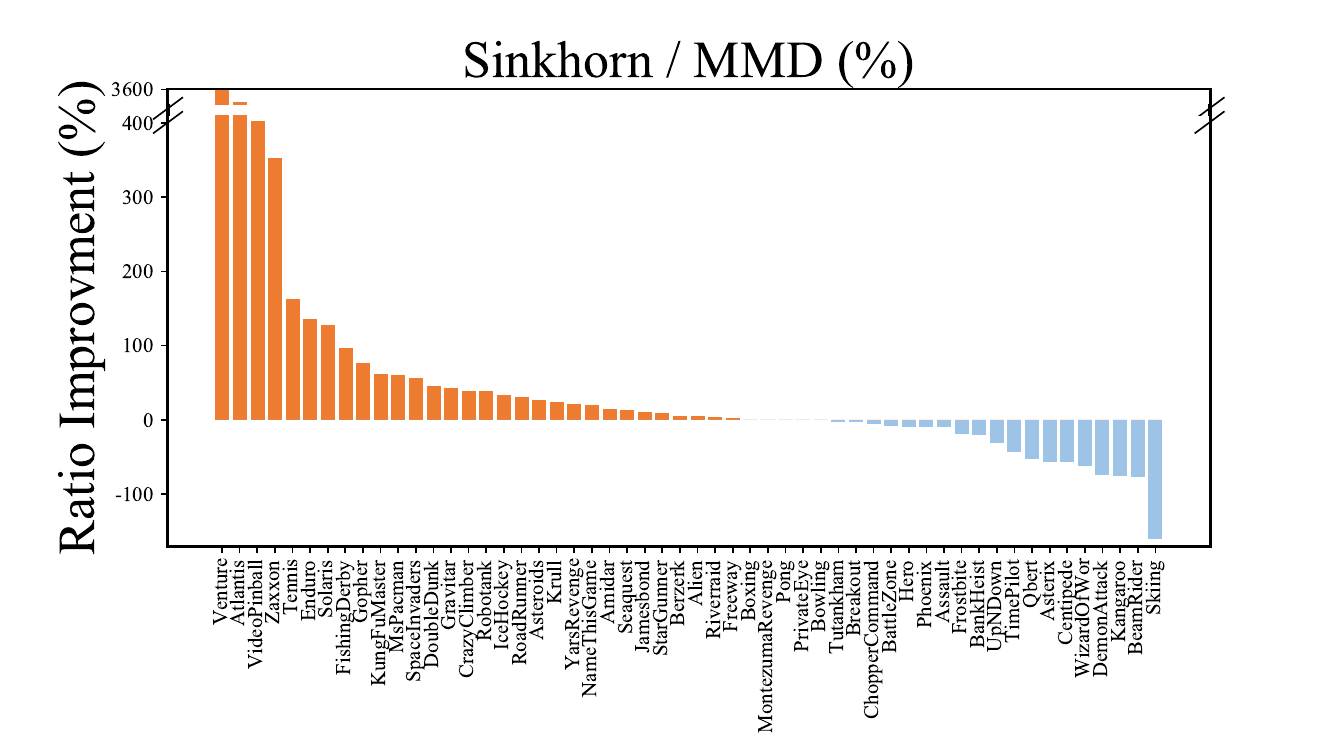}
		\caption{SinkhornDRL vs MMD-DQN}
	\end{subfigure}
	\caption{Ratio improvement of return for SinkhornDRL over QR-DQN~(left) and MMD-DQN~(right) averaged over 3 seeds. The ratio improvement is calculated by (SinkhornDRL - QR-DQN) / QR-DQN in (a) and (SinkhornDRL - MMD-DQN) / MMD-DQN in (b), respectively.}
	\label{fig:ratio_max}
\end{figure*}
	
	\noindent \textbf{Ratio Improvement Analysis across All Games.} Given the interpolation nature of Sinkhorn divergence between Wasserstein distance and MMD, as analyzed in Theorem~\ref{theorem:sinkhorn}, a pertinent question arises: \textit{In which environments does SinkhornDRL potentially perform better?} We empirically address this question by conducting a ratio improvement comparison between SinkhornDRL and both QR-DQN and MMD-DQN across all games. Figure~\ref{fig:ratio_max} showcases that SinkhornDRL surpasses both QR-DQN and MMD-DQN  in more than half of the games and significantly excels at them in a large proportion of games. Notably,  \textit{the games where SinkhornDRL achieves considerable improvement tend to have larger action spaces and more complex dynamics}. In particular, as illustrated in Figure~\ref{fig:ratio_max}, these games include Venture, Seaquest, Solaris, Tennis, Phoenix, Atlantis, and Zaxxon. Most of these games have an 18-dimensional action space and intricate dynamics, except for Atlantis, which has a 4-dimensional action space and simpler dynamics where MMD-DQN is substantially inferior to SinkhornDRL. For a detailed comparison, we provide the features of all games, including the number of action spaces, and complexity of environment dynamics in Table~\ref{table:games} of Appendix~\ref{appendix:featuresgames}.

	In summary, compared with QR-DQN, the empirical success of SinkhornDRL can be attributed to several key factors: 1. \textit{Enhanced return distribution representation:} SinkhornDRL captures return distribution characteristics more accurately by directly using samples, avoiding the non-crossing issue of learned quantile curves or the potential limitations of quantile representation. \textit{2. Smooth transport plan and stable convergence.} The induced smoother transport plan~(see Appendix~\ref{appendix:transportplan} for visualization) and the inherent smoothness of Sinkhhorn divergence contribute to more stable convergence, leading to performance improvement. In contrast to MMD-DQN, the benefits of SinkhornDRL arise from its richer data representation capability when comparing return distributions, rooted in the OT nature. This is in comparison to the potentially restricted kernel-specific distances, such as MMD.

	\subsection{Sensitivity Analysis and Computational Cost}\label{sec:experiments_sensitivity}

\begin{figure*}[htbp]
		\centering
		\begin{subfigure}[t]{0.32\textwidth}
			\centering
			\includegraphics[width=\textwidth,trim=0 0 0 10,clip]{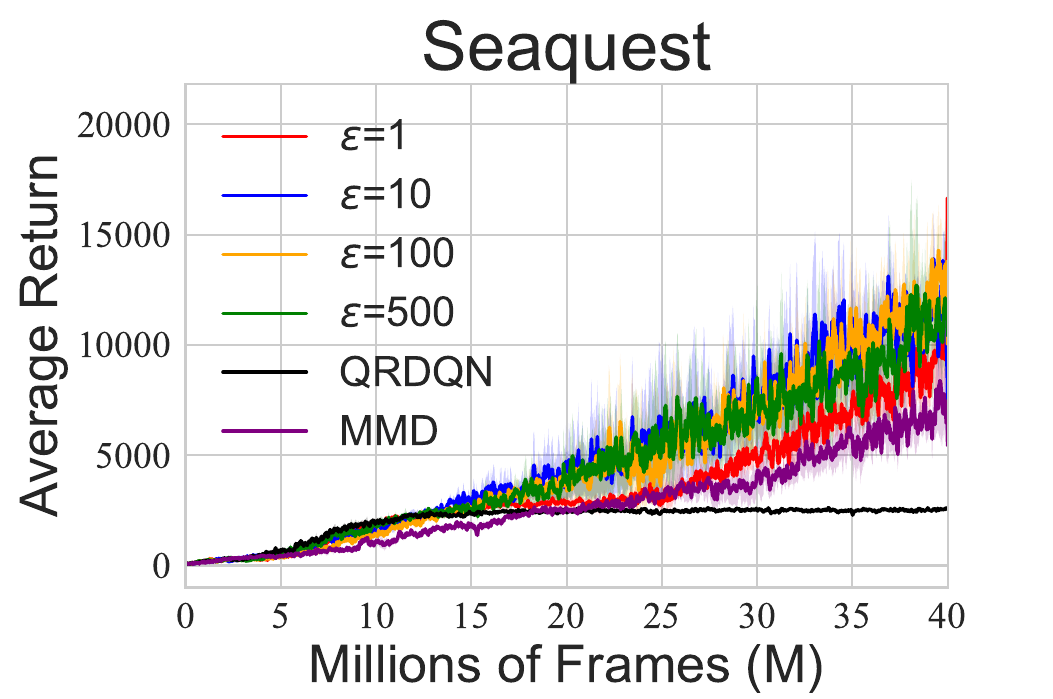}
			\caption{Hyper-parameter $\varepsilon$}
		\end{subfigure}
		\begin{subfigure}[t]{0.32\textwidth}
			\centering
			\includegraphics[width=\textwidth,trim=0 0 0 10,clip]{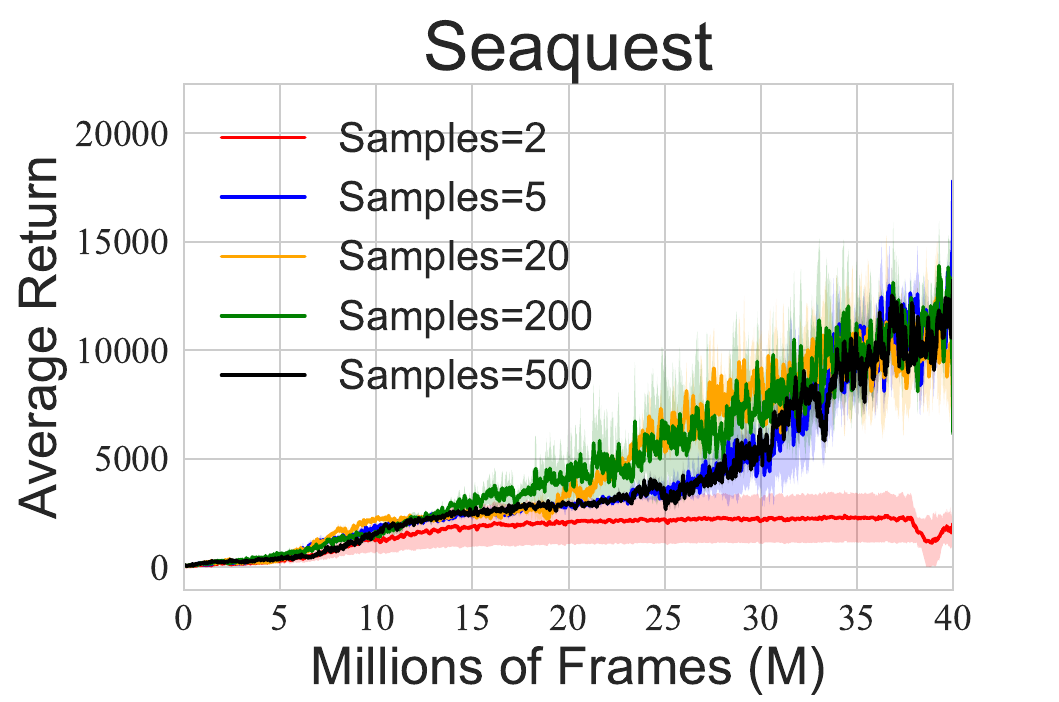}
			\caption{Number of Samples}
		\end{subfigure}
		\begin{subfigure}[t]{0.335\textwidth}
			\centering
			\includegraphics[width=\textwidth,trim=0 0 0 10,clip]{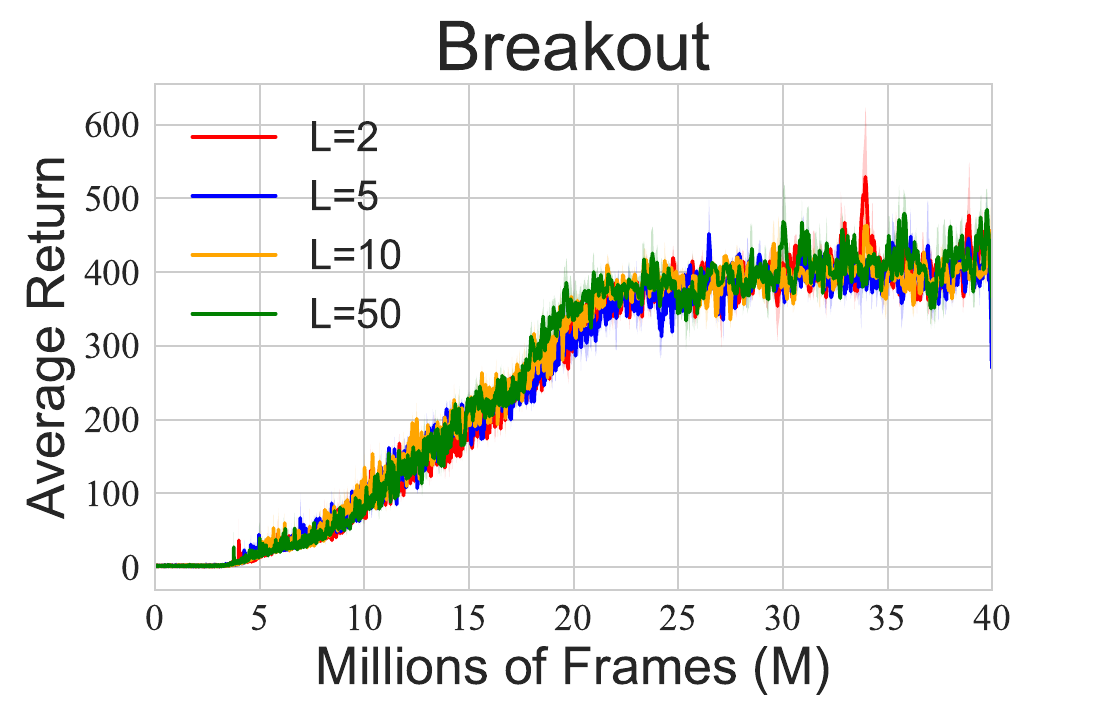}
			\caption{Sinkhorn Iterations $L$}
		\end{subfigure}
		\caption{Sensitivity analysis of SinkhornDRL on Breakout and Seaquest in terms of $\varepsilon$, number of samples, and number of iteration $L$. Learning curves are reported over three seeds.}
		\label{fig:sensivity}
	\end{figure*}

	\noindent \textbf{Sensitivity Analysis.} In practice, a proper $\varepsilon$ is preferable as an overly large or small $\varepsilon$ will lead to numerical instability of Sinkhorn iterations in Algorithm~\ref{alg:sinkhorn_iterations}~(see the discussion in Section 4.4 of \cite{peyre2019computational}), therefore worsening its performance, as shown in Figure~\ref{fig:sensivity}~(a). This implies that the potential interpolation nature of limiting behaviors between SinkhornDRL with QR-DQN and MMD-DQN revealed in Theorem~\ref{theorem:sinkhorn} may not be able to be rigorously verified in numerical experiments. SinkhornDRL also requires a proper number of iterations $L$ and samples $N$. For example, a small $N$, e.g., $N=2$ in Seaquest in Figure~\ref{fig:sensivity}~(b) leads to the divergence of algorithms, while an overly large $N$ can degrade the performance and meanwhile increases the computational burden~(Appendix~\ref{appendix:cost}). We conjecture that using larger networks to represent more samples is more likely to suffer from overfitting, yielding the instability in the RL training~\cite{bjorck2021towards}. Therefore, we choose $N=200$ to attain favorable performance and guarantee computational effectiveness simultaneously. We provide a more detailed sensitivity analysis and more results on StarGunner and Zaxxon in Appendix~\ref{appendix:sensitivity}.
	
	\noindent \textbf{Computation Cost.} In terms of the computation cost, SinkhornDRL slightly increases the computational overhead compared with C51, QR-DQN, and MMD-DQN. For instance, SinkhornDRL increases the average computational cost compared with MMD-DQN by around 20$\%$. Due to the space limit, we provide more computation cost comparison in terms of $L$ and $N$ in Appendix~\ref{appendix:cost}.
	
	\subsection{Modeling Joint Return Distribution for Multi-Dimensional Reward Functions}\label{sec:experiment_multi}
	
	Many RL tasks involve modeling multivariate return distributions. Following the multi-dimensional reward setting in \cite{zhang2021distributional}, we compare SinkhornDRL with MMD-DQN on six Atari games with multiple sources of rewards. In these tasks, the primitive scalar-based rewards are decomposed into reward vectors based on the respective reward structures~(see Appendix~\ref{appendix:multi} for more details). Figure~\ref{fig:multi} showcases that SinkhornDRL outperforms MMD-DQN in most cases for multi-dimensional reward functions. Of particular note, it remains an open question to directly approximate multi-dimensional Wasserstein distances via quantile regression or other efficient algorithms, particularly in RL tasks. 
	
	\begin{figure*}[htbp]
		\centering
		\begin{subfigure}[t]{0.161\textwidth}
			\centering
			\includegraphics[width=\textwidth,trim=0 0 40 0,clip]{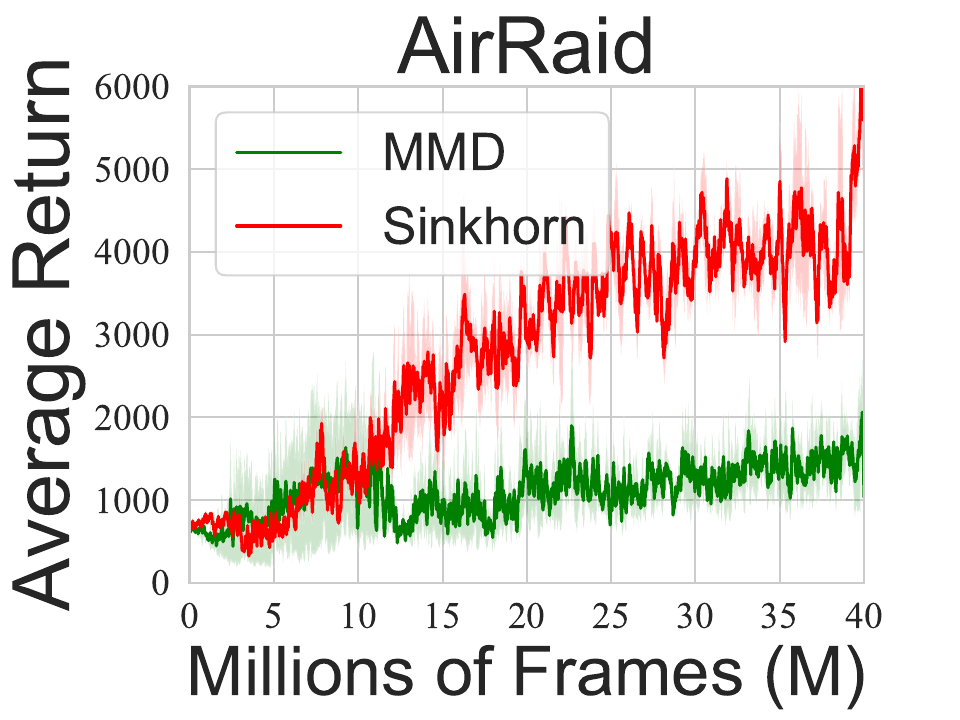}
		\end{subfigure}
		\begin{subfigure}[t]{0.161\textwidth}
			\centering
			\includegraphics[width=\textwidth,trim=0 0 40 0,clip]{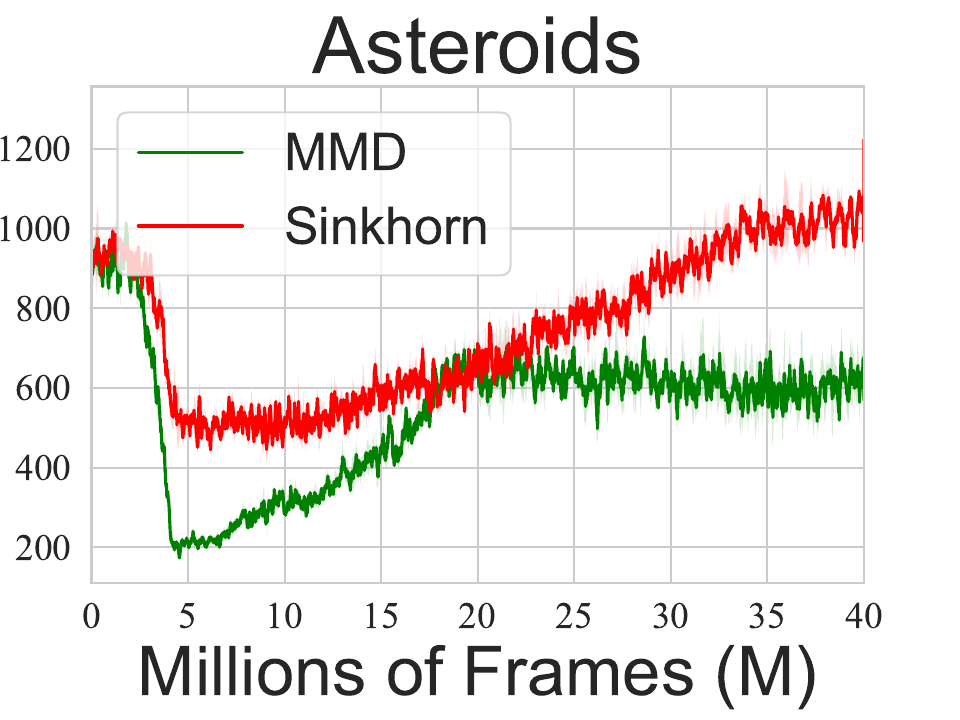}
		\end{subfigure}
		\begin{subfigure}[t]{0.161\textwidth}
			\centering
			\includegraphics[width=\textwidth,trim=0 0 38 0,clip]{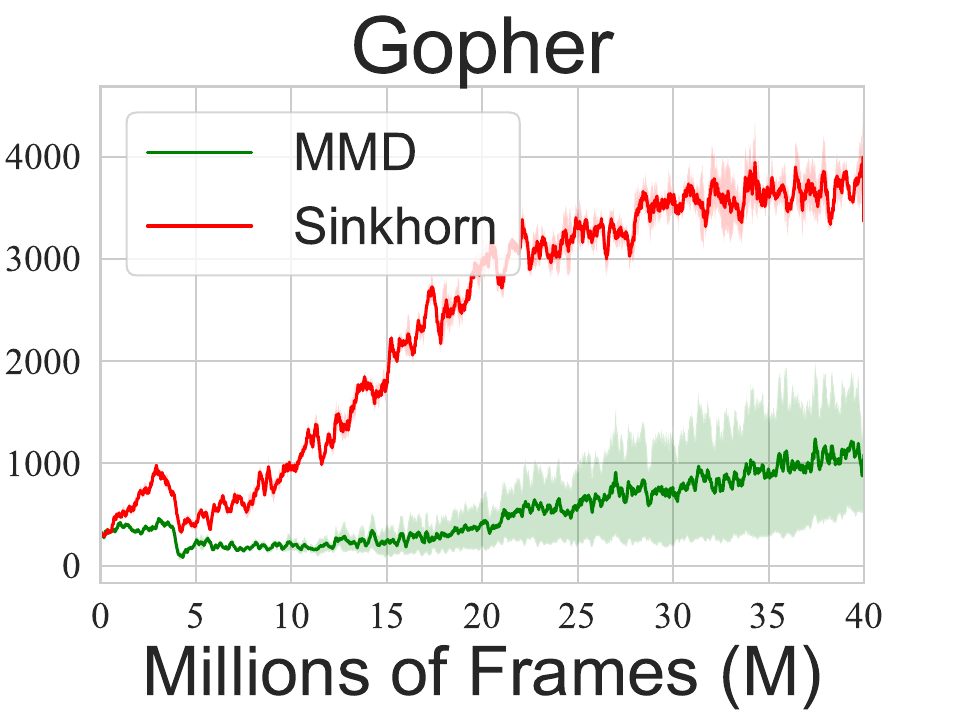}
		\end{subfigure}
		\begin{subfigure}[t]{0.161\textwidth}
			\centering
			\includegraphics[width=\textwidth,trim=0 0 40 0,clip]{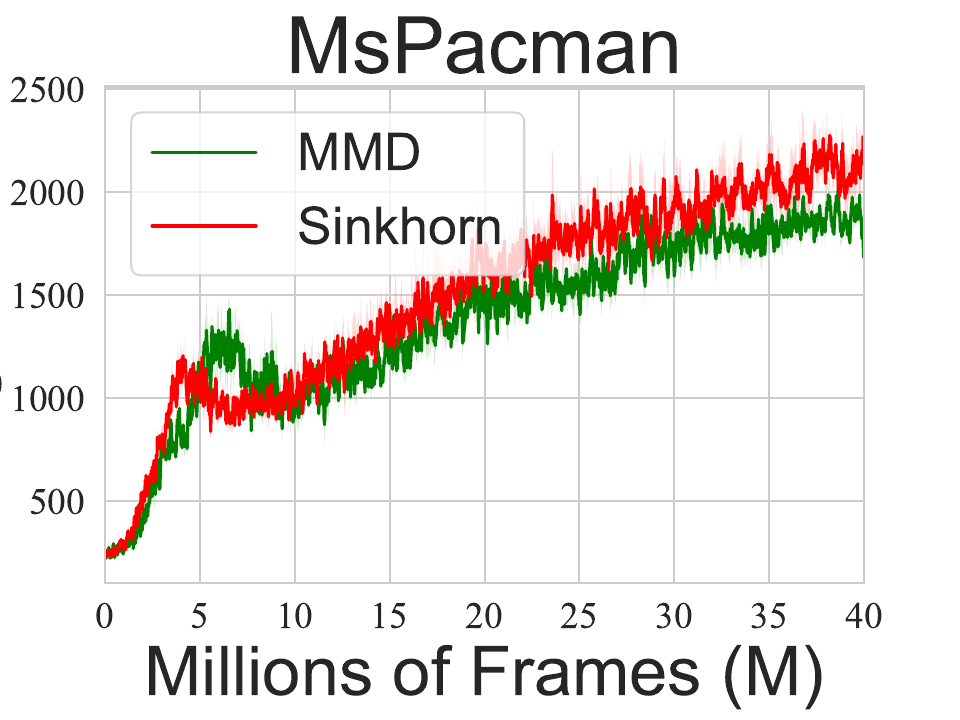}
		\end{subfigure}
		\begin{subfigure}[t]{0.161\textwidth}
			\centering
			\includegraphics[width=\textwidth,trim=0 0 40 0,clip]{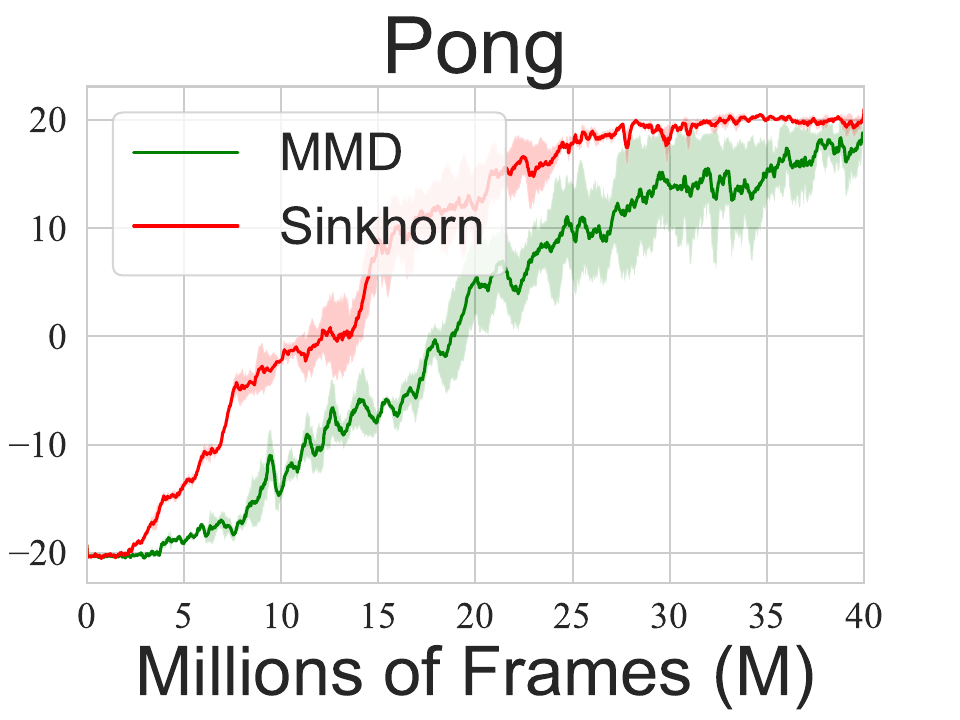}
		\end{subfigure}
		\begin{subfigure}[t]{0.161\textwidth}
			\centering
			\includegraphics[width=\textwidth,trim=0 0 39 0,clip]{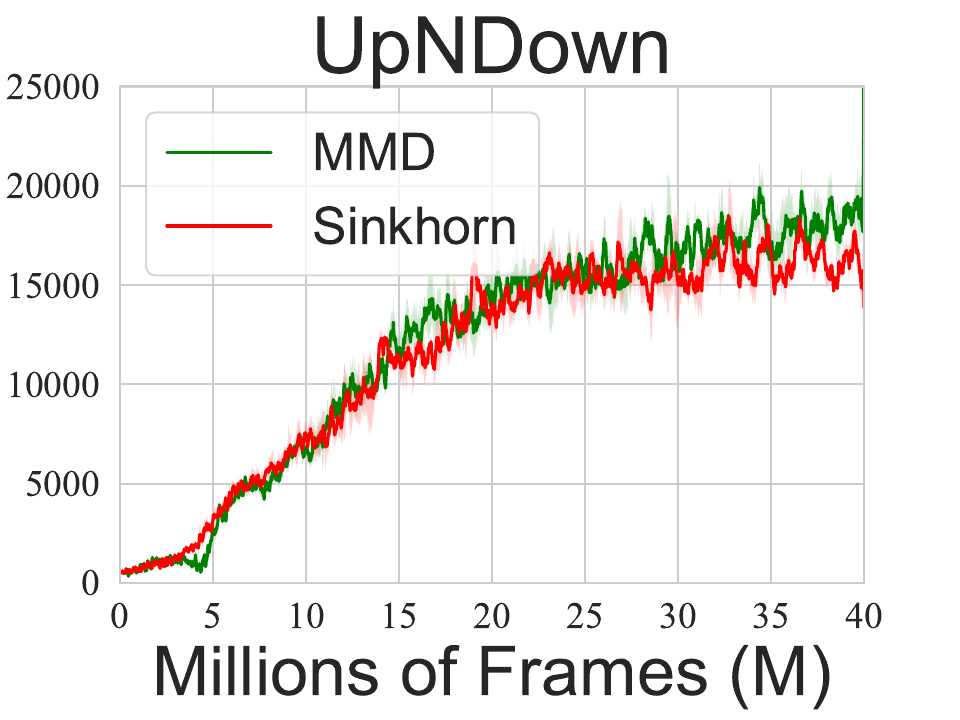}
		\end{subfigure}
		\caption{Performance of SinkhornDRL on six Atari games with multi-dimensional reward functions.}
		\label{fig:multi}
	\end{figure*}
	
	\section{Conclusion, Limitations and Future Work}\label{sec:discussion}
	
	In this work, we propose a novel family of distributional RL algorithms based on Sinkhorn divergence that accomplishes competitive performance compared with the typical distributional RL algorithms on the Atari games suite. Theoretical results about the properties of this regularized Wasserstein loss and its convergence guarantee in the context of RL are provided with rigorous empirical verification. 
	
	\noindent \textbf{Limitations.} While SinkhornDRL achieves competitive performance, it relatively increases the computational cost and requires tuning additional hyperparameters. This hints that the enhanced performance offered by SinkhornDRL may come with slightly greater efforts in practical deployment. Additionally, it remains elusive for a deeper connection between the theoretical properties of divergences and the practical performance of distributional RL algorithms given a specific environment. 
	
	
	\noindent \textbf{Future work.} Along the two dimensions of distributional RL algorithm evolution, we can further improve Sinkhorn distributional RL by incorporating implicit generative models, including parameterizing the cost function and increasing model capacity. Moreover,  Sinkhorn distributional RL also opens a door for new applications of Sinkhorn divergence and more optimal transport approaches in RL. It also becomes increasingly crucial to design a quantitative criterion for a given environment to recommend the choice of a specific distribution divergence before conducting costly experiments. 

	\section*{Acknowledgements}
	Yingnan Zhao and Ke Sun were supported by the State Scholarship Fund from China Scholarship Council (No:202006120405 and No:202006010082). Bei Jiang and Linglong Kong were partially supported by grants from the Canada CIFAR AI Chairs program, the Alberta Machine Intelligence Institute (AMII), and Natural Sciences and Engineering Council of Canada (NSERC), and Linglong Kong was also  partially supported by grants from the Canada Research Chair program from NSERC.  The authors express their gratitude for the insightful feedback provided by all reviewers and the area chairs, which significantly enhanced the initial version of this paper.

\bibliographystyle{plain}
\bibliography{Sinkhorn}

\newpage
\appendix

\onecolumn

\addcontentsline{toc}{section}{Appendix} 
\part{Appendix} 
\parttoc 

\clearpage

\section{Smoother Transport Plan via Sinkhorn Divergence by Increasing $\varepsilon$}\label{appendix:transportplan}

\begin{figure*}[htbp]
	\centering
	\begin{subfigure}[t]{0.7\textwidth}
		\centering
		\includegraphics[width=\textwidth,trim=50 20 50 20,clip]{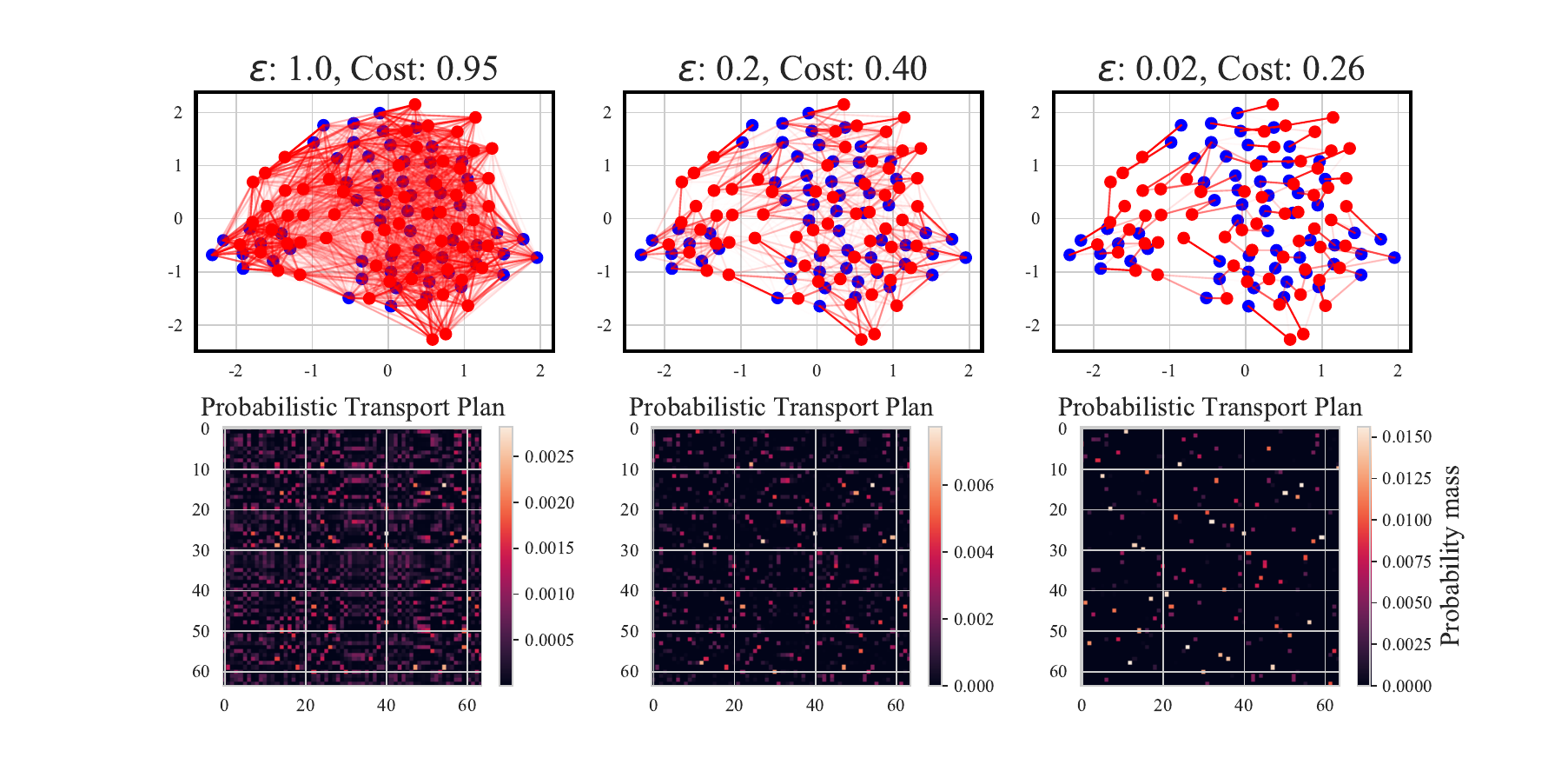}
		\caption{Enduro}
	\end{subfigure}
	\begin{subfigure}[t]{0.7\textwidth}
		\centering
	\includegraphics[width=\textwidth,trim=50 20 50 20,clip]{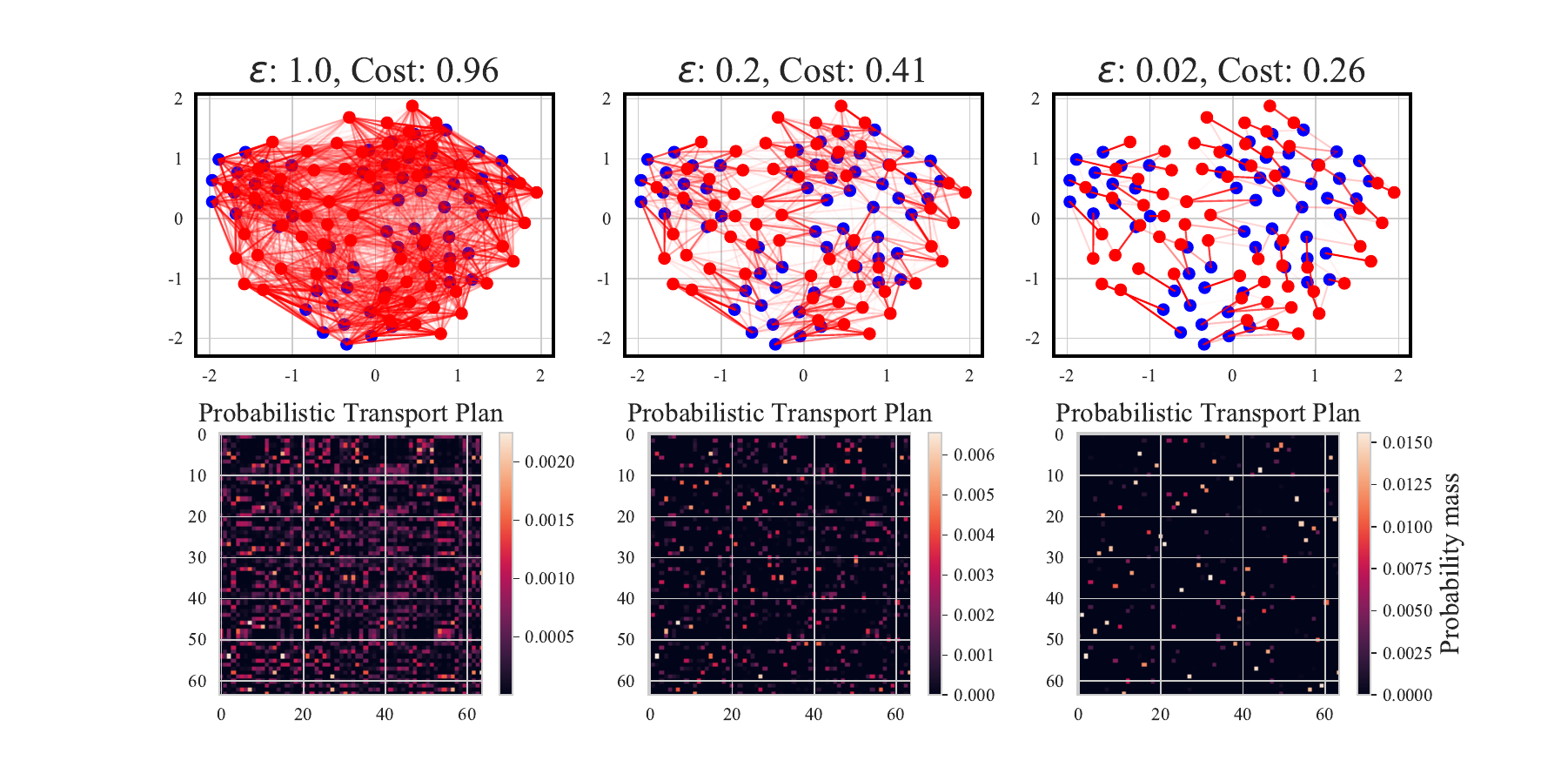}
		\caption{Qbert}
	\end{subfigure}
		\begin{subfigure}[t]{0.7\textwidth}
		\centering
		\includegraphics[width=\textwidth,trim=50 20 50 20,clip]{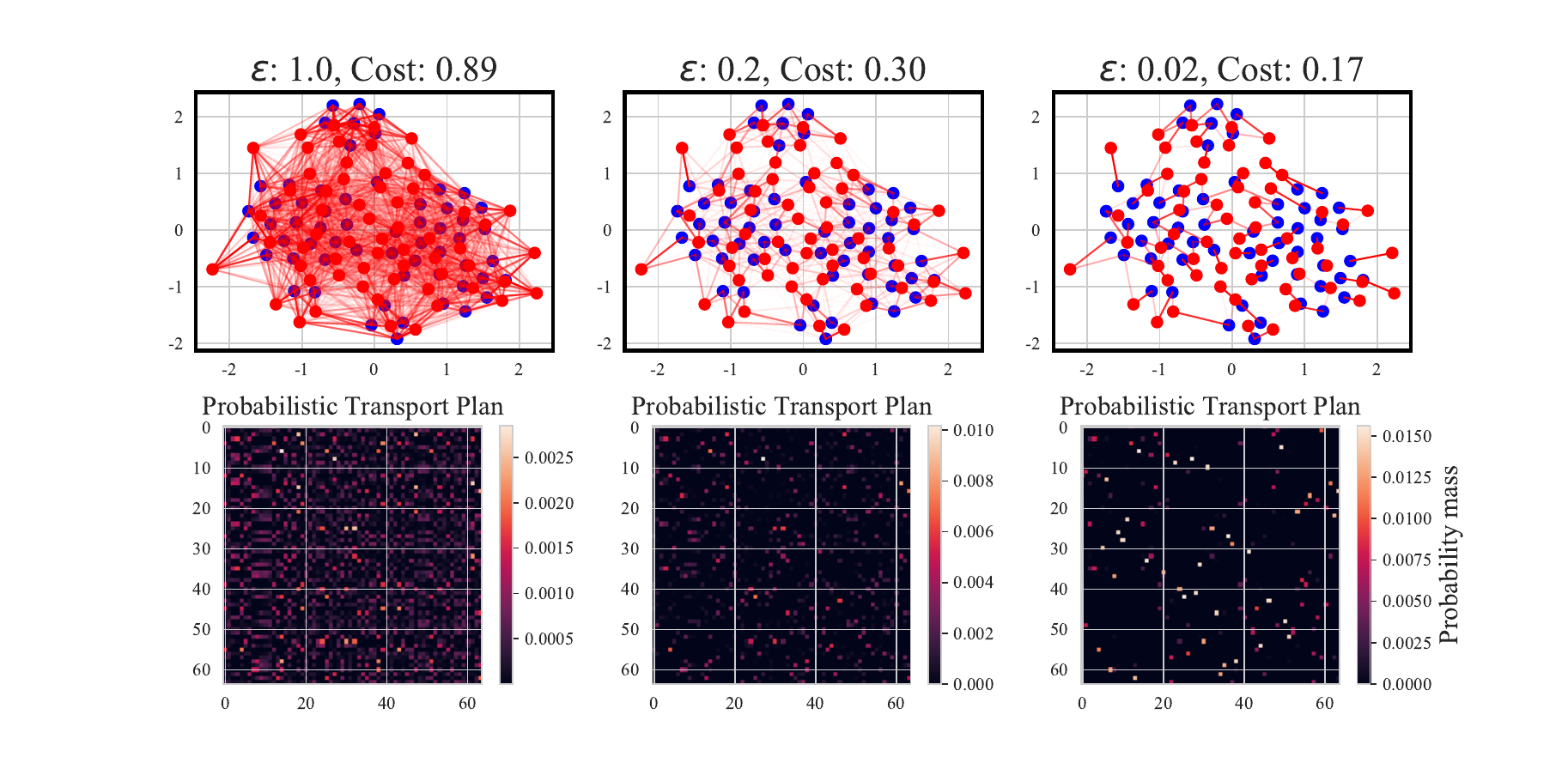}
		\caption{Seaquest}
	\end{subfigure}
	\caption{Optimal transport plans for via Sinkhorn Iterations in SinkhornDRL on three Atari games. The first row denotes the (two-dimensional) spatial transport plans across different data points, while the second row represents the heat map of the obtained transport plan (optimal coupling). }
	\label{fig:transportplan}
\end{figure*}

We visualize the optimal transport plans by solving Sinkhorn divergence with different $\varepsilon$ in well-trained SinkhornDRL models across three games in Figure~\ref{fig:transportplan} We evaluate (randomly selected 64) current and target state features to be compared and then apply t-SNE to reduce their dimension from 512  to 2 associated with a normalization for visualization. In each game of Figure~\ref{fig:transportplan}, as we increase the regularization strength $\varepsilon$ (from right to left), the resulting transport plans tend to be smoother, less concentrated, and more uniformly distributed by transporting the point mass between two distributions (in red and blue).

\section{Definition of Distribution Divergences and Contraction Properties}\label{appendix:distance}

\textbf{Definition of distances.} Given two random variables $X$ and $Y$, one-dimensional $p$-Wasserstein metric $W_p$ between the distributions of $X$ and $Y$ has a simplified form via the quantile functions:	 
\begin{equation}\begin{aligned}\label{eq:wasserstein}
		W_{p}(X, Y)=\left(\int_{0}^{1}\left|F_{X}^{-1}(\omega)-F_{Y}^{-1}(\omega)\right|^{p} d \omega\right)^{1 / p} = \Vert F_{X}^{-1} - F_{Y}^{-1} \Vert_p,
\end{aligned}\end{equation}
which $F^{-1}$ is the quantile function, also known as inverse cumulative distribution function, of a random variable with the cumulative distribution function as $F$. The supremal form of $W_{p}$, denoted by $W_{p}^\infty$, is defined as
\begin{equation}\begin{aligned}
		W^\infty_p(\mu, \nu) = \sup_{(s, a)\in \mathcal{S}\times \mathcal{A}} W_p^\infty(\mu(s, a), \nu(s, a)).
\end{aligned}\end{equation}
Further, $\ell_p$ distance~\cite{elie2020dynamic} is defined as 
\begin{equation}\begin{aligned}\label{eq:lp}
		\ell_{p}(X, Y):= \left( \int_{-\infty}^{\infty} | F_{X}(\omega)-F_{Y}(\omega)|^{p} \mathrm{~d} \omega \right)^{1 / p}  = \Vert F_{X} - F_{Y} \Vert_p.
\end{aligned}\end{equation}
The $\ell_{p}$ distance and Wasserstein metric are identical at $p=1$, but are otherwise distinct. Note that when $p=2$, $\ell_p$ distance is also called Cramér distance~\cite{bellemare2017cramer} $d_{C}(X, Y)$. Also, Cramér distance has a different representation given by
\begin{equation}\begin{aligned}\label{eq:cramer}
		d_{C}(X, Y)=\mathbb{E}|X-Y|-\frac{1}{2} \mathbb{E}\left|X-X^{\prime}\right|-\frac{1}{2} \mathbb{E}\left|Y-Y^{\prime}\right|,
\end{aligned}\end{equation}
where $X^\prime$ and $Y^\prime$ are the i.i.d. copies of $X$ and $Y$. Energy distance~\cite{szekely2003statistics, ziel2020energy} is a natural extension of Cramér distance to the multivariate case, which is defined as
\begin{equation}\begin{aligned}\label{eq:energy}
		d_{E}(\mathbf{X}, \mathbf{Y})=\mathbb{E}\Vert \mathbf{X}-\mathbf{Y} \Vert -\frac{1}{2} \mathbb{E}\Vert \mathbf{X}-\mathbf{X}^{\prime} \Vert -\frac{1}{2} \mathbb{E} \Vert \mathbf{Y}-\mathbf{Y}^{\prime} \Vert,
\end{aligned}\end{equation}
where $\mathbf{X}$ and $\mathbf{Y}$ are multivariate. Moreover, the energy distance is a special case of the maximum mean discrepancy~(MMD), which is formulated as
\begin{equation}\label{eq:MMD}\begin{aligned}
		\text{MMD}(\mathbf{X}, \mathbf{Y}; k) =\left(\mathbb{E}\left[k\left(\mathbf{X}, \mathbf{X}^{\prime}\right)\right]+\mathbb{E}\left[k\left(\mathbf{Y}, \mathbf{Y}^{\prime}\right)\right]-2 \mathbb{E}[k(\mathbf{X}, \mathbf{Y})]\right)^{1 / 2},
\end{aligned}\end{equation}
where $k(\cdot, \cdot)$ is a continuous kernel on $\mathcal{X}$. In particular, if $k$ is a trivial kernel, also called the unrectified kernel, MMD degenerates to energy distance. Additionally, we further define the supreme MMD, which is a functional $\mathcal{P}(\mathcal{X})^{\mathcal{S} \times \mathcal{A}} \times \mathcal{P}(\mathcal{X})^{\mathcal{S} \times \mathcal{A}} \rightarrow \mathbb{R}$ formulated as
\begin{equation}\begin{aligned}
		\text{MMD}_\infty(\mu, \nu) = \sup_{(s, a)\in \mathcal{S}\times \mathcal{A}} \text{MMD}_\infty(\mu(s, a), \nu(s, a)).
\end{aligned}\end{equation}

We further summarize the convergence rates of the distributional Bellman operator $\mathfrak{T}^{\pi}$ under different distribution divergences.

\begin{itemize}
	\item $\mathfrak{T}^{\pi}$ is $\gamma$-contractive under the supreme form of Wassertein distance $W_p$.
	\item $\mathfrak{T}^{\pi}$ is $\gamma^{1 \\/ p}$-contractive under the supreme form of $\ell_p$ distance.
	\item $\mathfrak{T}^{\pi}$ is $\gamma^{\alpha / 2}$-contractive under $\text{MMD}_\infty$ with the kernel $k_{\alpha}(x, y)=-\|x-y\|^{\alpha}, \forall \alpha > 0$.
\end{itemize}

\textbf{Proof of Contraction in Distributional Dynamic Programming.} 
\begin{itemize}
	\item Contraction under the supreme form of Wasserstein distance is provided in Lemma 3~\cite{bellemare2017distributional}.
	\item Contraction under supreme form of $\ell_{p}$ distance can refer to Theorem 3.4~\cite{elie2020dynamic}.
	\item Contraction under $\text{MMD}_\infty$ is provided in Lemma 6~\cite{nguyen2020distributional}.
\end{itemize}

\section{Proof of Proposition~\ref{prop:regularization}}\label{appendix:prop_regularization}

\begin{proof}
	We denote two marginal random variables $U$ and $V$ with the pdf $\mu(x)$ and $\nu(y)$. We next denote the $p_\Pi(x, y)$ as the pdf for $\Pi$ in $\text{MI}_\Pi(U, V) = \text{KL}(\Pi |U \otimes V)$. We first prove that the $\text{MI}_\Pi(U, V)$ is sum-invariant, which is based on the dual form of KL divergence via the variational representation~\cite{donsker1976asymptotic,agrawal2021optimal}:
	\begin{equation}
		\begin{aligned}
			D_{KL}(X, Y) = \sup_{f \in \mathcal{L}^b} \{ \mathbb{E}_X [f(x)] - \log \left(\mathbb{E}_Y \left[ e^{f(y)}\right] \right) \}, 
		\end{aligned}
	\end{equation}
	where $\mathcal{L}^b$ is the space of bounded measurable functions. The mutual information involves two-dimensional random variables in the KL divergence. Let $U^\prime = a + U$ and $V=a+V$ with pdf $\mu^\prime$ and $\nu^\prime$, we denote the joint distribution with margins $\mu^\prime(x) = \mu(x-a)$ and $\nu^\prime(y)=\nu(y-a)$ as $\Pi^\prime(x, y)$ whose pdf $p_{\Pi^\prime}$ satisfies $p_{\Pi^\prime}(x, y)=p_\Pi(x - a, y-a)$. Based on the two-dimensional variational representation of KL divergence $\text{MI}_\Pi(U, V) = \sup_{f \in \mathcal{L}^b} \{ \mathbb{E}_{\Pi} [f(x, y)] - \log \left(\mathbb{E}_{U, V} \left[ e^{f(x, y)}\right] \right) \} $, we have:
	\begin{equation}
		\begin{aligned}\label{eq:MI_sum}
			&\text{MI}_\Pi(A+U, A+V) \\
			&= \sup_{f \in \mathcal{L}^b} \{ \mathbb{E}_{\Pi^\prime} [f(x, y)] - \log \left(\mathbb{E}_{A+U, A+V} \left[ e^{f(x, y)}\right] \right) \} \\ 
			& \stackrel{(a)}{=} \sup _{f \in \mathcal{L}^b} \{ \mathbb{E}_A \left[\mathbb{E}_{\Pi(x-a,y-a)} \left[f(x, y)  \right]  \right]- \log \left( \mathbb{E}_A \left[\mathbb{E}_{a+U, a+V} \left[ e^{f(x, y)}  \right]  \right] \right) \} \\ 
			& = \sup _{f \in \mathcal{L}^b} \{ \mathbb{E}_A \left[\mathbb{E}_{\Pi(x,y)} \left[f(x+a, y+a)  \right]  \right]- \log \left( \mathbb{E}_A \left[\mathbb{E}_{U, V} \left[ e^{f(x+a, y+a)}  \right]  \right] \right) \} \\ 
			& \stackrel{(b)}{\leq} \sup _{f \in \mathcal{L}^b} \{ \mathbb{E}_A \mathbb{E}_{\Pi} [f(x+a, y+a)]- \mathbb{E}_A \log \left( \mathbb{E}_{U, V} \left[ e^{f(x+a, y+a)}\right] \right) \} \\ 
			& = \sup _{f \in \mathcal{L}^b} \{\mathbb{E}_A [\mathbb{E}_\Pi [f(x+a, y+a)]- \log \left( \mathbb{E}_{U, V} \left[ e^{f(x+a, y+a)}\right] \right)] \} \\ 
			& \stackrel{(c)}{ \leq } \mathbb{E}_A \sup _{f \in \mathcal{L}^b} \{ \mathbb{E}_\Pi [f(x+a, y+a)] - \log \left( \mathbb{E}_{U, V} \left[ e^{f(x+a, y+a)} \right] \right) \} \\ 
			& \stackrel{(d)}{=} \mathbb{E}_A \sup _{g \in \mathcal{L}^b} \{\mathbb{E}_\Pi [g(x, y)]- \log \left(\mathbb{E}_{U, V} \left[ e^{g(x, y)}\right] \right) \} \\ 
			& =\text{MI}_\Pi(U, V), 
		\end{aligned} 
	\end{equation}
	where (a) is by the independence of $A$ between $X,Y$, and the joint cdf $\Pi$. For instance, in the one-dimensional setting, we have $\mathbb{E}_{Z = A+X}\left[f(z)\right] = \int_a \int_x f(x+a) p_A(a) p_X(x) dx da = \mathbb{E}_A \left[ \mathbb{E}_X \left[ f(x+a) \right]\right]$. (b) and (c) are by Jensen's inequality in terms of the convex function $-\log(x)$ and $\sup_f$, and (d) is because the translated cdf is still within $\mathcal{L}^b$.

	Next, we show that $\text{MI}_\Pi$ is NOT scale-sensitive or with the zero-order $\tau$. This result is directly based on the similar property of KL divergence. With a slight abuse of notations, we denote $U^\prime = a U$ and $V^\prime = a V$, whose pdfs are $\mu^\prime(x) = \frac{1}{a} \mu(\frac{x}{a})$ and $\nu^\prime(y) = \frac{1}{a} \nu(\frac{y}{a})$, respectively. The scaled joint distribution $\Pi^\prime$ with the pdf $p_{\Pi^\prime}$ satisfying $p_{\Pi^\prime}(x, y) = \frac{1}{a^2} p_{\Pi}(x/a, y/a)$. Therefore, its marginal distributions are $\int_y \frac{1}{a^2} p_{\Pi}(x/a, y/a) dy = \frac{1}{a} \mu(\frac{x}{a})$ and $\int_x \frac{1}{a^2} p_{\Pi}(x/a, y/a) dy = \frac{1}{a} \nu(\frac{y}{a})$. We thus have the following result:
	\begin{equation}
		\begin{aligned} 
			\text{MI}_\Pi(aU, aV) 
			& = \text{KL}(\Pi^\prime(x, y) |U^\prime \otimes V^\prime) \\
			& = \int p_{\Pi^\prime} (x, y) \log \frac{ p_{\Pi^\prime} (x, y)}{\mu^\prime(x) \nu^\prime(y)} dx dy \\
			& = \int \frac{1}{a^2}p_{\Pi} (x/a, y/a) \log \frac{ \frac{1}{a^2}p_{\Pi} (x/a, y/a)}{\frac{1}{a^2} \mu(x/a) \nu(y/a)} dx dy \\
			&= \int p_{\Pi} (x, y) \log \frac{ p_{\Pi} (x, y)}{\mu(x) \nu(y)} dx dy \\
			& =\text{MI}_\Pi(U, V).
		\end{aligned} 
	\end{equation}
	Putting the two properties together and given two return distributions $Z_1(s, a)$ and $Z_2(s, a)$, we have the non-expansive contraction property of the supremal form of $\text{MI}_\Pi$ as follows.
	\begin{equation}\begin{aligned}
			\text{MI}_\Pi^\infty(\mathfrak{T}^\pi Z_1, \mathfrak{T}^\pi Z_2) 
			&=\sup_{s, a}  \text{MI}_\Pi(\mathfrak{T}^\pi Z_1(s, a), \mathfrak{T}^\pi Z_2(s, a))\\
			&= \sup_{s, a}  \text{MI}_\Pi(R(s, a) + \gamma Z_1(s^\prime, a^\prime), R(s, a) + \gamma Z_2(s^\prime, a^\prime))\\
			&\stackrel{(a)}{\leq}  \text{MI}_\Pi(\gamma Z_1(s^\prime, a^\prime), \gamma Z_2(s^\prime, a^\prime))\\
			&\stackrel{(b)}{=} \text{MI}_\Pi( Z_1(s^\prime, a^\prime),  Z_2(s^\prime, a^\prime)) \\
			& \leq \sup_{s, a} \text{MI}_\Pi( Z_1(s^\prime, a^\prime),  Z_2(s^\prime, a^\prime)) \\
			& =  	\text{MI}_\Pi^\infty(Z_1, Z_2),
	\end{aligned}\end{equation}
	where (a) relies on the sum invariant property of $\text{MI}_\Pi$ and (b) utilizes the non-scale sensitive property of $\text{MI}_\Pi$. By applying the well-known Banach fixed point theorem, we have a unique return distribution when convergence of distributional dynamic programming under $\text{MI}_\Pi$ for any non-trivial joint distribution $\Pi$.
	
\end{proof}

\section{Proof of Proposition~\ref{prop:scalesum}}\label{appendix:prop_scalesum}

		\subsection{Sum Invariant Property}  
		Given two random variables $U$ and $V$ with the marginal distributions as $\mu$ and $\nu$, and a random variable $A$ that is independent of them, we aim at proving
		\begin{equation}
			\begin{aligned}
					{\mathcal{W}}_{c, \varepsilon}(A + U, A + V) & 
				\leq {\mathcal{W}}_{c, \varepsilon}(U, V).
			\end{aligned}
		\end{equation}
		According to \cite{peyre2019computational}, we have the dual form of $\mathcal{W}_{c, \varepsilon}$:
		\begin{eqnarray}
			\begin{aligned}
			\mathcal{W}_{c, \varepsilon}(U,V) & = \sup _{\varphi, \psi} \left\{ \int_x \varphi(x) \mu_x dx + \int_y \psi(y) \nu_y dy - \varepsilon \int_{x, y} \exp{\frac{\varphi(x) + \psi(y) - c(x, y)}{\varepsilon}} \mu_x \nu_y dx dy \right\} \\
			& = \sup _{\varphi, \psi} \left\{ \mathbb{E}_{\mu}\left[ \varphi(x) \right] +  \mathbb{E}_\nu \left[\psi(y)\right] - \varepsilon \mathbb{E}_{\mu, \nu} \left[\exp{\frac{\varphi(x) + \psi(y) - c(x, y)}{\varepsilon}} \right]  \right\} 
			\end{aligned}
		\end{eqnarray}
		Therefore, we have:
			\begin{eqnarray}
			\begin{aligned}
				& \mathcal{W}_{c, \varepsilon}(A+U,A+V) \\
				& =  \sup _{\varphi, \psi} \left\{ \mathbb{E}_{A+U}\left[ \varphi(x) \right] +  \mathbb{E}_{A+V} \left[\psi(y)\right] - \varepsilon \mathbb{E}_{A+U, A+V} \left[\exp{\frac{\varphi(x) + \psi(y) - c(x, y)}{\varepsilon}} \right]  \right\} \\
				& \stackrel{(a)}{=}  \sup _{\varphi, \psi} \left\{ \mathbb{E}_A \left[\mathbb{E}_{\mu}\left[ \varphi(x+a) \right] +  \mathbb{E}_{\nu} \left[\psi(y+a)\right] - \varepsilon \mathbb{E}_{\mu, \nu} \left[\exp{\frac{\varphi(x+a) + \psi(y+a) - c(x, y)}{\varepsilon}} \right]\right]  \right\} \\
				& \stackrel{(b)}{\leq} \mathbb{E}_A \left[ \sup _{\varphi, \psi} \left\{ \mathbb{E}_{\mu}\left[ \varphi(x+a) \right] +  \mathbb{E}_{\nu} \left[\psi(y+a)\right] - \varepsilon \mathbb{E}_{\mu, \nu} \left[\exp{\frac{\varphi(x+a) + \psi(y+a) - c(x, y)}{\varepsilon}} \right]\right\}  \right]   \\
				& \stackrel{(c)}{=} \sup _{f, g} \left\{ \mathbb{E}_{\mu}\left[ f(x) \right] +  \mathbb{E}_{\nu} \left[g(y)\right] - \varepsilon \mathbb{E}_{\mu, \nu} \left[\exp{\frac{f(x) + g(y) - c(x, y)}{\varepsilon}} \right]\right\} \\
				& = \mathcal{W}_{c, \varepsilon}(U, V),
			\end{aligned}
		\end{eqnarray}
		where $(a)$ relies on the same techniques used in the proof of Eq.~\ref{eq:MI_sum} in Appendix~\ref{appendix:prop_regularization}, (b) utilizes the Jensen inequality of $\sup$, and $(c)$ is based on the fact that the translation operator is still within the same functional space of $\varphi, \psi$. 
		

	\subsection{A Variant of Scale Sensitive Property when $c=-k_\alpha$}
	
	\noindent \textbf{General Conclusion.} Let $\Pi^*$ be the optimal coupling for $\mathcal{W}_{c, \varepsilon}$, we define a ratio $\lambda_\varepsilon(U, V) = \frac{\varepsilon \text{KL}(\Pi^*|\mu \otimes \nu)}{\mathcal{W}_{c, \varepsilon}}\in (0, 1)$ for any considered $U, V$ with measures $\mu, \nu$ to compare, where the denominator $\mathcal{W}_{c, \varepsilon}$ is generally non-zero. We thus have the following result:
	\begin{equation}
		\begin{aligned}
			{\mathcal{W}}_{c, \varepsilon}(aU, aV) & 
			\leq \Delta_\varepsilon(a, \alpha) 	{\mathcal{W}}_{c, \varepsilon}(U, V),
		\end{aligned}
	\end{equation}
	where the scaling factor $\Delta_\varepsilon(a, \alpha) =|a|^\alpha (1 - \sup_{U, V} \lambda_\varepsilon(U, V))  +  \sup_{U, V} \lambda_\varepsilon(U, V)) \in (|a|^\alpha, 1)$ with $\sup_{U, V} \lambda_\varepsilon(U, V) > 0$. \textit{The ratio $\lambda_\varepsilon(U, V)$ measures the proportion of the entropic regularization term over the whole divergence term $\mathcal{W}_{c, \varepsilon}$}, i.e., $\lambda_\varepsilon(U, V) = \frac{\varepsilon \text{KL}(\Pi^*|\mu \otimes \nu)}{\mathcal{W}_{c, \varepsilon}}\in (0, 1)$.  Under the mild assumption of a finite set of probability measures, we have $\sup_{U, V} \lambda_\varepsilon(U, V) > 0$. To elaborate the reason behind it, we first know that $\lambda_\varepsilon(U, V) < 1$ for any $U$ and $V$ with their measures on the probability measure set.  If this set is finite, the ratio set that contains all $\{\lambda_\varepsilon(U, V)\}$  is also finite. Based on the fact that the real set is dense, we can directly find a positive lower bound $\lambda^*$ for the ratio set, such that $\{\lambda_\varepsilon(U, V)\} \leq \lambda^* < 1$. This implies that $\sup_{U, V} \lambda_\varepsilon(U, V) = \max_{U, V} \lambda_\varepsilon(U, V) < 1$. Notably, this finite set property of the ratio avoids the extreme case that may lead to a conservative conclusion about a non-expansive distribution Bellman operator, which we will give more details later.
	
	\noindent \textbf{Scale-sensitive Property.} By definition of Sinkhorn divergence~\cite{eckstein2022quantitative, peyre2019computational}, the pdf of Gibbs kernel in the equivalent form of Sinkhorn divergence is $\mathcal{K}(U, V)$, which satisfies $\mathcal{K}(U, V) \propto e^{\frac{-c(x,y)}{\varepsilon}} \mu(x) \nu(y)$. In particular, the pdf of Gibbs kernel is defined as $\frac{d \mathcal{K}}{d\left(\mu \otimes \nu \right)}(x, y)=\frac{\exp (-c/\varepsilon)}{\int \exp (-c/\varepsilon) d\left(\mu \otimes \nu\right)}$, where the denominator is the normalization factor. After a scaling transformation, the pdf of $aU$ and $aV$ with respect to $x$ and $y$ would be $\frac{1}{a}\mu(\frac{x}{a})$ and $\frac{1}{a}\nu(\frac{y}{a})$. Thus $\mathcal{K}(aU, aV) \propto e^{\frac{-c(x,y)}{\varepsilon}} \frac{1}{a}\mu(\frac{x}{a}) \frac{1}{a}\nu(\frac{y}{a})$. In the following proof, we use the change variable formula (multivariate version) constantly, while changing the joint pdf $\pi(x, y)$ and keep the cost function term $c(x, y)$. In particular,  we denote $\Pi^*$ and $\Pi^{0}$ as the optimal joint distribution of $\mathcal{W}_{c, \varepsilon}(\mu, \nu)$ and $\mathcal{W}_{c, \varepsilon}(a\mu, a\nu)$. Then we have:
	\begin{equation}
		\begin{aligned} \label{eq:scale}
			\mathcal{W}_{c, \varepsilon}(aU, aV) & 
			= \int c(x, y) \mathrm{d} \Pi^0(x, y) + \varepsilon \text{KL}(\Pi^0|a\mu \otimes a\nu) \\
			&\leq \int c(x, y) \mathrm{d} \Pi^*(x, y) + \varepsilon \text{KL}(\Pi^*|a\mu \otimes a\nu) \\
			&\stackrel{c=-k_\alpha}{=} \int (x-y)^\alpha \frac{1}{a^2} \pi^*(\frac{x}{a}, \frac{y}{a})\mathrm{d} x \mathrm{d} y + \varepsilon \int \frac{1}{a^2} \pi^*(\frac{x}{a}, \frac{y}{a}) \log  \frac{\frac{1}{a^2} \pi^*(\frac{x}{a}, \frac{y}{a})}{\frac{1}{a^2}\mu(\frac{x}{a})\nu(\frac{y}{a})} \mathrm{d} x \mathrm{d} y \\
			&= |a|^\alpha   \int (x-y)^\alpha \pi^*(x, y)\mathrm{d} x \mathrm{d} y + \varepsilon \int  \pi^*(x, y) \log  \frac{\pi^*(x, y)}{\mu(x)\nu(y)} \mathrm{d} x \mathrm{d} y\\
			& =  |a|^\alpha   \int (x-y)^\alpha \pi^*(x, y)\mathrm{d} x \mathrm{d} y + (|a|^\alpha + 1 - |a|^\alpha) \varepsilon \int  \pi^*(x, y) \log  \frac{\pi^*(x, y)}{\mu(x)\nu(y)} \mathrm{d} x \mathrm{d} y\\
			& = |a|^\alpha	\mathcal{W}_{c, \varepsilon}(U, V) + (1 - |a|^\alpha) \varepsilon \text{KL}(\Pi^*| \mu \otimes \nu)\\
			& = \Delta_\varepsilon^{U, V}(a, \alpha) 	\mathcal{W}_{c, \varepsilon}(U, V) \\
		\end{aligned}
	\end{equation}
	where $\Delta_\varepsilon^{U, V}(a, \alpha)= |a|^\alpha + (1 - |a|^\alpha) \lambda_\varepsilon(U, V) = |a|^\alpha ( 1- \lambda_\varepsilon(U, V)) + \lambda_\varepsilon(U, V) \in (|a|^\alpha, 1)$ for $\varepsilon \in (0, +\infty)$ and $a<1$ due to the fact that $\lambda_\varepsilon(U, V) \in (0, 1)$ for any non-trivial $\mathcal{W}_{c, \varepsilon}(U, V)$. The non-trivial $\mathcal{W}_{c, \varepsilon}(U, V)$ rules out the case when the regularization term is zero, e.g., $\epsilon=0$ or the optimal coupling is the product of two margins. In other words, $\Delta_\varepsilon^{U, V}(a, \alpha)$ is a function less than 1, which depends on the two margins, including their independence and distribution similarity,  the scale factor $a$ and the order $\alpha$.

	\noindent \textbf{Ruling Out Extreme Cases in the Convergence via a Finite Set.} However, the fact that $\Delta_\varepsilon^{U, V}(a, \alpha) < 1$ can only guarantee a "conservative" non-expansive contraction rather than a desirable contraction of the distributional Bellman operator. This is because there will be extreme cases in the power of series in general, although it is very unlikely to occur  given a certain MDP in practice. For example, denote the non-constant factor as $q_k$ for the k-th distributional Bellman update, where $q_k < 1$. We can construct a counterexample as $q_k=1-1/(k+2)^2$. In this case, $\Pi_{k=1}^{+\infty} q_k = (\frac{2}{3}\frac{4}{3})(\frac{3}{4}\frac{5}{4}) \cdots > 0$ instead of the convergence to 0 and the non-zero limit can not guarantee the contraction. It also intuitively implies that iteratively applying distribution Bellman operator under $\mathcal{W}_{c, \varepsilon}$ may not lead to convergence \textit{in general by considering all possible return distributions} given the non-constant factor $\Delta_\varepsilon^{U, V}(a, \alpha)$. Although we know these extreme cases are very unlikely to happen, we have to rule out these extreme cases for a rigorous proof. As we have the assumption of a finite set of probability measures, the set of $\{\lambda_\varepsilon (U, V) \}$ is also finite. As the real set is dense, we can always find a positive constant that can be used as the contraction factor. Alternatively, we can directly use the $\sup_{U, V}\lambda_\varepsilon(U, V)$ as the uniform upper bound across the whole set of interested probability measures. Under this condition, we can immediately find a universal upper bound of $\Delta_\varepsilon^{U, V}(a, \alpha)$:
	\begin{equation}
		\begin{aligned}
			\sup_{U, V} \Delta_\varepsilon^{U, V}(a, \alpha) & =  |a|^\alpha + (1 - |a|^\alpha) \sup_{U, V} \lambda_\varepsilon(U, V)  \\
			& =  |a|^\alpha  (1 - 	\sup_{U, V} \lambda_\varepsilon(U, V))+ \sup_{U, V} \lambda_\varepsilon(U, V) \\
			& \stackrel{\cdot}{=}\Delta_\varepsilon (a, \alpha)
		\end{aligned}
	\end{equation}
	where the upper bound $\sup_{U, V} \Delta_\varepsilon^{U, V}(a, \alpha)$ has an interpolation form, which can be viewed as the convex combination between $|a|^\alpha$ and 1 with the coefficient $ \sup_{U, V} \lambda_\varepsilon(U, V)$ determined by the probability measure set. More importantly,  $\sup_{U, V} \Delta_\varepsilon^{U, V}(a, \alpha)$ is strictly less than 1, which is guaranteed by the finite set of $\{\lambda_\varepsilon(U, V)\}$ thanks to a finite set of interested probability measures. Finally, we have the variant of scale-sensitive property as follows, where the factor $\Delta_\varepsilon(a, \alpha)$ depends on $\alpha, a$ and the probability measure set.
	\begin{equation}
	\begin{aligned}
		{\mathcal{W}}_{c, \varepsilon}(aU, aV) & 
		\leq \Delta_\varepsilon(a, \alpha) 	{\mathcal{W}}_{c, \varepsilon}(U, V).
	\end{aligned}
\end{equation}

\section{Proof of Theorem~\ref{theorem:sinkhorn}}\label{appendix:sinkhorn}
	\subsection{$\varepsilon \rightarrow 0$ and $c=-k_\alpha$.} 
	
	We study the uniform convergence when $\varepsilon \rightarrow 0$. The proof is summarized from the optimal transport literature~\cite{genevay2018learning, feydy2019interpolating} and we here provide the detailed proof for completeness. On the one hand, $\mathcal{W}_{c, \varepsilon}\geq \int (x-y)^\alpha d \Pi^*(x, y) dxdy \geq W_\alpha^\alpha$ as $\text{KL}\geq 0$. We want to provide the inequality on the other side. Denote $\Pi^\prime$ as the minimizer in the Wasserstein distance $W_\alpha^\alpha$. For any $\delta > 0$, there always exists a joint distribution $\Pi^\delta$ such that 
	\begin{equation}
		\begin{aligned}
			 | \int (x-y)^\alpha d\Pi^\prime(x, y) - \int (x-y)^\alpha d\Pi^\delta(x, y) | \leq \delta
		\end{aligned}
	\end{equation}
	and $\text{KL}(\Pi^\delta | \mu \otimes \nu) < +\infty$, i.e., $ \int (x-y)^\alpha d\Pi^\delta(x, y) -  \int (x-y)^\alpha d\Pi^\prime(x, y)  \leq \delta$. One possible way to find $\Pi^\delta$ is provided in notes of Lecture 6 in Optimal Transport Course\footnote{{\url{https://lchizat.github.io/ot2021orsay.html}}} and we invite interested readers for reference. It follows that 
	\begin{equation}
		\begin{aligned}
			W_\alpha^\alpha \leq \mathcal{W}_{c, \varepsilon} \leq \int (x-y)^\alpha d\Pi^\delta(x, y) + \varepsilon \text{KL}(\Pi^\delta | \mu \otimes \nu) \leq \int (x-y)^\alpha d\Pi^\prime(x, y) + \delta + \varepsilon \text{KL}(\Pi^\delta | \mu \otimes \nu),
		\end{aligned}
	\end{equation}
	where the RHS $\int (x-y)^\alpha d\Pi^\prime(x, y) + \delta + \varepsilon \text{KL}(\Pi^\delta | \mu \otimes \nu) \rightarrow \int (x-y)^\alpha d\Pi^\prime(x, y) + \delta  = W_\alpha^\alpha + \delta$ as $\varepsilon \rightarrow 0$. As $\delta > 0$ is arbitrary, combing the two sides, it shows that $\mathcal{W}_{c, \epsilon}\rightarrow W_\alpha^{\alpha}$ as $\varepsilon \rightarrow 0$. Thus, Sinkhorn divergence maintains the properties of Wasserstein distance when $\varepsilon\rightarrow0$.
	
	When $\varepsilon = 0$, it has been shown that $W_\alpha$ can guarantee a $\gamma$-contraction property for distributional Bellman operator~\cite{bellemare2017distributional}. The crux of proof is that $W_\alpha$ is $\gamma$-scale sensitive:
	\begin{equation}
		\begin{aligned}
			W_\alpha(a U, a V) &=\left(\inf _{\Pi \in \Pi(a U, a V)} \int a^\alpha (x-y)^p d \Pi(x, y)\right)^{1 / \alpha} \\
			& \leq a \left(\inf _{\Pi \in \Pi(U, V)} \int  (x-y)^p d \Pi(x, y)\right)^{1 / \alpha} \\
			& = a W_\alpha(U, V),
		\end{aligned}
	\end{equation}
	where the inequality comes from the change of optimal joint distribution. Therefore, $W_\alpha(a U, a V)\leq a W_\alpha(U, V)$ guarantees a $\gamma$-contraction property for the distributional Bellman operator. As such, for $W_\alpha^\alpha$, when $\varepsilon=0$, it suggest that $\overline{\mathcal{W}}_{c, 0} = W_\alpha^\alpha$ corresponds to a $\gamma^\alpha$-contraction for the distributional Bellman operator $\mathfrak{T}^\pi$.

	\subsection{ $\varepsilon \rightarrow \infty$ and $c=-k_\alpha$.} Our complete proof is inspired by \cite{ramdas2017wasserstein,genevay2018learning}. Recap the definition of squared MMD is
	\begin{equation}
		\begin{aligned}
			\mathbb{E}\left[k\left(\mathbf{X}, \mathbf{X}^{\prime}\right)\right]+\mathbb{E}\left[k\left(\mathbf{Y}, \mathbf{Y}^{\prime}\right)\right]-2 \mathbb{E}[k(\mathbf{X}, \mathbf{Y})].
		\end{aligned}
	\end{equation}
	When the kernel function $k$ degenerates to an unrectified $k_\alpha(x, y):=-\Vert x - y \Vert^\alpha$ for $\alpha \in (0, 2)$, the squared MMD would degenerate to
	\begin{equation}
		\begin{aligned}
			2 \mathbb{E}\Vert \mathbf{X} - \mathbf{Y} \Vert^\alpha - \mathbb{E} \Vert \mathbf{X} -  \mathbf{X}^{\prime} \Vert^\alpha - \mathbb{E}\Vert \mathbf{Y} - \mathbf{Y}^{\prime}\Vert^\alpha.
		\end{aligned}
	\end{equation}
	where  $\mathbf{X}, \mathbf{X}^{\prime} \stackrel{\text { i.i.d. }}{\sim} \mu, \mathbf{Y}, \mathbf{Y}^{\prime} \stackrel{\text { i.i.d. }}{\sim} \nu$ and $\mathbf{X}, \mathbf{X}^{\prime},\mathbf{Y}, \mathbf{Y}^{\prime}$ are mutually independent. On the other hand, by definition, we have the Sinkhorn loss as
	\begin{equation}
		\begin{aligned}
			\overline{\mathcal{W}}_{c, \infty}(\mu, \nu)=2 \mathcal{W}_{c, \infty}(\mu, \nu)-\mathcal{W}_{c, \infty}(\mu, \mu)-\mathcal{W}_{c, \infty}(\nu, \nu).
		\end{aligned}
	\end{equation}
	Denoting $\Pi_\varepsilon$ be the unique minimizer for $\overline{\mathcal{W}}_{c, \varepsilon}$, it holds that $\Pi_{\varepsilon} \rightarrow \mu \otimes \nu$ as $\varepsilon \rightarrow \infty$, which is the product of two marginal distributions. That being said, $\mathcal{W}_{c, \infty}(\mu, \nu) \rightarrow \int c(x, y) \mathrm{d} \mu(x) \mathrm{d} \nu(y) + 0 = \int c(x, y) \mathrm{d} \mu(x) \mathrm{d} \nu(y)$. \textit{One important proof insight here is although $\varepsilon \rightarrow +\infty$, the KL term tends to zero, which is faster than $\varepsilon$. Therefore, the whole regularization term still tends to 0 as $\varepsilon \rightarrow +\infty$.} If $c=-k_\alpha=\Vert x - y \Vert^\alpha$, we eventually have $\mathcal{W}_{-k_\alpha, \infty}(\mu, \nu) \rightarrow   \int \Vert x - y \Vert^\alpha \mathrm{d} \mu(x) \mathrm{d} \nu(y) =  \mathbb{E}\Vert \mathbf{X} - \mathbf{Y}\Vert^\alpha$, where $\mu$ and $\nu$ can be inherently correlated, although the minimizer degenerates to the product of the two marginal distributions. Finally, we can have
	\begin{equation}
		\begin{aligned}
			\overline{\mathcal{W}}_{-k_\alpha, \infty} \rightarrow  2 \mathbb{E}\Vert \mathbf{X} - \mathbf{Y} \Vert^\alpha - \mathbb{E} \Vert \mathbf{X} -  \mathbf{X}^{\prime} \Vert^\alpha - \mathbb{E}\Vert \mathbf{Y} - \mathbf{Y}^{\prime}\Vert^\alpha,
		\end{aligned}
	\end{equation}
	which is exactly the form of squared MMD with the unrectified kernel $k_\alpha$. Now the key is to prove that $\Pi_{\varepsilon} \rightarrow \mu \otimes \nu$ as $\varepsilon \rightarrow \infty$. We give the detailed proof as follows.
	
	Firstly, it is apparent that $\mathcal{W}_{c, \varepsilon}(\mu, \nu) \leq \int c(x, y) \mathrm{d} \mu(x) \mathrm{d} \nu(y)$ as $\mu \otimes \nu \in \Pi(\mu, \nu)$. Let $\{ \varepsilon_k \}$ be a positive sequence that diverges to $\infty$, and $\Pi_k$ be the corresponding sequence of unique minimizers for $\mathcal{W}_{c, \varepsilon}$. According to the optimality condition, it must be the case that $\int c(x, y) \mathrm{d} \Pi_k + \varepsilon_k \text{KL}(\Pi_k, \mu \otimes \nu) \leq \int c(x, y) \mathrm{d}  \mu \otimes \nu +0$ (when $ \Pi(\mu, \nu) = \mu \otimes \nu$). Thus, 
	$$\mathrm{KL}\left(\Pi_k, \mu \otimes \nu\right) \leqslant \frac{1}{\varepsilon_{k}}\left(\int c \mathrm{~d} \mu \otimes \nu-\int c \mathrm{~d} \Pi_{k}\right) \rightarrow 0.$$
	
	Besides, by the compactness of $\Pi(\mu, \nu)$, we can extract a converging subsequence $\Pi_{n_k} \rightarrow \Pi_\infty$. Since KL is weakly lower-semicontinuous, it holds that 
	$$\mathrm{KL}\left(\Pi_{\infty}, \mu \otimes \nu\right) \leqslant \lim_{k \rightarrow \infty} \inf  \mathrm{KL}\left(\Pi_{n_{k}}, \mu \otimes \nu\right)=0$$
	Hence $\Pi_\infty = \mu \otimes \nu$. That being said that the optimal coupling is simply the product of the marginals, indicating that $\Pi_{\varepsilon} \rightarrow \mu \otimes \nu$ as $\varepsilon \rightarrow \infty$. As a special case, when $\alpha=1$, $\overline{\mathcal{W}}_{-k_1, \infty}(u, v)$ is equivalent to the energy distance 
	\begin{equation}\begin{aligned}
			d_{E}(\mathbf{X}, \mathbf{Y}):=2\mathbb{E}\Vert \mathbf{X}-\mathbf{Y} \Vert -   \mathbb{E}\Vert \mathbf{X}-\mathbf{X}^{\prime} \Vert - \mathbb{E} \Vert \mathbf{Y}-\mathbf{Y}^{\prime} \Vert.
	\end{aligned}\end{equation}
	
	In summary, if the cost function is the rectified kernel $k_\alpha$, it is the case that $\overline{\mathcal{W}}_{-k_\alpha, \varepsilon}$ converges to the squared MMD as $\varepsilon \rightarrow \infty$. According to \cite{nguyen2020distributional}, $\mathfrak{T}^\pi$ is $\gamma^{\alpha / 2}$-contractive in the supremal form of MMD with the unrectified kernel $k_\alpha$. As $\overline{\mathcal{W}}_{c, \varepsilon}(\mu, \nu) \rightarrow \text{MMD}_{k_\alpha}^2(\mu, \nu)$, which is a squared MMD instead of MMD, it implies that $\mathfrak{T}^\pi$ is $\gamma^{\alpha}$-contractive under the squared MMD / $\overline{\mathcal{W}}_{c, +\infty}$.

	\subsection{$\varepsilon \in (0, +\infty)$ and $c = -\kappa_\alpha$}  
	
	In the proof of Proposition~\ref{prop:scalesum}, we have shown that the Sinkhorn loss ${\mathcal{W}}_{c, \varepsilon}$ satisfies the sum-invariant~\textbf{(I)} and a new variant of scale-sensitive properties as follows:
	\begin{equation}
		\begin{aligned}
				{\mathcal{W}}_{c, \varepsilon}(A + U, A + V) & 
			\leq {\mathcal{W}}_{c, \varepsilon}(U, V)\\
			{\mathcal{W}}_{c, \varepsilon}(aU, aV) & 
			\leq \Delta_\varepsilon(a, \alpha) 	{\mathcal{W}}_{c, \varepsilon}(U, V).
		\end{aligned}
	\end{equation}
	The Sinkhorn divergence $\overline{\mathcal{W}}_{c, \varepsilon}$ is defined by additionally subtracting two self-distance terms ($	{\mathcal{W}}_{c, \varepsilon}(\mu, \mu)$ and $	{\mathcal{W}}_{c, \varepsilon}(\nu, \nu)$) based on $	{\mathcal{W}}_{c, \varepsilon}(\mu, \nu)$ in order to guarantee the non-negativity, tri-angularity and metric properties. These two self-distance terms do not change the \textbf{(I)} and \textbf{(S)} properties when extending ${\mathcal{W}}_{c, \varepsilon}$  to $\overline{\mathcal{W}}_{c, \varepsilon}$, and some proof techniques can refer to Section 2 in \cite{feydy2019interpolating}. The only difference is that the scaling factor will be $\overline{\Delta}_\varepsilon^{U, V}(a, \alpha)$, which is the counterpart of Eq.~\ref{eq:scale} satisfying
		\begin{equation}
		\begin{aligned}
			\overline{\mathcal{W}}_{c, \varepsilon}(aU, aV) & 
			\leq \overline{\Delta}^{U, V}_\varepsilon(a, \alpha) 	\overline{\mathcal{W}}_{c, \varepsilon}(U, V).
		\end{aligned}
	\end{equation}
	where $\overline{\Delta}_\varepsilon^{U, V}(a, \alpha)=  |a|^\alpha ( 1- \overline{\lambda}_\varepsilon(U, V)) + \overline{\lambda}_\varepsilon(U, V) \in (|a|^\alpha, 1)$ for $\varepsilon \in (0, +\infty)$ and $a<1$ due to the fact that $\overline{\lambda}_\varepsilon(U, V) \in (0, 1)$ for any non-trivial $\overline{\mathcal{W}}_{c, \varepsilon}(U, V)$. The new ratio 
	$\overline{\lambda}_\varepsilon(U, V) = \frac{\varepsilon \text{KL}(\Pi^*|\mu \otimes \nu)}{\overline{\mathcal{W}}_{c, \varepsilon}}\in (0, 1)$ for any considered $U, V$ with measures $\mu, \nu$ in the interested probability measure set. In particular, in the context of distributional RL, the set of interested probability measures would be the return distribution set of $\{Z(s, a) \}$ for $s \in \mathcal{S}$ and  $a \in \mathcal{A}$ in a given finite MDP. We now want to find the universal upper bound $\overline{\Delta}_\varepsilon(a, \alpha)$, which is defined as
	\begin{equation}
			\begin{aligned}
			\overline{\Delta}_\varepsilon(a, \alpha) = |a|^\alpha (1 - \sup_{U, V} \overline{\lambda}_\varepsilon(U, V))  +  \sup_{U, V} \overline{\lambda}_\varepsilon(U, V)) \in (|a|^\alpha, 1).
		\end{aligned}
	\end{equation}
	Following the proof in Appendix~\ref{appendix:prop_scalesum}, the finite MDP guarantees a finite ratio set of $\{\overline{\lambda}_\varepsilon(U, V) \}$, and thus we can find a universal upper bound $\overline{\lambda}^*$ of the ratio set such that $\{\overline{\lambda}_\varepsilon(U, V) \} \leq \overline{\lambda}^* < 1$. This also implies that $\sup_{U, V} \overline{\lambda}_\varepsilon(U, V) \in (0, 1)$ and thus the scaling factor $\overline{\Delta}_\varepsilon(a, \alpha) \in (|a|^\alpha, 1)$, which is strictly less than 1. Therefore, we have the  \textbf{(I)} and \textbf{(S)} properties of $\overline{\mathcal{W}}_{c, \varepsilon}$:
		\begin{equation}
		\begin{aligned}
			\overline{\mathcal{W}}_{c, \varepsilon}(A + U, A + V) & 
			\leq \overline{\mathcal{W}}_{c, \varepsilon}(U, V)\\
			\overline{\mathcal{W}}_{c, \varepsilon}(aU, aV) & 
			\leq \overline{\Delta}_\varepsilon(a, \alpha) 	\overline{\mathcal{W}}_{c, \varepsilon}(U, V).
		\end{aligned}
	\end{equation}

	Putting all together, we now derive the convergence of distributional Bellman operator $\mathfrak{T}^\pi$ under the supreme form of $\overline{\mathcal{W}}_{c, \varepsilon}$, i.e., $\overline{\mathcal{W}}_{c, \varepsilon}^\infty$:
	\begin{equation}\begin{aligned}
			\overline{\mathcal{W}}_{c, \varepsilon}^\infty(\mathfrak{T}^\pi Z_1, \mathfrak{T}^\pi Z_2) 
			&=\sup_{s, a} \overline{\mathcal{W}}_{c, \varepsilon}(\mathfrak{T}^\pi Z_1(s, a), \mathfrak{T}^\pi Z_2(s, a))\\
			&=\sup_{s, a} \overline{\mathcal{W}}_{c, \varepsilon}(R(s, a) + \gamma Z_1(s^\prime, a^\prime), R(s, a) + \gamma Z_2(s^\prime, a^\prime))\\
			&\stackrel{(a)}{\leq} \overline{\mathcal{W}}_{c, \varepsilon}(\gamma Z_1(s^\prime, a^\prime), \gamma Z_2(s^\prime, a^\prime))\\
			&\stackrel{(b)}{\leq}  \overline{\Delta}_\varepsilon^{Z_1(s^\prime, a^\prime), Z_2(s^\prime, a^\prime)}(\gamma, \alpha) \overline{\mathcal{W}}_{c, \varepsilon}(Z_1(s^\prime, a^\prime), Z_2(s^\prime, a^\prime)) \\
			& \leq \sup_{s^\prime, a^\prime} \overline{\Delta}_\varepsilon^{Z_1(s^\prime, a^\prime), Z_2(s^\prime, a^\prime)}(\gamma, \alpha) \sup_{s^\prime, a^\prime} \overline{\mathcal{W}}_{c, \varepsilon}(Z_1(s^\prime, a^\prime), Z_2(s^\prime, a^\prime)) \\
			& =  \overline{\Delta}_\varepsilon(\gamma, \alpha) \overline{\mathcal{W}}_{c, \varepsilon}^\infty(Z_1, Z_2)
	\end{aligned}\end{equation}
	where the inequality (a) is based on the sum invariant property \textbf{(I)} of Sinkhorn divergence. (b) is based on the new variant of scale-sensitive property \textbf{(S)} of Sinkhorn divergence and the leverage of $c=-k_\alpha$. Notably, $\overline{\Delta}_\varepsilon(\gamma, \alpha) \in (|\gamma|^\alpha, 1)$ is an MDP-dependent constant (determined by the return distribution set), which is also determined by $\gamma, \varepsilon$ and $\alpha$. As such, we conclude that distributional Bellman operator is \textit{at least} $\overline{\Delta}_\varepsilon(\gamma, \alpha) $-contractive, where the contraction factor $\overline{\Delta}_\varepsilon(\gamma, \alpha)$ is strictly less than 1 in a given finite MDP. Based on the existing Banach fixed point theorem, we have a unique optimal return distribution by applying the distributional Bellman operator $\mathfrak{T}^\pi$ in the distributional dynamic programming when convergence.
	

\section{Proof of Corollary~\ref{corollary:multi} }\label{appendix:corollary_multi}

\begin{proof}
		The contraction conclusion that extends to the multi-dimensional return distributions is straightforward. As the definition of Sinkhorn divergence inherently allows the multi-dimensional measures, the sum-invariant and the variant of scale-sensitive properties hold naturally. Specifically, after recapping to proof of these properties, we only need to change $c(x,y)=(x-y)^\alpha$ to $c(\mathbf{x},  \mathbf{y})=\Vert \mathbf{x} - \mathbf{y} \Vert^\alpha$ and re-define two $d$-dimensional random vector $\mathbf{U}$ and $\mathbf{V}$ with the $d$-dimensional probability measure $\mathbf{\mu}$ and $\mathbf{\nu}$. Therefore, the \textbf{(I)} and \textbf{(S)} properties in the multi-dimensional reward settings are:
	\begin{equation}
		\begin{aligned}
			\overline{\mathcal{W}}_{c, \varepsilon}(\mathbf{A} + \mathbf{U}, \mathbf{A} + \mathbf{V}) & 
			\leq \overline{\mathcal{W}}_{c, \varepsilon}(\mathbf{U}, \mathbf{V})\\
			\overline{\mathcal{W}}_{c, \varepsilon}(a\mathbf{U}, a\mathbf{V}) & 
			\leq \overline{\Delta}_\varepsilon(a, \alpha) 	\overline{\mathcal{W}}_{c, \varepsilon}(\mathbf{U}, \mathbf{V}),
		\end{aligned}
	\end{equation}
	where $\mathbf{A}$ is a $d$-dimensional random vector independent of $\mathbf{U}$ and $\mathbf{V}$.
	
	By leveraging these two properties, we now derive the convergence of distributional Bellman operator $\mathfrak{T}_d^\pi$ under $\overline{\mathcal{W}}_{c, \varepsilon}^\infty$ in the joint return distribution setting. Given two $d$-dimensional return distributions $\mathbf{Z}_1$ and $\mathbf{Z}_2$, we have
	\begin{equation}\begin{aligned}
			\overline{\mathcal{W}}_{c, \varepsilon}^\infty(\mathfrak{T}_d^\pi \mathbf{Z}_1, \mathfrak{T}_d^\pi \mathbf{Z}_2) 
			&=\sup_{s, a} \overline{\mathcal{W}}_{c, \varepsilon}(\mathfrak{T}_d^\pi \mathbf{Z}_1(s, a), \mathfrak{T}_d^\pi \mathbf{Z}_2(s, a))\\
			&=\sup_{s, a} \overline{\mathcal{W}}_{c, \varepsilon}(\mathbf{R}(s, a) + \gamma \mathbf{Z}_1(s^\prime, a^\prime), \mathbf{R}(s, a) + \gamma \mathbf{Z}_2(s^\prime, a^\prime))\\
			&\stackrel{(a)}{\leq} \overline{\mathcal{W}}_{c, \varepsilon}(\gamma \mathbf{Z}_1(s^\prime, a^\prime), \gamma \mathbf{Z}_2(s^\prime, a^\prime))\\
			&\stackrel{(b)}{\leq}  \overline{\Delta}_\varepsilon^{\mathbf{Z}_1(s^\prime, a^\prime), \mathbf{Z}_2(s^\prime, a^\prime)}(\gamma, \alpha) \overline{\mathcal{W}}_{c, \varepsilon}(\mathbf{Z}_1(s^\prime, a^\prime), \mathbf{Z}_2(s^\prime, a^\prime)) \\
			& \leq \sup_{s^\prime, a^\prime} \overline{\Delta}_\varepsilon^{\mathbf{Z}_1(s^\prime, a^\prime), \mathbf{Z}_2(s^\prime, a^\prime)}(\gamma, \alpha) \sup_{s^\prime, a^\prime} \overline{\mathcal{W}}_{c, \varepsilon}(\mathbf{Z}_1(s^\prime, a^\prime), \mathbf{Z}_2(s^\prime, a^\prime)) \\
			& =  \overline{\Delta}_\varepsilon(\gamma, \alpha) \overline{\mathcal{W}}_{c, \varepsilon}^\infty(\mathbf{Z}_1, \mathbf{Z}_2)
	\end{aligned}\end{equation}
	where the inequality (a) is based on the sum invariant property \textbf{(I)} of Sinkhorn divergence that cancels the additive $d$-dimensional random vector $\mathbf{R}(s, a)$. (b) is based on the new variant of scale-sensitive property \textbf{(S)} of Sinkhorn divergence and the leverage of $c=-k_\alpha$, where the contraction factor $\overline{\Delta}_\varepsilon(\gamma, \alpha)$ will depend on the set of $d$-dimensional probability measures/distributions. Notably, the analysis of $\overline{\Delta}_\varepsilon(\gamma, \alpha)$ in the one-dimensional return setting established in Appendix~\ref{appendix:prop_scalesum} and Appendix~\ref{appendix:sinkhorn} is also applicable in the multi-dimensional setting.
\end{proof}

\section{Algorithm: Sinkhorn Iterations and Sinkhorn Distributional RL}\label{appendix:algorithm}

\begin{algorithm}[htbp]
	\setstretch{0.01}
	\caption{Sinkhorn Iterations to Approximate $\overline{\mathcal{W}}_{c, \varepsilon}\left(\left\{Z_{i}\right\}_{i=1}^{N},\left\{\mathfrak{T} Z_{j}\right\}_{j=1}^{N}\right)$}
	\textbf{Input}: Two samples sequences $\left\{Z_{i}\right\}_{i=1}^{N},\left\{\mathfrak{T} Z_{j}\right\}_{j=1}^{N}$, number of iterations $L$ and hyperparameter $\varepsilon$.
	\begin{algorithmic}[1] 
		\STATE $\hat{c}_{i,j}=c(Z_i, \mathfrak{T} Z_{j})$ for $\forall i=1,...,N, j=1,...,N$
		\STATE $\mathcal{K}_{i,j}=\exp(-\hat{c}_{i, j}/\varepsilon)$
		\STATE $b_0 \leftarrow \mathbf{1}_N$
		\FOR{$l=1,2,...,L$}
		\STATE $a_l\leftarrow \frac{\mathbf{1}_N}{\mathcal{K} b_{l-1}}$, $b_l\leftarrow \frac{\mathbf{1}_N}{\mathcal{K} a_l}$
		\ENDFOR
		\STATE $\widehat{\overline{\mathcal{W}}}_{c, \varepsilon}\left(\left\{Z_{i}\right\}_{i=1}^{N},\left\{\mathfrak{T} Z_{j}\right\}_{j=1}^{N}\right) = \left< (K \odot \hat{c}) b, a \right>$
	\end{algorithmic}
	\textbf{Return}:  $\widehat{\overline{\mathcal{W}}}_{c, \varepsilon}\left(\left\{Z_{i}\right\}_{i=1}^{N},\left\{\mathfrak{T} Z_{j}\right\}_{j=1}^{N}\right)$
	\label{alg:sinkhorn_iterations}
\end{algorithm}

Given two sample sequences $\left\{Z_{i}\right\}_{i=1}^{N},\left\{\mathfrak{T} Z_{j}\right\}_{j=1}^{N}$ in the distributional RL algorithm, the optimal transport distance is equivalent to the form:
\begin{equation}
\begin{aligned}
\min _{P \in \mathbb{R}_{+}^{N \times N}}\left\{\langle P, \hat{c}\rangle ; P \mathbf{1}_{N}=\mathbf{1}_{N}, P^{\top} \mathbf{1}_{N}=\mathbf{1}_{N}\right\},
\end{aligned}
\end{equation}
where the empirical cost function is $\hat{c}_{i,j}=c(Z_{i}, \mathfrak{T} Z_{j})$. By adding entropic regularization on optimal transport distance, Sinkhorn divergence can be viewed to restrict the search space of $P$ in the following scaling form:
\begin{equation}
\begin{aligned}
P_{i, j}=a_{i} \mathcal{K}_{i, j} b_{j},
\end{aligned}
\end{equation}
where $\mathcal{K}_{i, j} = e^{-\hat{c}_{i, j} / \varepsilon}$ is the Gibbs kernel defined in Eq.~\ref{eq:sinkhorn_gibbs}. This allows us to leverage iterations regarding the vectors $a$ and $b$. More specifically, we initialize $b_0 = \mathbf{1}_N$, and then the Sinkhorn iterations are expressed as
\begin{equation}
	\begin{aligned}
		a_{l+1} \leftarrow \frac{\mathbf{1}_{N}}{\mathcal{K} b_{l}} \quad \text { and } \quad b_{l+1} \leftarrow \frac{\mathbf{1}_{N}}{\mathcal{K}^{\top} a_{l+1}},
	\end{aligned}
\end{equation}
where $\frac{\cdot}{\cdot}$ indicates an entry-wise division. Combining Sinkhorn Iteration in Algorithm~\ref{alg:sinkhorn_iterations} and the generic update of Sinkhorn Distributional RL in Algorithm~\ref{alg:sinkhorn}, we provide a full version of Sinkhorn Distributional RL algorithm in Algorithm~\ref{algorithm:full}.

\begin{algorithm}[htbp]
	\caption{Sinkhorn Distributional RL}
	\begin{algorithmic}[1]
		\REQUIRE Number of generated samples $N$, the kernel $k$ (e.g., unrectified kernel), discount factor $\gamma \in [0, 1]$, learning rate $\alpha$, replay buffer $M$, main network $Z_{\theta}$, target network $Z_{\theta^*}$, number of iterations $L$, hyperparameter $\varepsilon$, and a behavior policy $\pi$ based on $Z_\theta$ following an $\epsilon$-greedy rule
		\STATE Initialize $\theta$ and $\theta^* \leftarrow \theta$
		\FOR{$t = 1, 2, \ldots$}
		\STATE Take action $a_t \sim \pi(\cdot | s_t; \theta)$, receive reward $r_t \sim R(\cdot | s_t, a_t)$, and observe $s_{t+1} \sim P(\cdot | s_t, a_t)$
		\STATE Store $(s_t, a_t, r_t, s_{t+1})$ to the replay buffer $M$
		\STATE Randomly draw a batch of transition samples $(s, a, r, s')$ from the replay buffer $M$
		\STATE Compute a greedy action: 
		$a^* = \arg\max_{a' \in A} \frac{1}{N} \sum_{i=1}^N Z_{\theta^*}(s', a')_i$
		
		\STATE Compute the target Bellman return distribution:
		$\mathfrak{T} Z_{i} \leftarrow r+\gamma Z_{\theta^*}\left(s^{\prime}, a^{*}\right)_{i}, \forall 1 \leq i \leq N$
		
		\STATE Evaluate Sinkhorn divergence via Sinkhorn Iterations in Algorithm~\ref{alg:sinkhorn_iterations}:
		
		$$\overline{\mathcal{W}}_{c, \varepsilon}\left(\left\{Z_{\theta}(s, a)_{i}\right\}_{i=1}^{N},\left\{\mathfrak{T}  Z_{j}\right\}_{j=1}^{N}\right)$$
		
		\STATE Update the main network $Z_\theta$:
		$\theta \leftarrow \theta - \alpha \nabla_{\theta} \overline{\mathcal{W}}_{c, \varepsilon}\left(\left\{Z_{\theta}(s, a)_{i}\right\}_{i=1}^{N},\left\{\mathfrak{T}  Z_{j}\right\}_{j=1}^{N}\right)$
		
		\STATE Periodically update the target network $\theta^* \leftarrow \theta$
		\ENDFOR
	\end{algorithmic}
	\label{algorithm:full}
\end{algorithm}

\section{Summary Table  for Human Normalized Scores~(HNS)}\label{appendix:table_HNS}

\begin{table}[htbp]
	\centering
	\scalebox{1.0}{
		\begin{tabular}{c|r|r|r|c}
			\hline
			\hline
			& \textbf{Mean} & \textbf{IQM~(5\%)} & \textbf{Median} & \textbf{$>$DQN} \\
			\hline \text{DQN } & 452.6 \% & 181.2 \%& 32.8 \% &0 \\
			\text{C51 } & 640.2 \% &368.5 \%& 68.5 \% &  35 \\
			\text{QR-DQN-1} & 780.9 \%&401.9 \% & 85.8 \% &  38 \\
			\text{MMD-DQN} & 781.7 \% &428.4 \%& \textbf{96.6} \% &  37 \\
			\hline
			\text{SinkhornDRL} & \bf{1306.1} \% & \bf{477.0} \%& \underline{91.1} \% & \bf{41} \\
			\hline
			\hline
		\end{tabular}
	}
	\vspace{0.2cm}
	\caption{Evaluation of best Human Normalized Scores~(HNS) across 55 Atari games. Results are averaged over 3 seeds. Our proposed SinkhornDRL achieves the best performance in terms of Mean and IQM($5\%$) HNS as well as the ``$>\text{DQN}$'' metric, and is on par with MMD-DQN in terms of Median of HNS.} 
		\label{table:allresults_summary}
\end{table}

\clearpage
\section{Learning Curves on 55 Atari Games}\label{appendix:experiment_games}
\begin{figure}[htbp]
	\centering
	\begin{subfigure}[t]{0.16\textwidth}
		\centering
		\includegraphics[width=\textwidth]{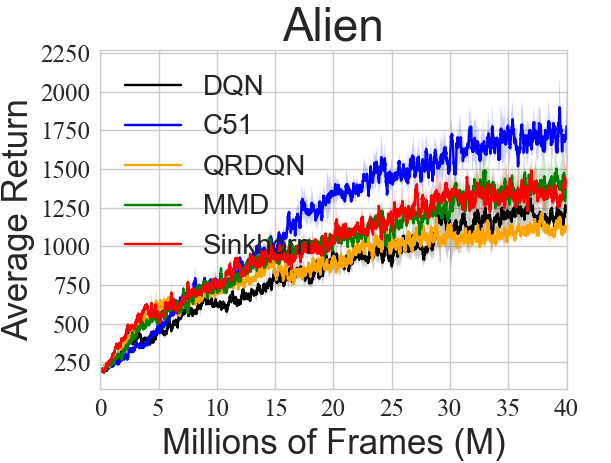}
	\end{subfigure}
	\begin{subfigure}[t]{0.16\textwidth}
		\centering
		\includegraphics[width=\textwidth]{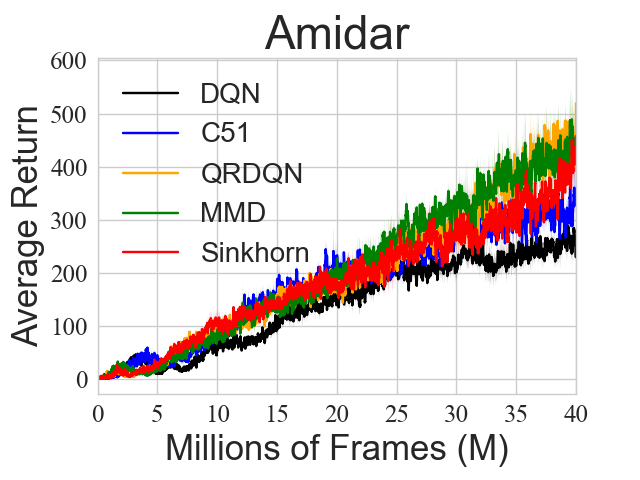}
	\end{subfigure}
	\begin{subfigure}[t]{0.16\textwidth}
		\centering
		\includegraphics[width=\textwidth]{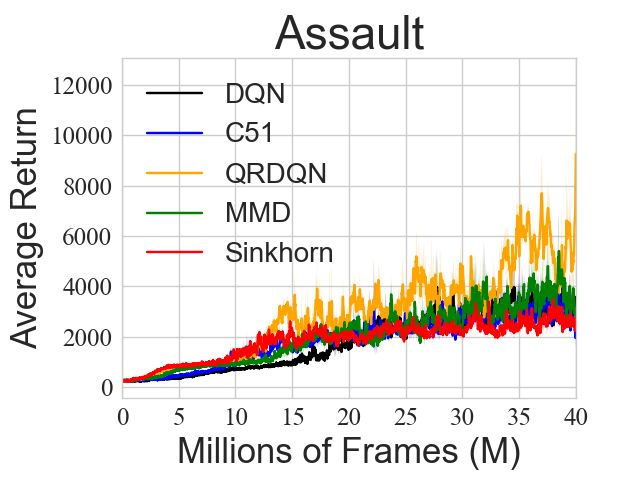}
	\end{subfigure}
	\begin{subfigure}[t]{0.16\textwidth}
		\centering
		\includegraphics[width=\textwidth]{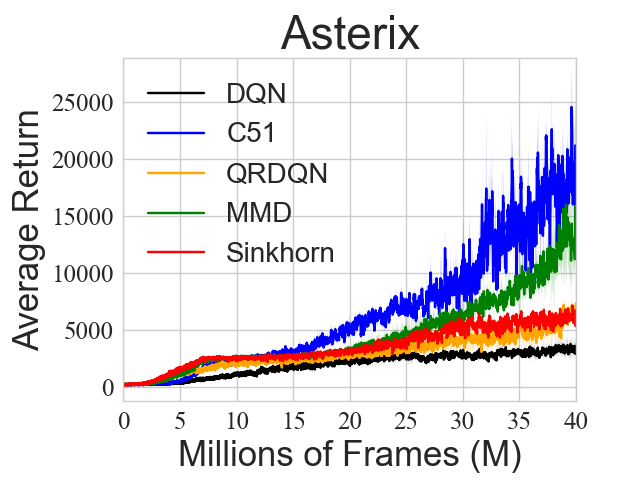}
	\end{subfigure}
	\begin{subfigure}[t]{0.16\textwidth}
		\centering
		\includegraphics[width=\textwidth]{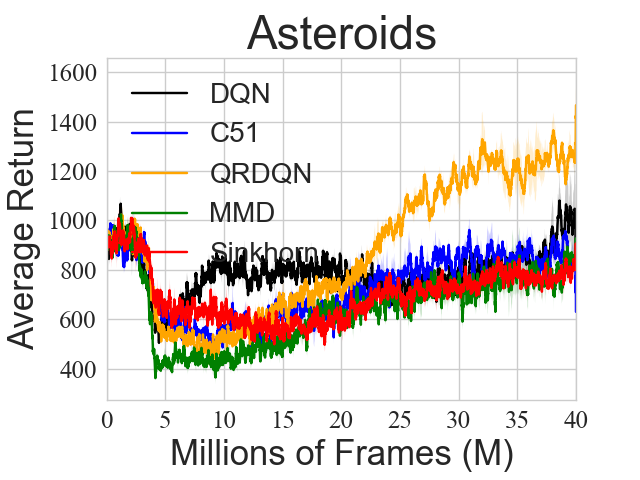}
	\end{subfigure}
	\begin{subfigure}[t]{0.16\textwidth}
		\centering
		\includegraphics[width=\textwidth]{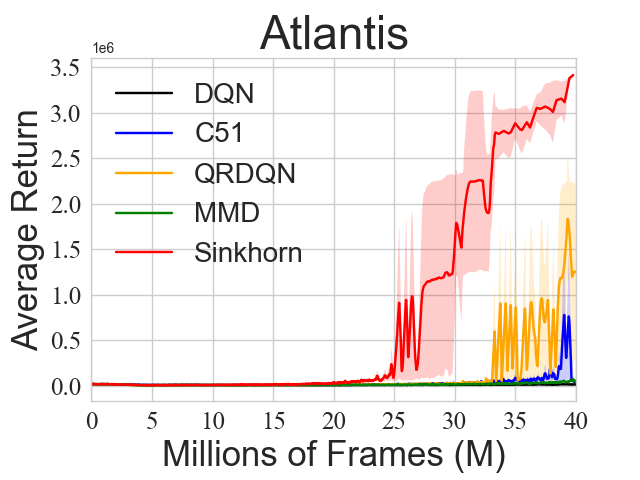}
	\end{subfigure}
	\begin{subfigure}[t]{0.16\textwidth}
		\centering
		\includegraphics[width=\textwidth]{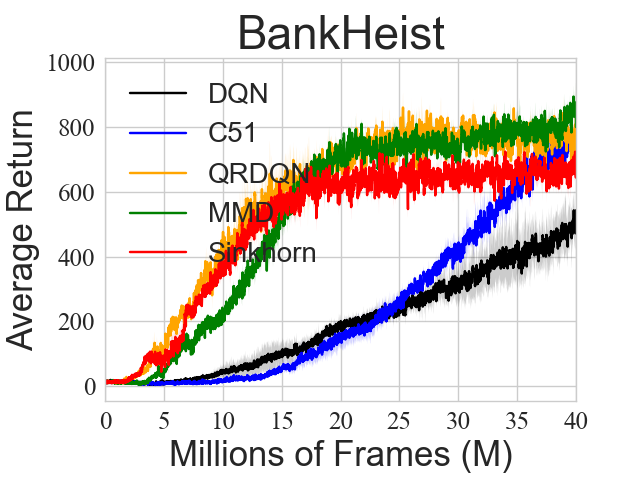}
	\end{subfigure}
	\begin{subfigure}[t]{0.16\textwidth}
		\centering
		\includegraphics[width=\textwidth]{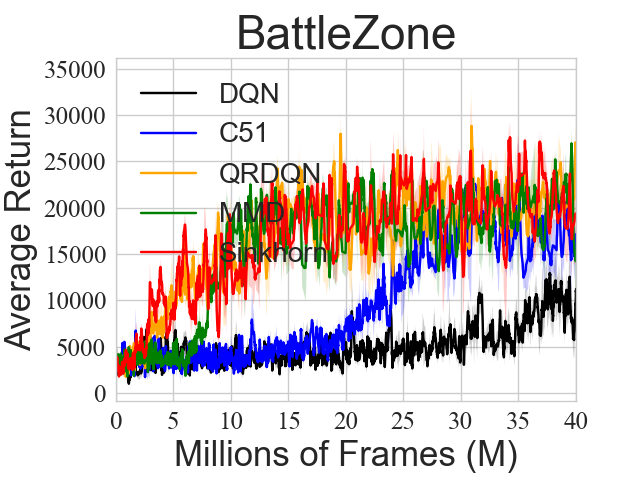}
	\end{subfigure}
	\begin{subfigure}[t]{0.16\textwidth}
		\centering
		\includegraphics[width=\textwidth]{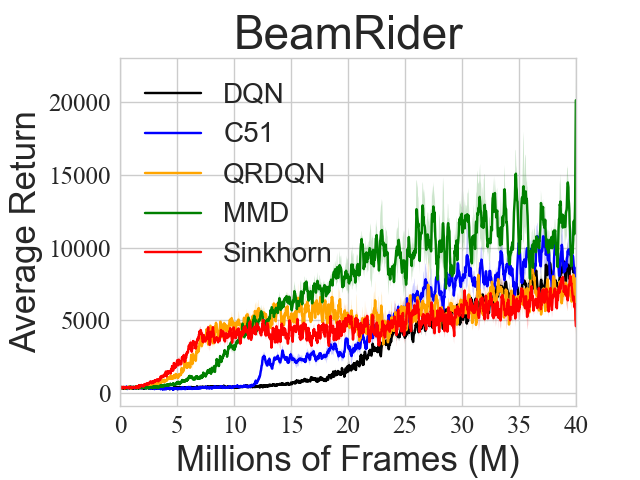}
	\end{subfigure}
	\begin{subfigure}[t]{0.16\textwidth}
		\centering
		\includegraphics[width=\textwidth]{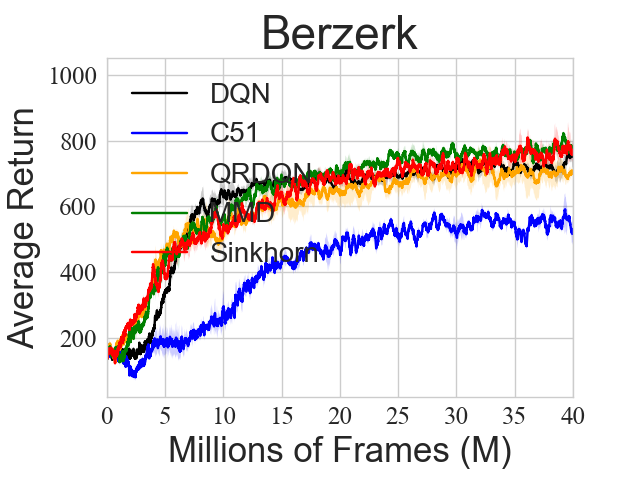}
	\end{subfigure}
	\begin{subfigure}[t]{0.16\textwidth}
		\centering
		\includegraphics[width=\textwidth]{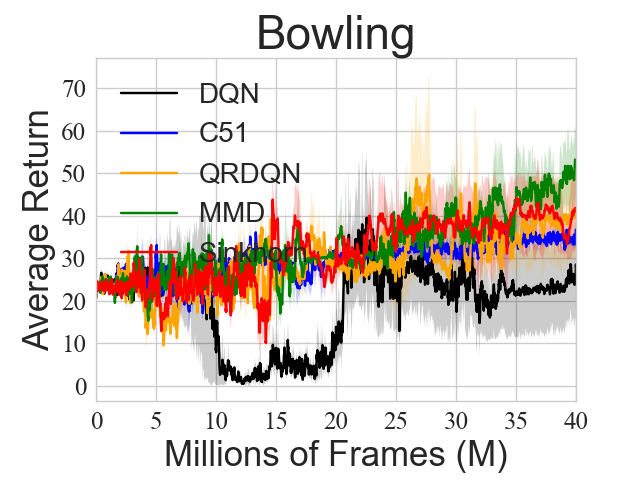}
	\end{subfigure}
	\begin{subfigure}[t]{0.16\textwidth}
		\centering
		\includegraphics[width=\textwidth]{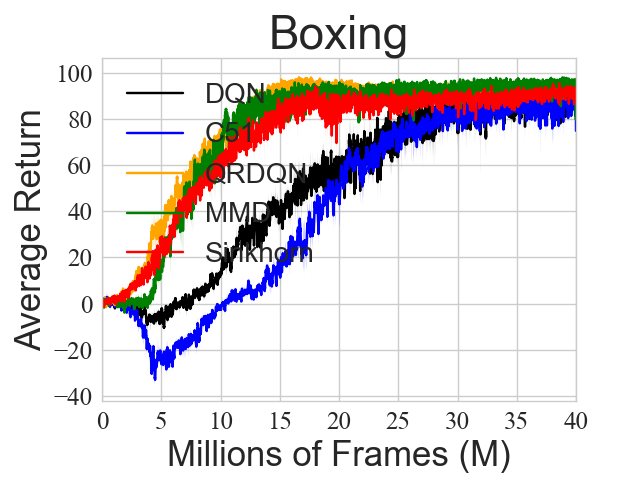}
	\end{subfigure}
	\begin{subfigure}[t]{0.16\textwidth}
		\centering
		\includegraphics[width=\textwidth]{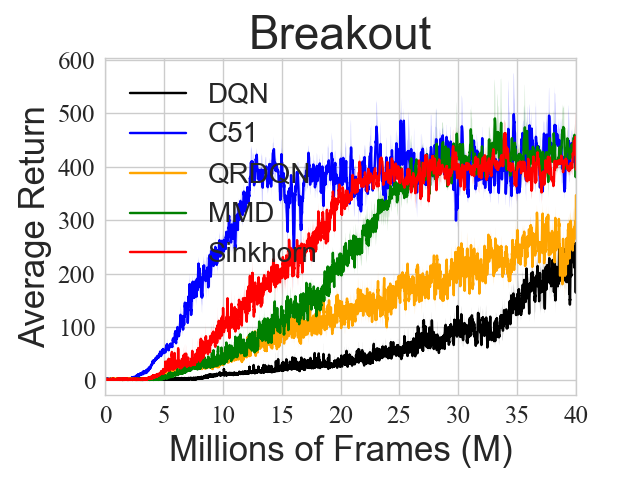}
	\end{subfigure}
	\begin{subfigure}[t]{0.16\textwidth}
		\centering
		\includegraphics[width=\textwidth]{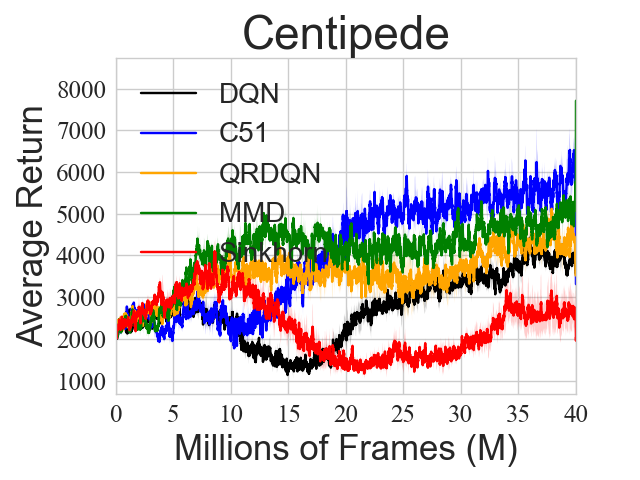}
	\end{subfigure}
	\begin{subfigure}[t]{0.16\textwidth}
		\centering
		\includegraphics[width=\textwidth]{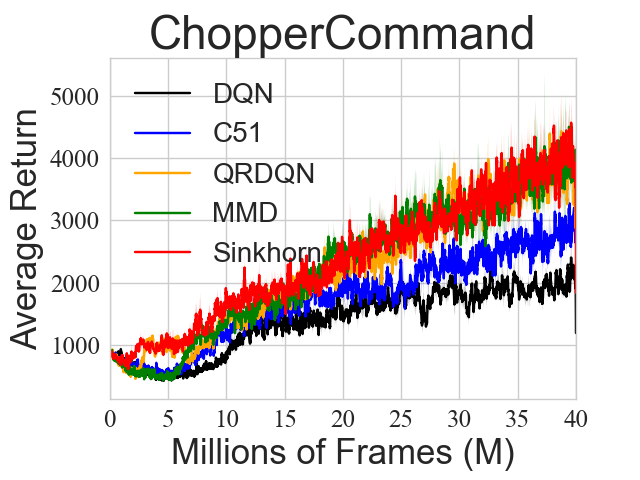}
	\end{subfigure}
	\begin{subfigure}[t]{0.16\textwidth}
		\centering
		\includegraphics[width=\textwidth]{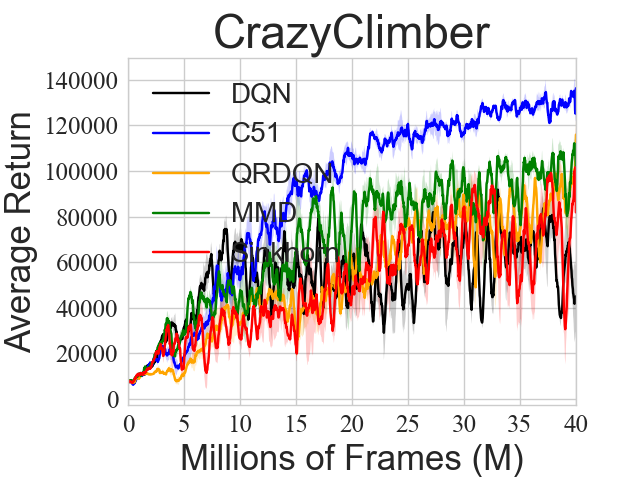}
	\end{subfigure}
	\begin{subfigure}[t]{0.16\textwidth}
		\centering
		\includegraphics[width=\textwidth]{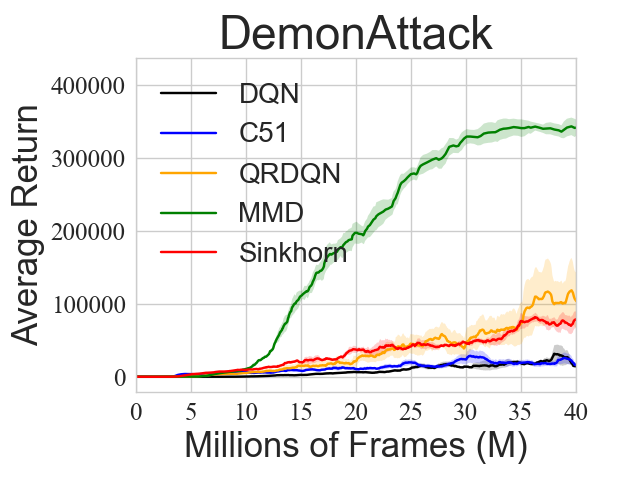}
	\end{subfigure}
	\begin{subfigure}[t]{0.16\textwidth}
		\centering
		\includegraphics[width=\textwidth]{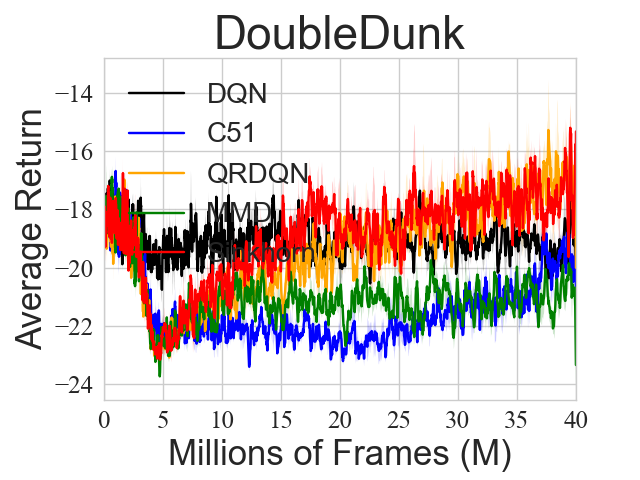}
	\end{subfigure}
	\begin{subfigure}[t]{0.16\textwidth}
		\centering
		\includegraphics[width=\textwidth]{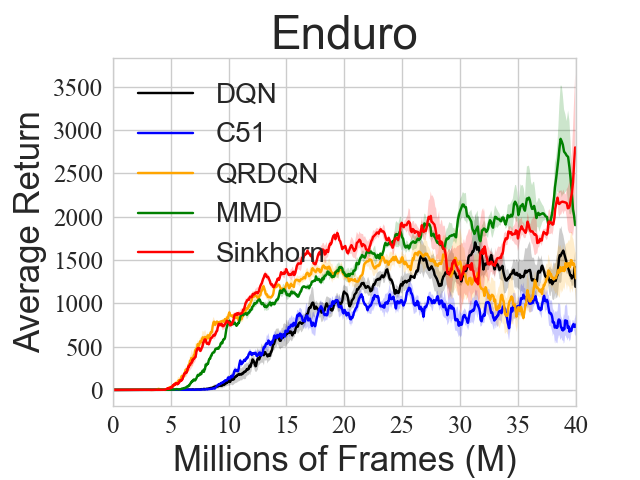}
	\end{subfigure}
	\begin{subfigure}[t]{0.16\textwidth}
		\centering
		\includegraphics[width=\textwidth]{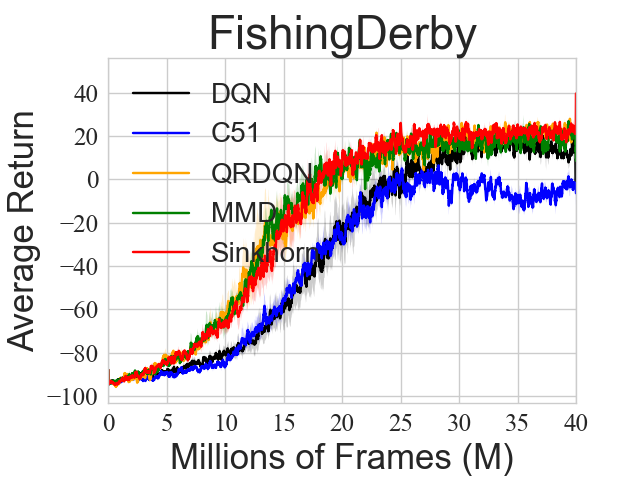}
	\end{subfigure}
	\begin{subfigure}[t]{0.16\textwidth}
		\centering
		\includegraphics[width=\textwidth]{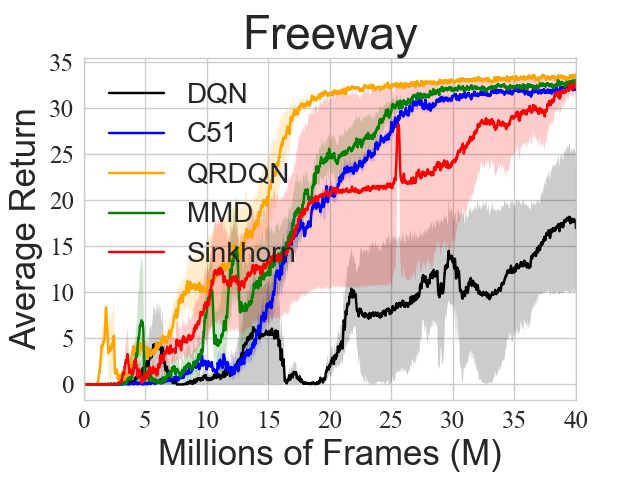}
	\end{subfigure}
	\begin{subfigure}[t]{0.16\textwidth}
		\centering
		\includegraphics[width=\textwidth]{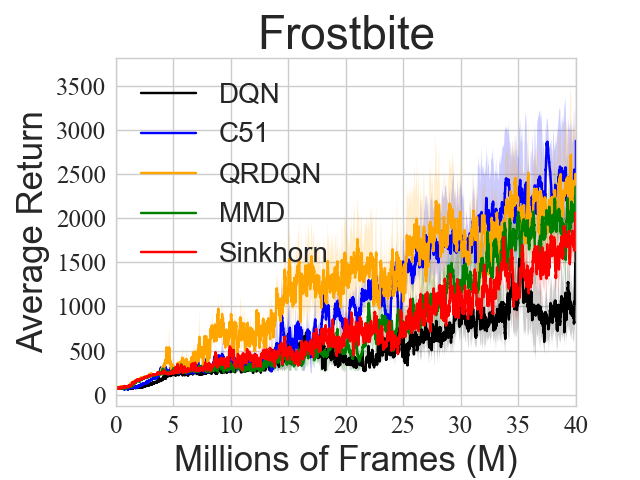}
	\end{subfigure}
	\begin{subfigure}[t]{0.16\textwidth}
		\centering
		\includegraphics[width=\textwidth]{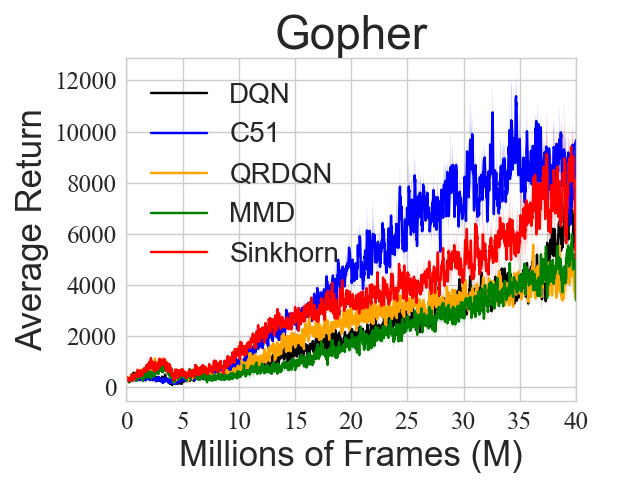}
	\end{subfigure}
	\begin{subfigure}[t]{0.16\textwidth}
		\centering
		\includegraphics[width=\textwidth]{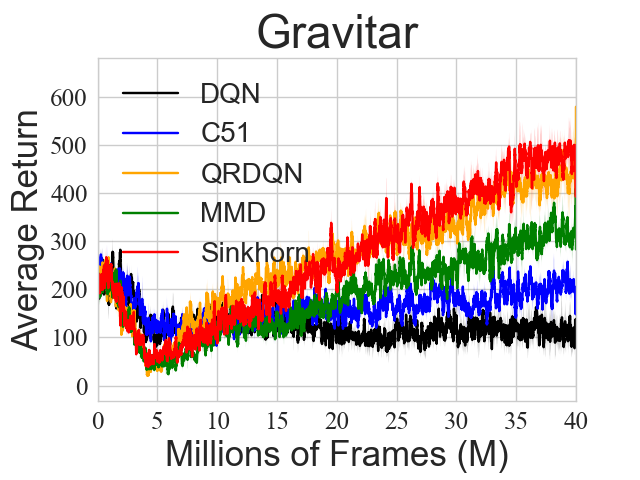}
	\end{subfigure}
	\begin{subfigure}[t]{0.16\textwidth}
		\centering
		\includegraphics[width=\textwidth]{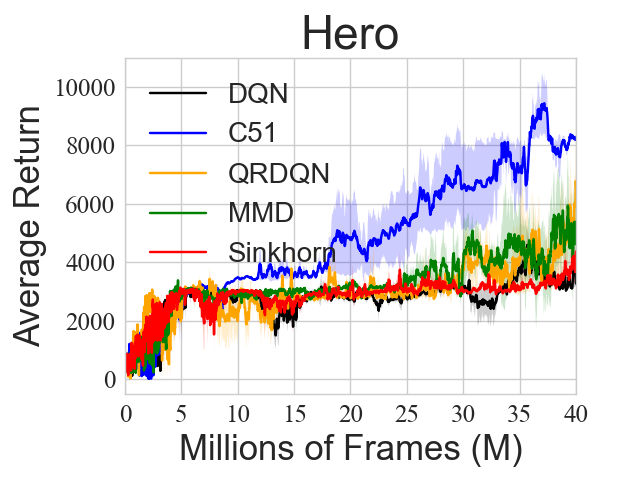}
	\end{subfigure}
	\begin{subfigure}[t]{0.16\textwidth}
		\centering
		\includegraphics[width=\textwidth]{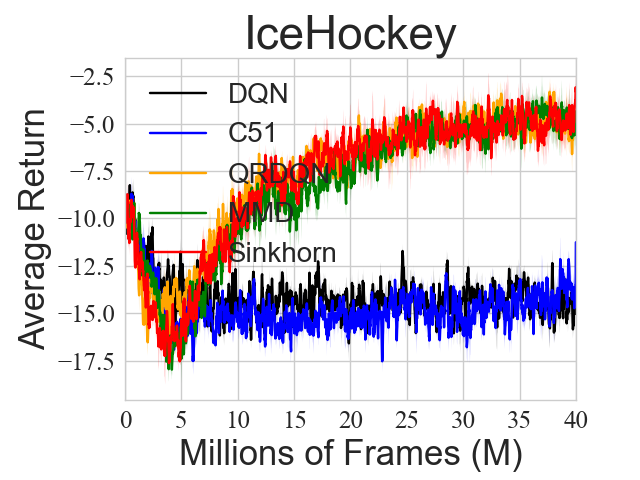}
	\end{subfigure}
	\begin{subfigure}[t]{0.16\textwidth}
		\centering
		\includegraphics[width=\textwidth]{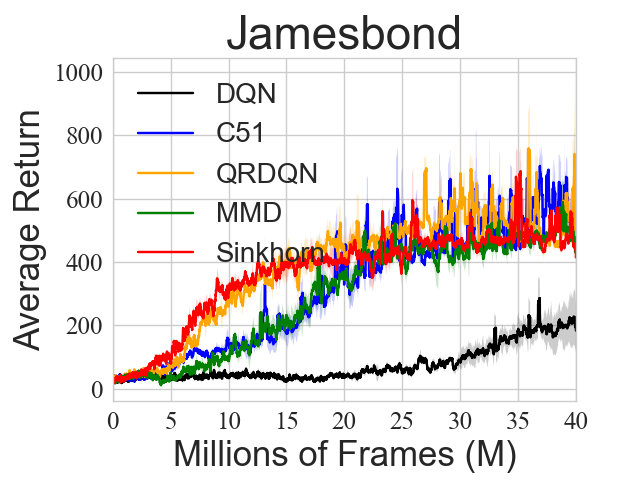}
	\end{subfigure}
	\begin{subfigure}[t]{0.16\textwidth}
		\centering
		\includegraphics[width=\textwidth]{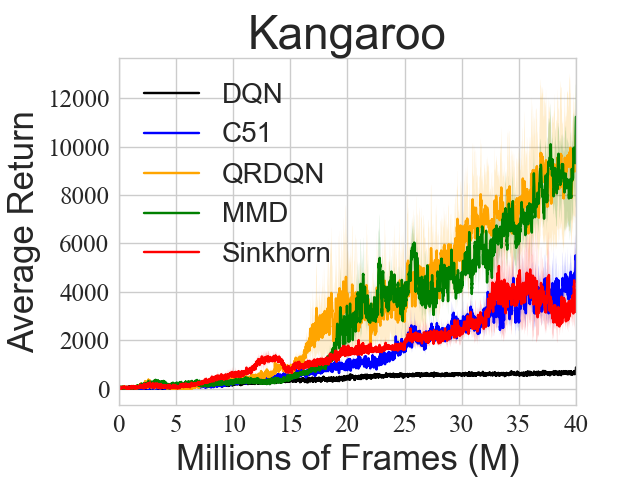}
	\end{subfigure}
	\begin{subfigure}[t]{0.16\textwidth}
		\centering
		\includegraphics[width=\textwidth]{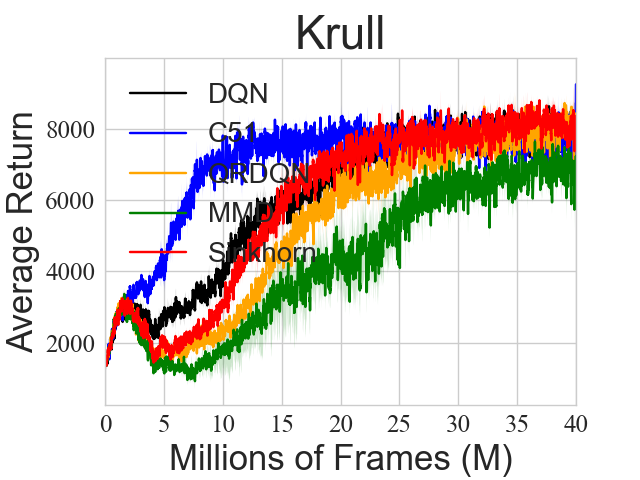}
	\end{subfigure}
	\begin{subfigure}[t]{0.16\textwidth}
		\centering
		\includegraphics[width=\textwidth]{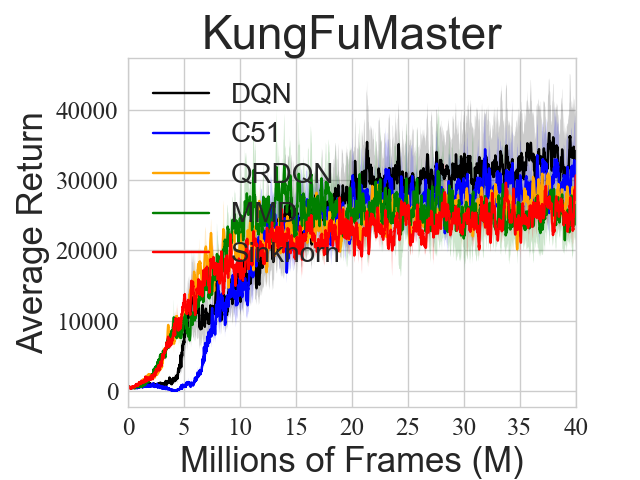}
	\end{subfigure}
	\begin{subfigure}[t]{0.16\textwidth}
		\centering
		\includegraphics[width=\textwidth]{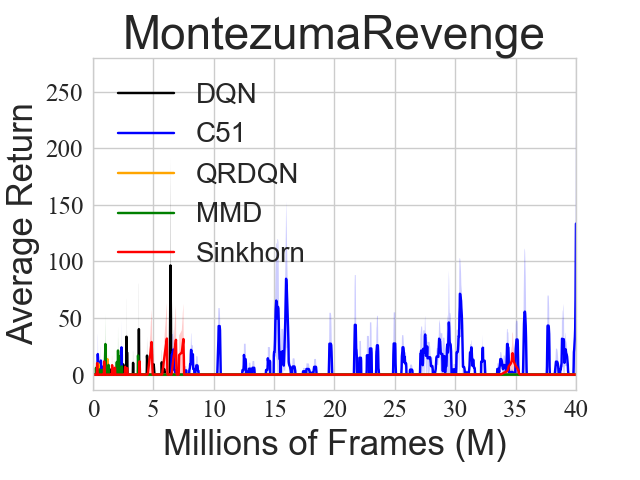}
	\end{subfigure}
	\begin{subfigure}[t]{0.16\textwidth}
		\centering
		\includegraphics[width=\textwidth]{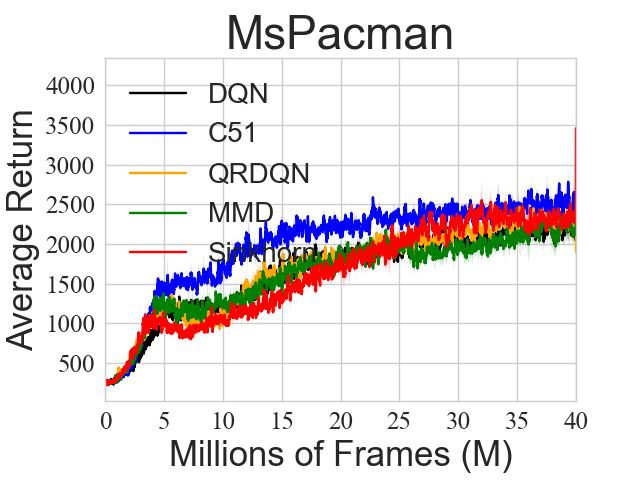}
	\end{subfigure}
	\begin{subfigure}[t]{0.16\textwidth}
		\centering
		\includegraphics[width=\textwidth]{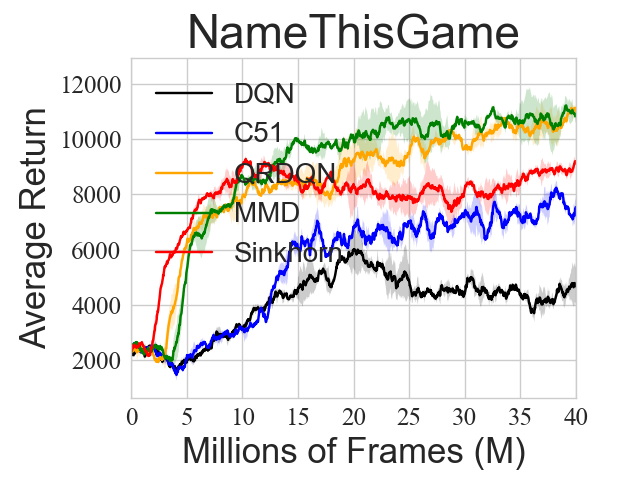}
	\end{subfigure}
	\begin{subfigure}[t]{0.16\textwidth}
		\centering
		\includegraphics[width=\textwidth]{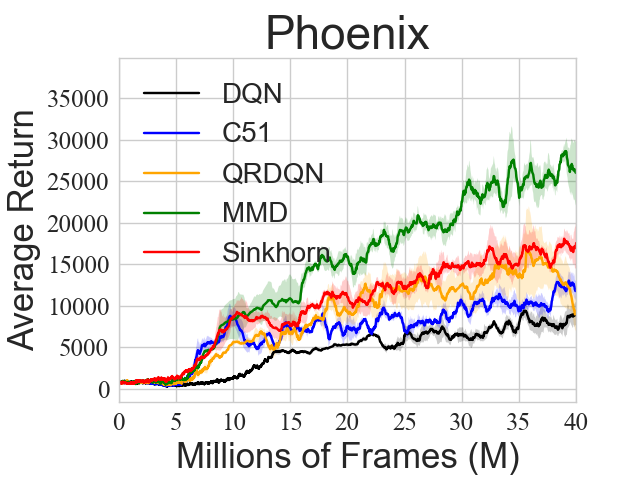}
	\end{subfigure}
	\begin{subfigure}[t]{0.16\textwidth}
		\centering
		\includegraphics[width=\textwidth]{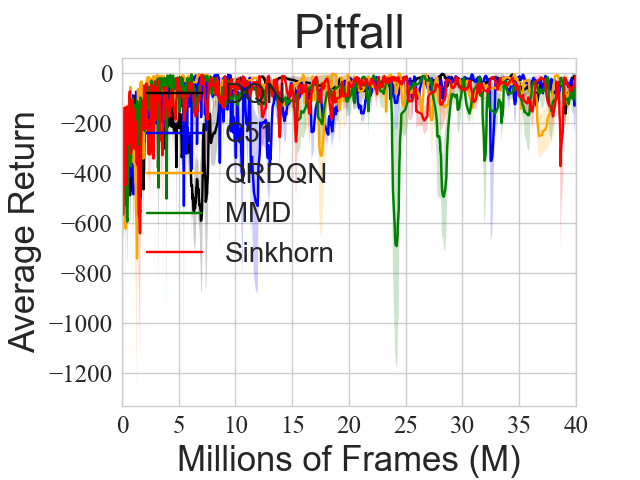}
	\end{subfigure}
	\begin{subfigure}[t]{0.16\textwidth}
		\centering
		\includegraphics[width=\textwidth]{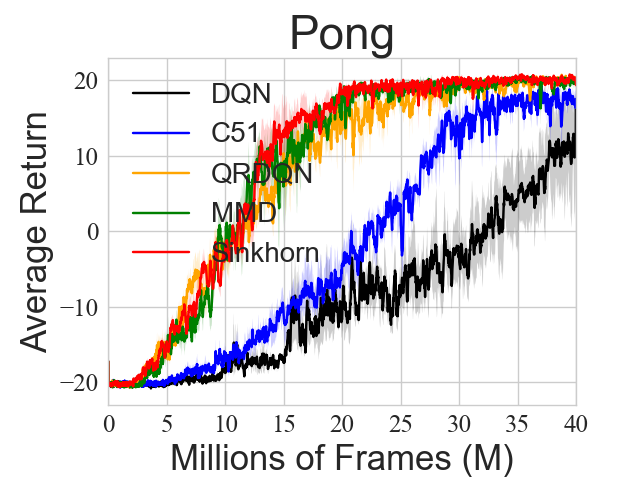}
	\end{subfigure}
	\begin{subfigure}[t]{0.16\textwidth}
		\centering
		\includegraphics[width=\textwidth]{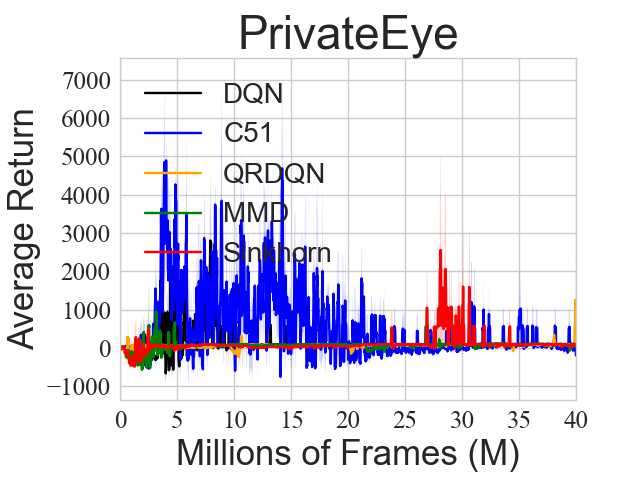}
	\end{subfigure}
	\begin{subfigure}[t]{0.16\textwidth}
		\centering
		\includegraphics[width=\textwidth]{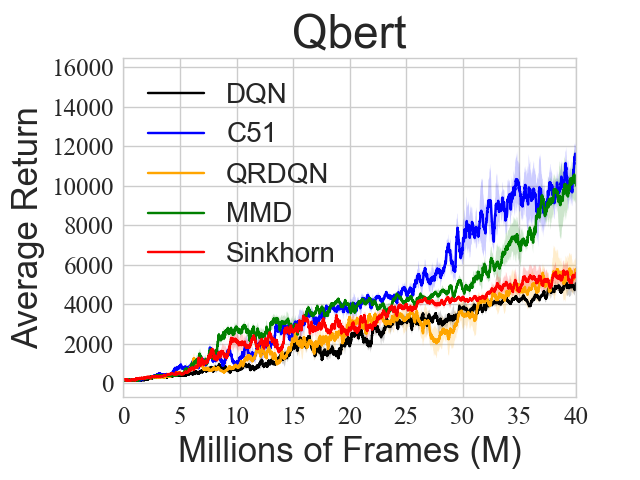}
	\end{subfigure}
	\begin{subfigure}[t]{0.16\textwidth}
		\centering
		\includegraphics[width=\textwidth]{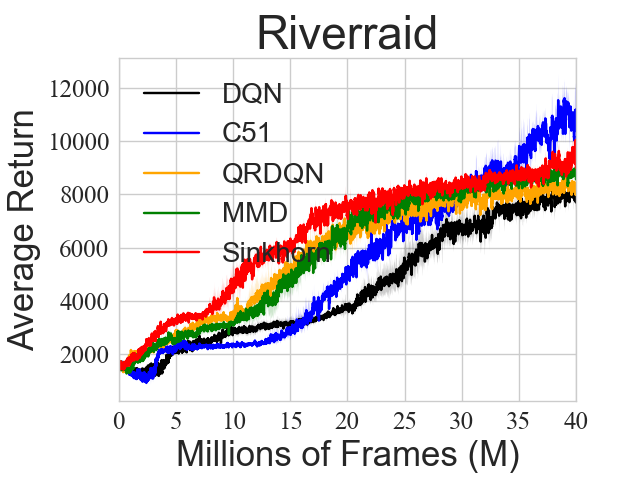}
	\end{subfigure}
	\begin{subfigure}[t]{0.16\textwidth}
		\centering
		\includegraphics[width=\textwidth]{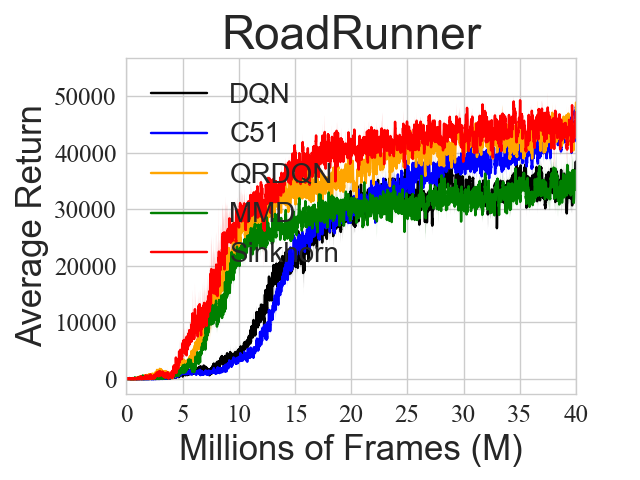}
	\end{subfigure}
	\begin{subfigure}[t]{0.16\textwidth}
		\centering
		\includegraphics[width=\textwidth]{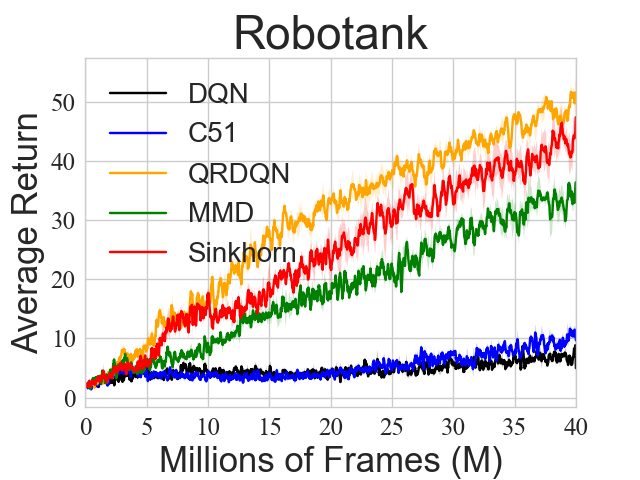}
	\end{subfigure}
	\begin{subfigure}[t]{0.16\textwidth}
		\centering
		\includegraphics[width=\textwidth]{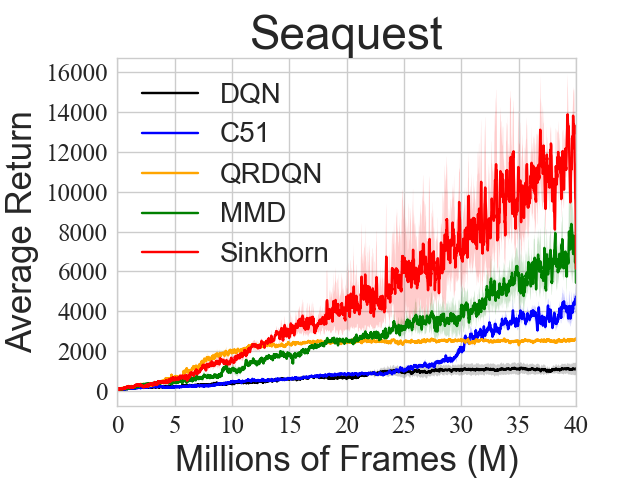}
	\end{subfigure}
	\begin{subfigure}[t]{0.16\textwidth}
		\centering
		\includegraphics[width=\textwidth]{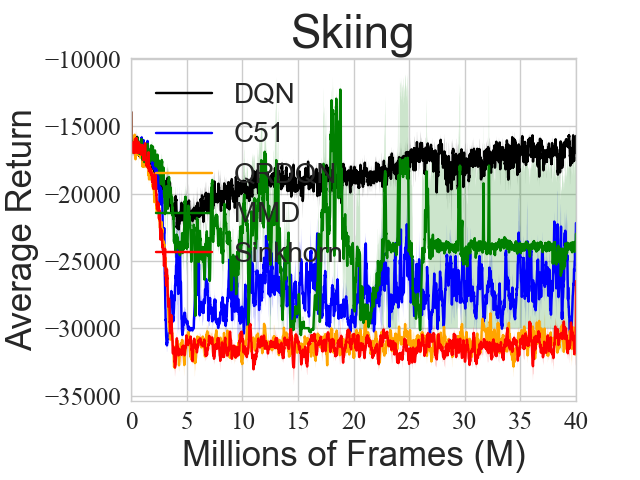}
	\end{subfigure}
	\begin{subfigure}[t]{0.16\textwidth}
		\centering
		\includegraphics[width=\textwidth]{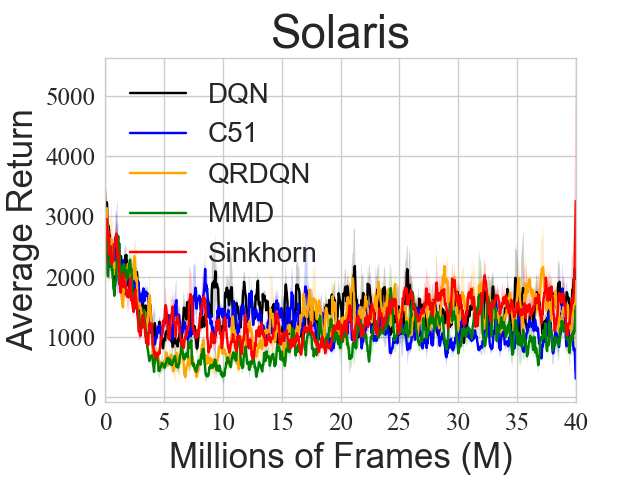}
	\end{subfigure}
	\begin{subfigure}[t]{0.16\textwidth}
		\centering
		\includegraphics[width=\textwidth]{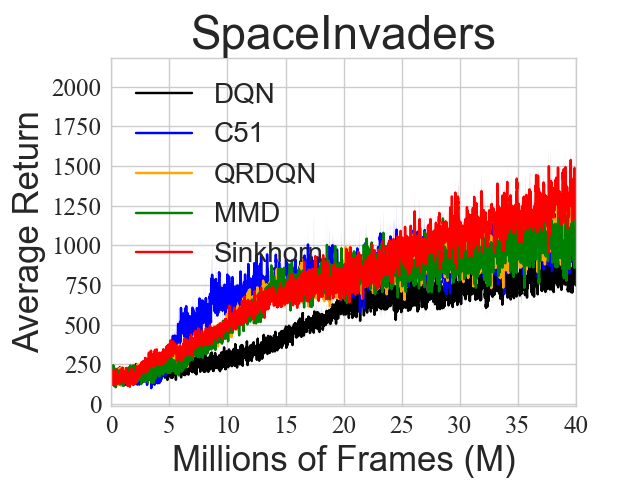}
	\end{subfigure}
	\begin{subfigure}[t]{0.16\textwidth}
		\centering
		\includegraphics[width=\textwidth]{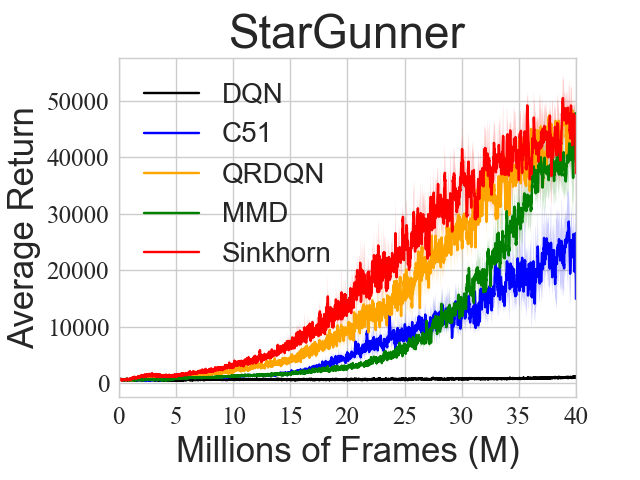}
	\end{subfigure}
	\begin{subfigure}[t]{0.16\textwidth}
		\centering
		\includegraphics[width=\textwidth]{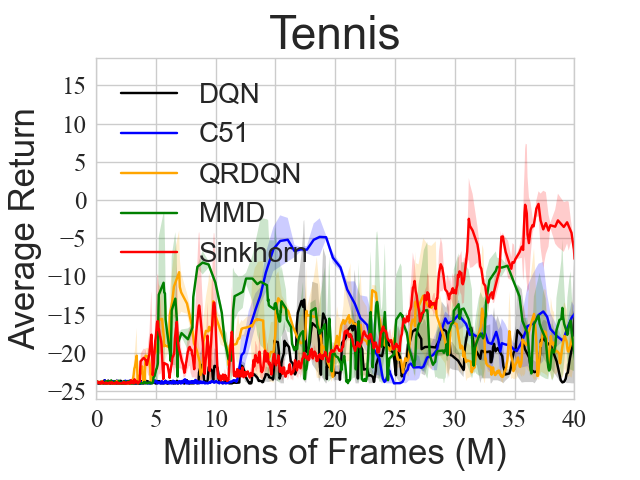}
	\end{subfigure}
	\begin{subfigure}[t]{0.16\textwidth}
		\centering
		\includegraphics[width=\textwidth]{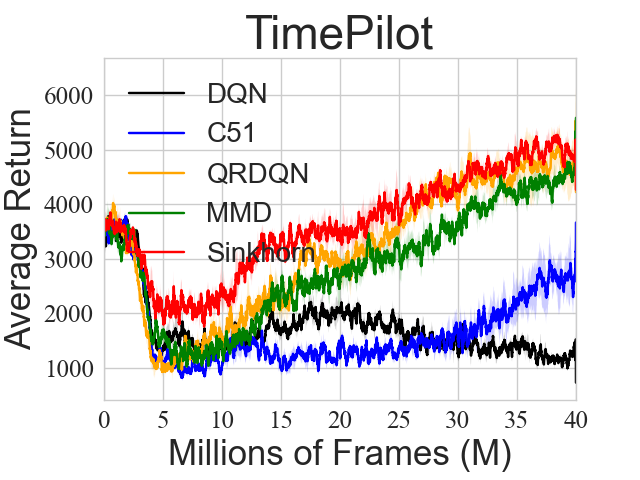}
	\end{subfigure}
	\begin{subfigure}[t]{0.16\textwidth}
		\centering
		\includegraphics[width=\textwidth]{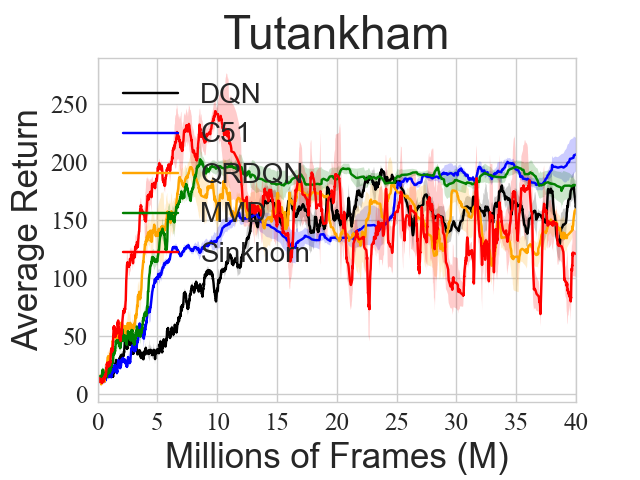}
	\end{subfigure}
	\begin{subfigure}[t]{0.16\textwidth}
		\centering
		\includegraphics[width=\textwidth]{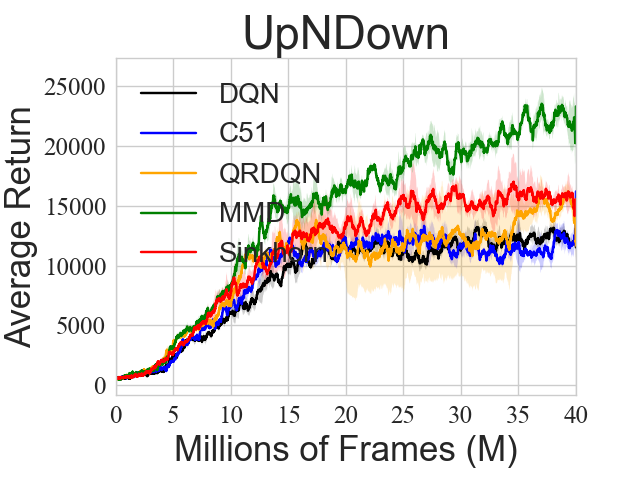}
	\end{subfigure}
	\begin{subfigure}[t]{0.16\textwidth}
		\centering
		\includegraphics[width=\textwidth]{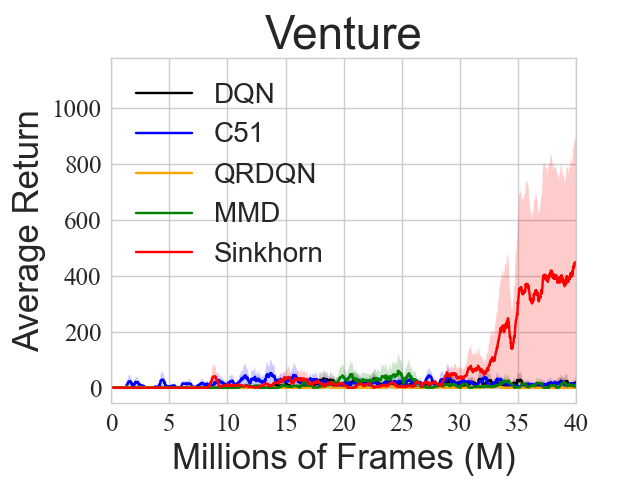}
	\end{subfigure}
	\begin{subfigure}[t]{0.16\textwidth}
		\centering
		\includegraphics[width=\textwidth]{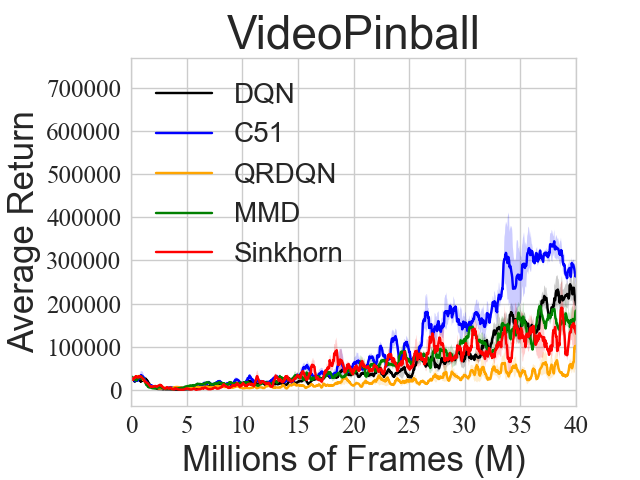}
	\end{subfigure}
	\begin{subfigure}[t]{0.16\textwidth}
		\centering
		\includegraphics[width=\textwidth]{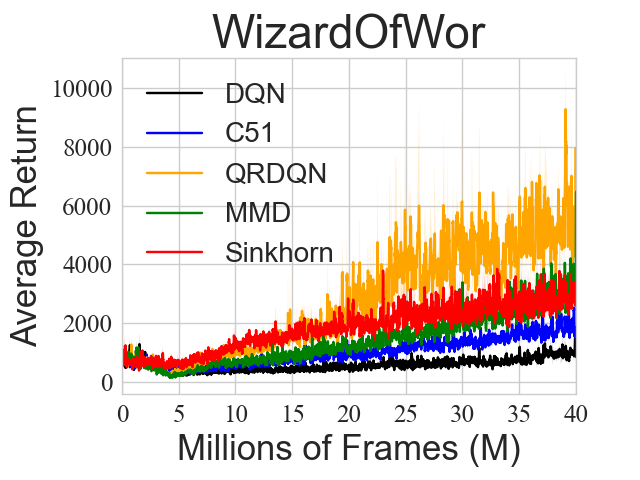}
	\end{subfigure}
	\begin{subfigure}[t]{0.16\textwidth}
		\centering
		\includegraphics[width=\textwidth]{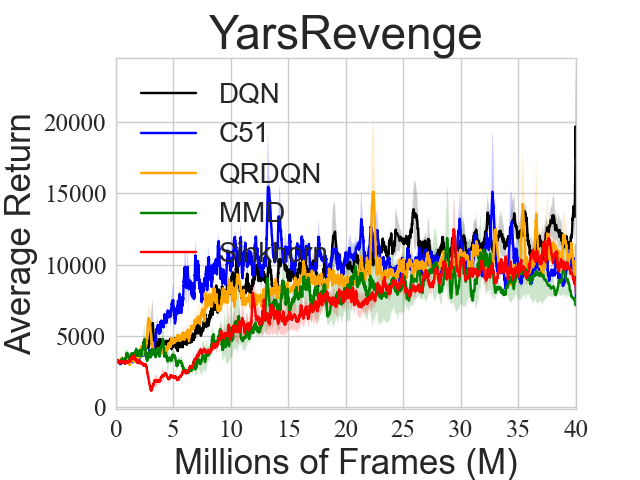}
	\end{subfigure}
	\begin{subfigure}[t]{0.16\textwidth}
		\centering
		\includegraphics[width=\textwidth]{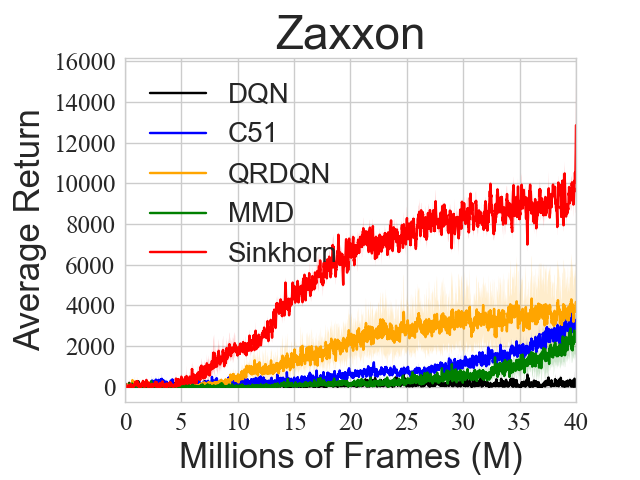}
	\end{subfigure}
	\caption{Learning curves of SinkhornDRL on 55 Atari games after training 40M frames averaged over 3 seeds.}
	\label{fig:allgames}
\end{figure}

\clearpage
%
%
%
%
%

\section{Raw Score Table Across 55 Atari Games}\label{appendix:allresults}

\begin{table*}[htbp]
	\centering
	\scalebox{0.8}{
			\begin{tabular}{l|rr|rrrrr}
					\toprule[1pt]
					\textbf{ GAMES } & \textbf{RANDOM} & \textbf {HUMAN} & \textbf{DQN}  & \textbf{C51} & \textbf{QR-DQN}  & \textbf {MMD-DQN} & \textbf{SinkhornDRL}  \\		
					\hline \textbf { Alien} & 211.9 & 7,127.7 & 1030 & \underline{1510} & 1030  & 1480  & \textbf{1560} \\
					\textbf { Amidar} 	   & 2.34   & 1,719.5 & 341 & 424 & \bf{ 677} & 510 & \underline{588} \\
					\textbf { Assault} & 283.5 & 742.0 & 3232 & \underline{3647} & \bf{12943} &  3295 & 2960 \\
					\textbf { Asterix} & 268.5 & 8,503.3 & 3000 & \bf{34900} & 11500 & \underline{14900}  & 6500 \\
					\textbf { Asteroids} & 1008.6 & 47,388.7 & 1180 & 780 & \bf{1650} & 1080 & \underline{1370} \\
					\textbf { Atlantis} & 22188 & 29,028.1 & 15500 & 84900 & \underline{3316700} &  93600  & \bf{3447100} \\
					\textbf { BankHeist} & 14 & 753.1 & 570 & \underline{960} & \bf{980} & 880 & 700 \\
					\textbf { BattleZone} & 3000 & 37,187.5 & 15000 & 19000 & 26000 & \bf{35000} & \underline{32000} \\
					\textbf { BeamRider} & 414.3 & 16,926.5 & \underline{8200} & 7476 & 7642 &  \bf{25602} & 6022 \\
					\textbf { Berzerk} & 165.6 & 2,630.4 & \bf{970} & 650 & 640 &  860 &  \underline{910} \\
					\textbf { Bowling} & 23.48 & 160.7 &54 & 43 & \bf{60} & \bf{60} &  \bf{60} \\
					\textbf { Boxing} & -0.69 & 12.1 & 94 & 90 & \bf{100} &  \bf{100} & \bf{100} \\
					\textbf { Breakout} & 1.5 & 30.5 &343 & \bf{452} & 414 &  \underline{432} &  418 \\
					\textbf { Centipede} & 2064.77 & 12,017.0 & \underline{7551} & 4133 & 5388 & \bf{9342} &  4070 \\
					\textbf { ChopperCommand} & 794 & 7,387.8 & 1500 & \bf{3600} & 3500 & \bf{3600} & 3400 \\
					\textbf{ CrazyClimber} & 8043 & 35,829.4 &94300 & \bf{153100} & \underline{139500} & 98500 & 137400 \\
					\textbf{ DemonAttack} & 162.25 & 1,971.0 & 31420 & 50240 & \underline{240660}  & \bf{407030} & 105185 \\
					\textbf { DoubleDunk} & -18.14 & -16.4 &\underline{-16} & -20 & -18 &  -22 &  \bf{-12} \\
					\textbf { Enduro} & 0.01 & 860.5 & 1387 & 1086 & \underline{1972} & 1953 &  \bf{4608} \\
					\textbf { FishingDerby} & -93.06 & -38.7& 23 & -1 & \underline{31} & \underline{31} & \bf{61} \\
					\textbf { Freeway} & 0.01 & 29.6 & 31 & 32 & \bf{34} & 33 & \bf{34} \\
					\textbf { Frostbite} & 73.2 & 4,334.7 &3330 & \bf{3690} & \underline{3470} & 3250  & 2640 \\
					\textbf { Gopher} & 364 & 2,412.5 &\underline{11400} & \bf{14780} & 5440 & 3740 &  6620 \\
					\textbf { Gravitar} & 226.5 & 3,351.4 &350 & 350 & \bf{750} & 350  & \underline{500} \\
					\textbf { Hero } & 551 & 30,826.4 &3440 & \underline{8535} & \bf{10155} &  7195 & 6540 \\
					\textbf { IceHockey} & -10.3 & 0.9 &-13 & -10 & -4 &  \underline{-3} & \bf{-2} \\
					\textbf { Jamesbond} & 27 & 302.8 & 350 & \underline{600} & \bf{650} &  450 &  500 \\
					\textbf { Kangaroo} & 54 & 3,035.0 & 1300 & 6500 & \underline{14600} & \bf{14800} & 3600 \\
					\textbf { Krull} & 1,566.59 & 2,665.5 & 8892 & 9336 & \bf{10053} & 7762 &  \underline{9630} \\
					\textbf { KungFuMaster} & 451 & 22,736.3 & \bf{46500} & 38000 & 27900 & 26900 &  \underline{43600} \\
					\textbf { MontezumaRevenge} & 0.0 & 4,753.3 & 1 & \bf{400} & 1 & 1 & 1 \\
					\textbf { MsPacman} & 242.6 & 6,951.6 &\underline{3230} & 2440 & 1860 & 3130  & \bf{5120} \\
					\textbf { NameThisGame} & 2404.9 & 8,049.0 & 6160 & 5750 & \bf{13580}  & 9350 & \underline{11250} \\
					\textbf { Phoenix} & 757.2 & 7,242.6 & 9430 & 18780 & 9390 & \bf{25690} & \underline{23300} \\
					\textbf { Pitfall} & -265 & 6,463.7 & 1 & 1 & 1  & 1  & 1 \\
					\textbf { Pong } & -20.34 & 14.6 & \bf{21} & 20 & 20 & \bf{21} & \bf{21} \\
					\textbf { PrivateEye} & 34.49 & 69,571.3 & 100 & 100 & 100 &  100  & 100 \\
					\textbf { Qbert } & 188.75 & 13,455.0 &  7425 & \bf{16375} & 7800 & \underline{16225} &  7750 \\
					\textbf { RiverRaid} & 1575.4 & 17,118.0 & 8470 & \bf{13310} & 8710 &  9190 &  \underline{9530} \\
					\textbf { RoadRunner} & 7 & 7,845.0 & 45500 & \bf{60900} & 52500 &  45600 &  \underline{59500} \\
					\textbf { Robotank} & 2.24 & 11.9 & 8 & 11 & \bf{58} &  39 & \underline{54} \\
					\textbf { Seaquest} & 88.2 & 42,054.7 & 1740 & 5940 & 2640 &  \underline{7370} &  \bf{8350} \\
					\textbf { Skiing} & -16267.9 & -4,336.9 &\underline{-13681} & -20495 & -29970  & \bf{-8986} &  -23455 \\
					\textbf { Solaris} & 2346.6 & 12,326.7 & 1640 & 660 & 2200 &  \underline{3380} & \bf{7720} \\
					\textbf { SpaceInvaders} & 136.15 & 1,668.7 & 940 & \bf{2480} & 1170 & 770 &  \underline{1200} \\
					\textbf{ StarGunner} & 631 & 10,250.0 & 1200 & 17200 & \underline{52900} &  52500 &  \bf{57500} \\
					\textbf{ Tennis} & -23.92 & -8.3 & -23 & \underline{-1} & -7 &  -8 &  \bf{5} \\
					\textbf { TimePilot} & 3682 & 5,229.2 &800 & 4100 & 4400 &\bf{8000} &  \underline{4500} \\
					\textbf { Tutankham} & 15.56 & 167.6 &  201 & \underline{213} & \bf{220} & 141 &  137 \\
					\textbf { UpNDown} & 604.7 & 11,693.2 & 14560 & 18440 & 13710 &  \bf{27370} & \underline{18910} \\
					\textbf { Venture} & 0.0 & 1,187.5 & 1 & 1 & 1 &  1 &  \bf{700} \\
					\textbf { VideoPinball} & 15720.98 & 17,667.9 & 155165 & \bf{576843} & 189460 & 69175 & \underline{347700} \\
					\textbf { WizardOfWor} & 534 & 4,756.5 & 1400 & 2400 & \bf{14300} & \underline{11500} &  4300 \\
					\textbf { YarsRevenge} & 3271.42 & 54,576.9 & \bf{28048} & 7882 & \underline{17729} &  7520 &  9120 \\
					\textbf{ Zaxxon} & 8 & 9,173.3 &1 & 3900 & \underline{9100} & 4300 &  \bf{19500} \\
					\hline
					{\color{blue} \textbf{Number of Best}} &   &  &4 & 12 & 15 & 13 &  \bf{17}\\
					{\color{blue}\textbf{Number of Second Best} }&   &  &6 & 7 & 10 & 8 &  \bf{16}\\
					\hline
					\bottomrule[1pt]
				\end{tabular}
		} 	
	\caption{Best score of all algorithms over 3 seeds across 55 Atari games after training 40M Frames. \textbf{Bold} denotes the best performance, while the \underline{underline} represents the second best performance. The number of games with the best and second best performance substantiate the superiority of our SinkhornDRL across all considered baseline algorithms.}
	\label{table:allresults}
\end{table*}

\clearpage

\section{Features of Atari Games}\label{appendix:featuresgames}

\begin{table}[htbp]
	\centering
	\scalebox{0.85}{
		\begin{tabular}{c|c|c}
			\toprule[1pt]
			\hline
			\textbf{ GAMES } & \textbf{Action Space} & \textbf {Dynamics} \\
			\hline
			\textbf{Alien} & 18 & Complex  \\
			\textbf{Amidar} & 6 & Simple  \\ 
			\textbf{Assault} & 7 & Complex  \\ 
			\textbf{Asterix} & 18 & Complex  \\	
			\textbf{Asteroids} & 4 & Simple  \\	 
			\textbf{Atlantis} & 4 & Simple  \\	 
			\textbf{BankHeist} & 18 & Simple  \\ 
			\textbf{BattleZone} & 18 & Simple  \\ 
			\textbf{BeamRider} & 18 & Complex  \\
			\textbf{Berzerk} & 18 & Complex  \\
			\textbf{Bowling} & Continuous & Simple  \\	 
			\textbf{Boxing} & 6 & Simple  \\
			\textbf{Breakout} & 4 & Simple  \\ 
			\textbf{Centipede} & 18 & Complex  \\ 
			\textbf{ChopperCommand} & Continuous & Complex  \\
			\textbf{CrazyClimber} & 18 & Complex  \\
			\textbf{DemonAttack} & 18 & Complex  \\	
			\textbf{DoubleDunk} & 18 & Simple  \\
			\textbf{Enduro} & 9 & Simple \\
			\textbf{FishingDerby} & 18 & Simple \\ 
			\textbf{Freeway} & 3 & Simple \\
			\textbf{Frostbite} & 18 & Complex \\
			\textbf{Gopher} & 18 & Simple \\
			\textbf{Gravitar} & Continuous & Complex \\
			\textbf{Hero} & 18 & Simple \\
			\textbf{IceHockey} & Continuous & Simple \\
			\textbf{Jamesbond} & 18 & Complex \\
			\textbf{Kangaroo} & 18 & Complex \\
			\textbf{Krull} & 18 & Complex \\
			\textbf{KungFuMaster} & 18 & Complex \\
			\textbf{MontezumaRevenge} & 18 & Complex \\ 
			\textbf{MsPacman} & 9 & Simple \\
			\textbf{NameThisGame} & 18 & Complex \\
			\textbf{Phoenix} & 18 & Complex \\
			\textbf{Pitfall} & 18 & Complex \\
			\textbf{Pong} & 3 & Simple \\
			\textbf{PrivateEye} & 18 & Complex \\
			\textbf{Qbert} & 6 & Complex \\
			\textbf{Riverraid} & 18 & Complex  \\
			\textbf{RoadRunner} & 18 & Simple  \\
			\textbf{Robotank} & 9 & Simple  \\
			\textbf{Seaquest} & 18 & Complex  \\
			\textbf{Skiing} & 9 & Simple  \\
			\textbf{Solaris} & 18 & Complex  \\
			\textbf{SpaceInvaders} & 6 & Simple  \\
			\textbf{StarGunner} & 18 & Complex  \\
			\textbf{Tennis} & 18 & Simple  \\
			\textbf{TimePilot} & 18 & Complex  \\
			\textbf{Tutankham} & 18 & Complex  \\
			\textbf{UpNDown} & 18 & Complex  \\
			\textbf{Venture} & 18 & Complex  \\
			\textbf{VideoPinball} & 6 & Simple  \\
			\textbf{WizardOfWor} & 12 & Complex  \\
			\textbf{YarsRevenge} & 18 & Complex  \\
			\textbf{Zaxxon} & 18 & Complex  \\
			\hline
			\bottomrule[1pt]
		\end{tabular}
	} 	
	\caption{Number of Action space and difficulty of environmental dynamics of 55 Atari games.}
	\label{table:games}
\end{table}

\clearpage

\section{Sensitivity Analysis and Computational Cost}\label{appendix:sensitivity_cost}

\subsection{More results in Sensitivity Analysis}\label{appendix:sensitivity}

\begin{figure*}[htbp]
	\centering
	\begin{subfigure}[t]{0.24\textwidth}
		\centering
		\includegraphics[width=\textwidth,trim=0 0 0 10,clip]{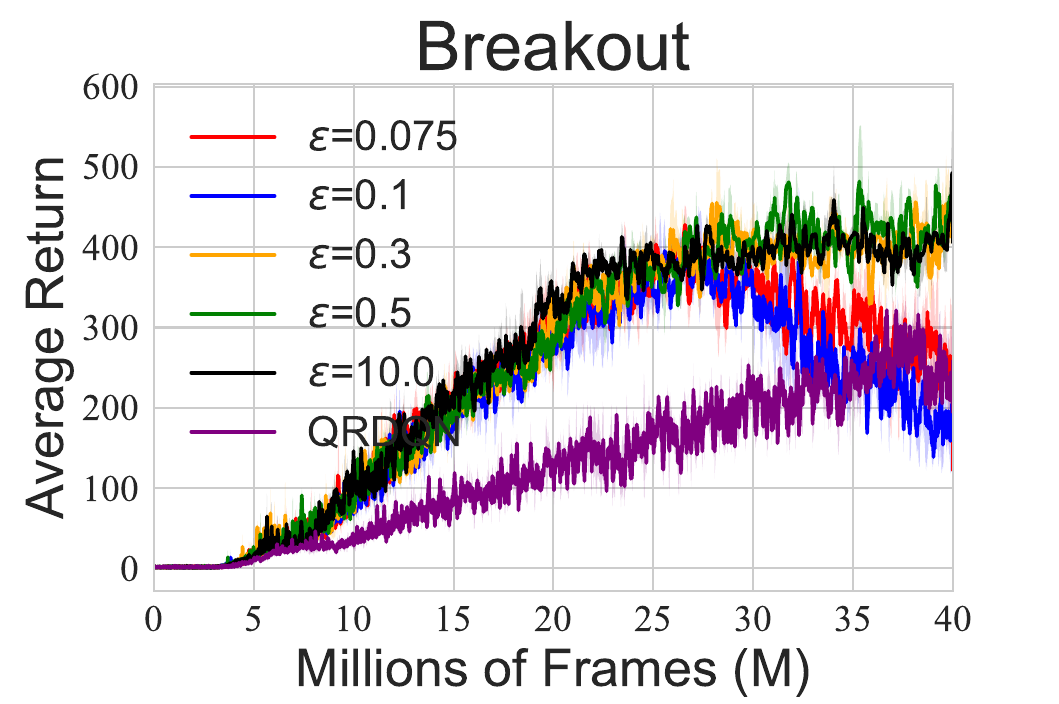}
		\caption{Small $\varepsilon$ vs QR-DQN}
	\end{subfigure}
	\begin{subfigure}[t]{0.24\textwidth}
		\centering
		\includegraphics[width=\textwidth,trim=0 0 0 10,clip]{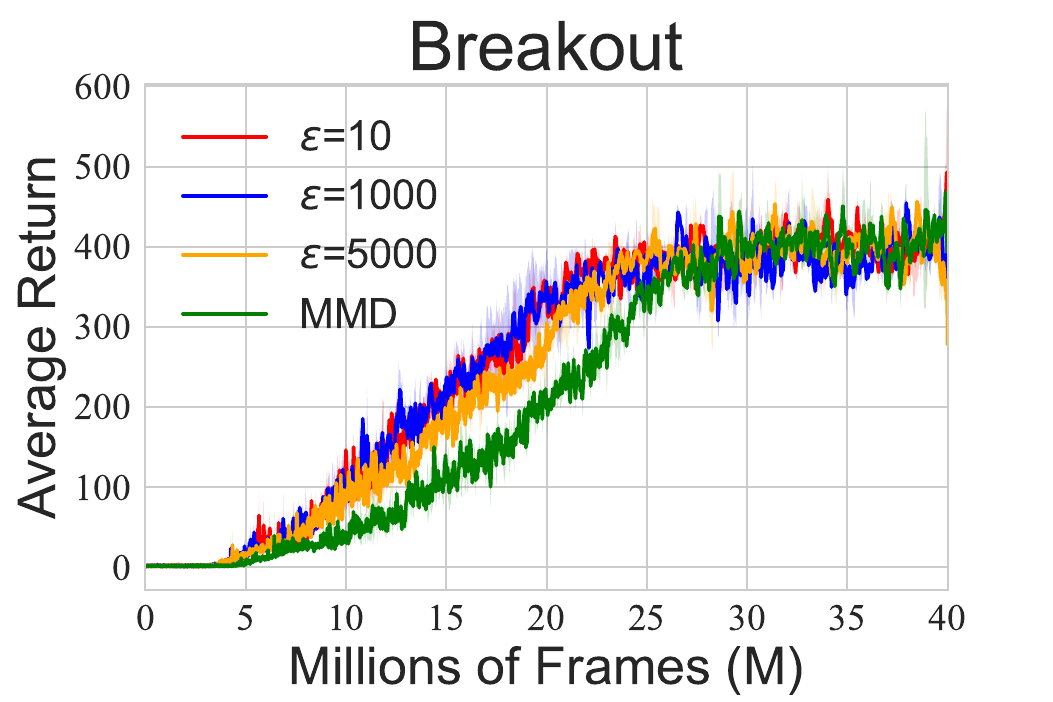}
		\caption{Large $\varepsilon$ vs MMD-DQN}
	\end{subfigure}
	\begin{subfigure}[t]{0.24\textwidth}
		\centering
		\includegraphics[width=\textwidth,trim=0 0 0 0,clip]{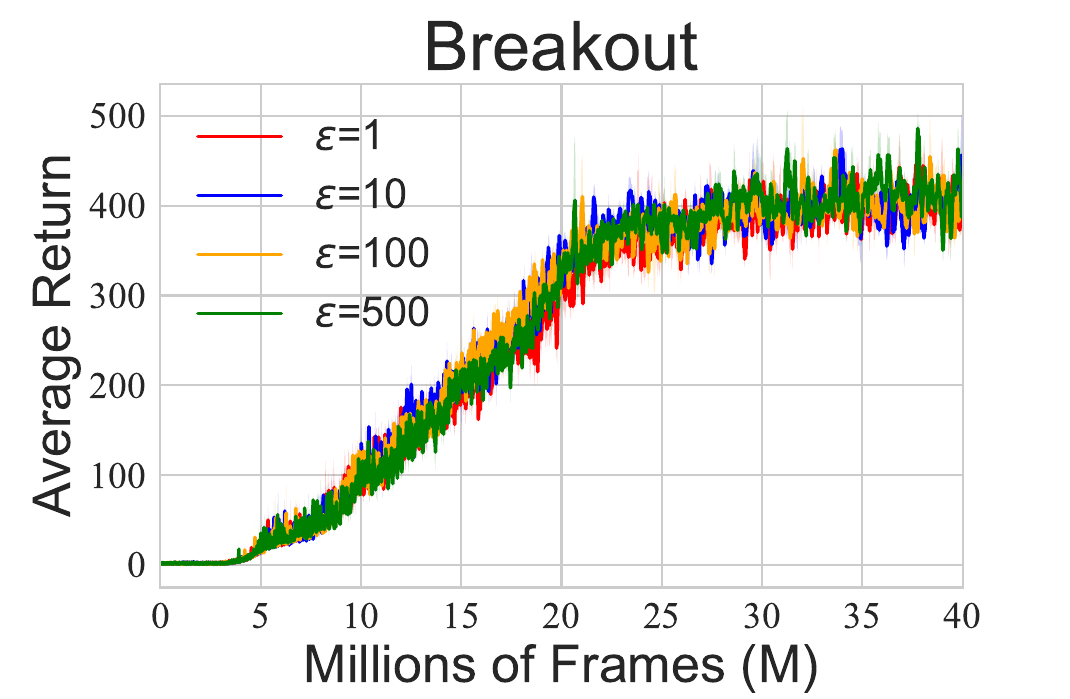}
		\caption{$\varepsilon$ on Breakout}
	\end{subfigure}
	\begin{subfigure}[t]{0.24\textwidth}
		\centering
		\includegraphics[width=\textwidth,trim=0 0 0 0,clip]{figures/0Figure_seaquest_epsilon_QRDQN.pdf}
		\caption{$\varepsilon$ on Seaquest}
	\end{subfigure}
	\caption{(a) Sensitivity analysis w.r.t. a small level of $\varepsilon$ SinkhornDRL to compare with QR-DQN that approximates Wasserstein distance on Breakout. (b) Sensitivity analysis w.r.t. a large level of $\varepsilon$ SinkhornDRL algorithm to compare with MMD-DQN on Breakout. All learning curves are reported over 2 seeds. (c) and (d) are results for a general $\varepsilon$ on Breakout and Seaquest, respectively.}
	\label{fig:sensivity_more}
\end{figure*}

\paragraph{Decreasing $\varepsilon$.} We argue that the limit behavior connection as stated in Theorem~\ref{theorem:sinkhorn} may not be able to be verified rigorously via numeral experiments due to the numerical instability of Sinkhorn Iteration in Algorithm~\ref{alg:sinkhorn_iterations}. From Figure~\ref{fig:sensivity_more}~(a), we can observe that if we gradually decline $\varepsilon$ to 0, SinkhornDRL's performance tends to degrade and approach QR-DQN. Note that an overly small $\varepsilon$ will lead to a trivial almost 0 $\mathcal{K}_{i, j}$ in Sinkhorn iteration in Algorithm~\ref{alg:sinkhorn_iterations}, and will cause $\frac{1}{0}$ numerical instability issue for $a_l$ and $b_l$ in Line 5 of  Algorithm~\ref{alg:sinkhorn_iterations}. In addition, we also conducted experiments on Seaquest, a similar result is also observed in Figure~\ref{fig:sensivity_more}~(d). As shown in Figure~\ref{fig:sensivity_more}~(d), the performance of SinkhornDRL is robust when $\varepsilon =10, 100, 500$, but a small $\epsilon=1$ tends to worsen the performance.

\paragraph{Increasing $\varepsilon$.} Moreover, for breakout, if we increase $\varepsilon$, the performance of SinkhornDRL tends to degrade and be close to MMD-DQN as suggested in Figure~\ref{fig:sensivity_more}~(b). It is also noted that an overly large $\varepsilon$ will let the $\mathcal{K}_{i, j}$ explode to $\infty$. This also leads to the numerical instability issue in Sinkhorn iteration in Algorithm~\ref{alg:sinkhorn_iterations}.

\begin{figure*}[htbp]
	\centering
	\begin{subfigure}[t]{0.25\textwidth}
		\centering
		\includegraphics[width=\textwidth,trim=0 0 0 10,clip]{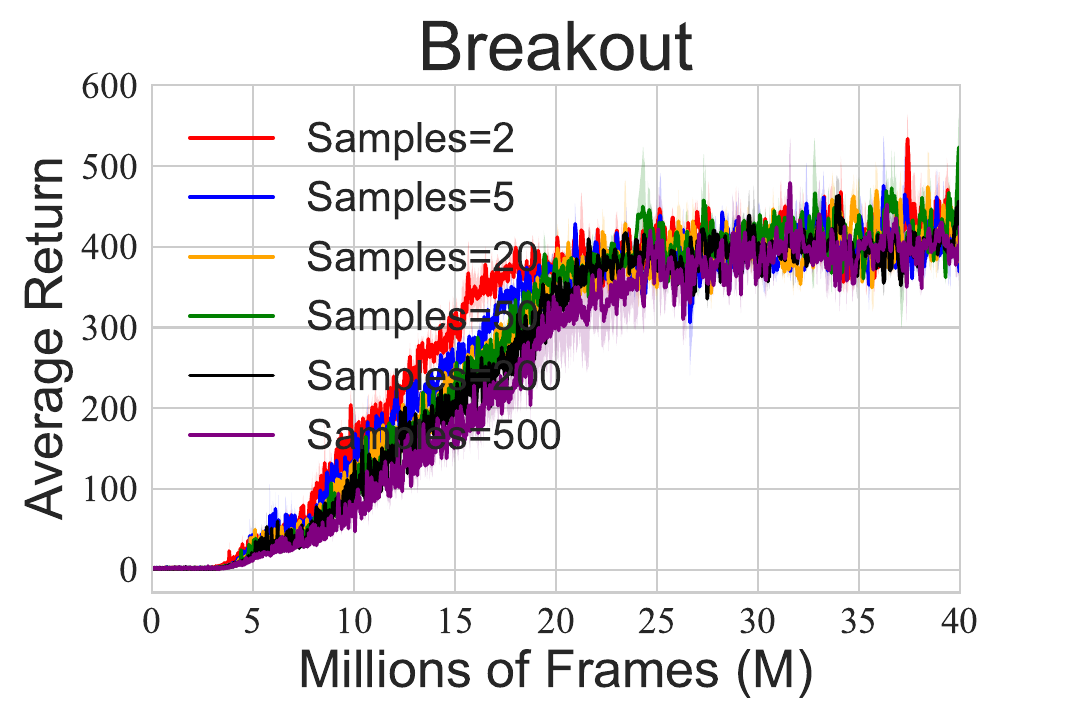}
		\caption{Samples on Breakout}
	\end{subfigure}
	\begin{subfigure}[t]{0.25\textwidth}
		\centering
		\includegraphics[width=\textwidth,trim=0 0 0 10,clip]{figures/0Figure_seaquest_samples_more.pdf}
		\caption{Samples on Seaquest}
	\end{subfigure}
	\caption{Sensitivity analysis of Sinkhorn in terms of the number of samples $N$ on Breakout (a) and Seaquest (b).}
	\label{fig:sensivity_more_samples}
\end{figure*}

\paragraph{Samples $N$.} We find that SinkhornDRL requires a proper number of samples $N$ to perform favorably, and the sensitivity w.r.t $N$ depends on the environment. As suggested in Figure~\ref{fig:sensivity_more_samples} (a), a smaller $N$, e.g., $N=2$ on breakout has already achieved favorable performance and even accelerates the convergence in the early phase, while $N=2$ on Seaquest will lead to the divergence issue. Meanwhile, an overly large $N$ worsens the performance across two games. We conjecture that using larger network networks to generate more samples may suffer from the overfitting issue, yielding the training instability~\cite{bjorck2021towards}. In practice, we choose a proper number of samples, i.e., $N=200$ across all games.

\begin{figure}[htbp]
	\centering
	\begin{subfigure}[t]{0.25\textwidth}
		\centering
		\includegraphics[width=\textwidth,trim=0 0 0 10,clip]{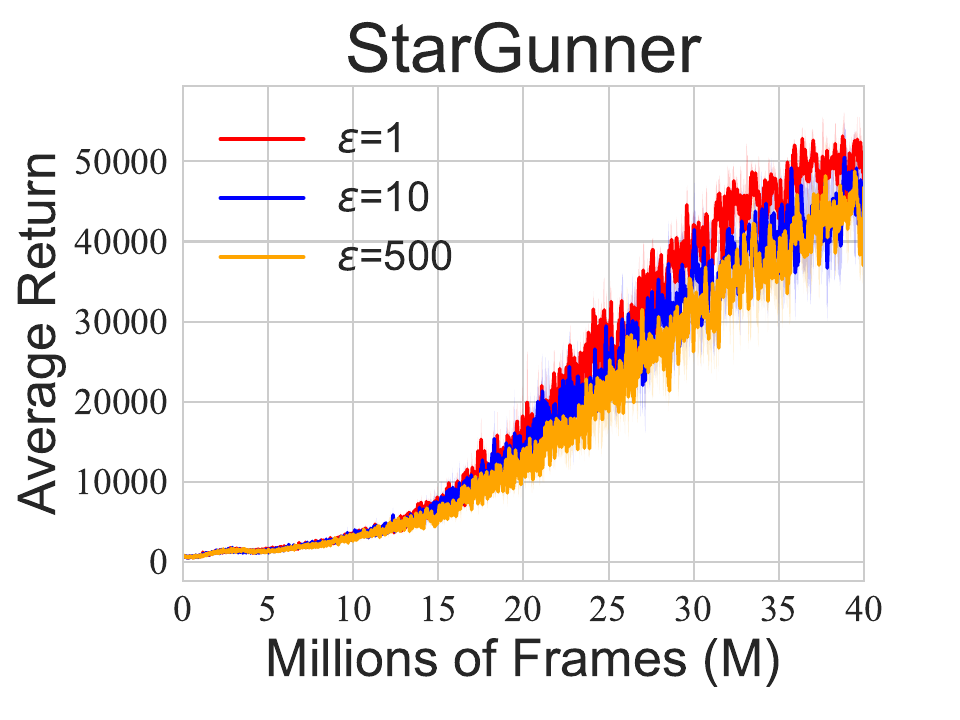}
		\caption{StarGunner: $\varepsilon$}
	\end{subfigure}
	\begin{subfigure}[t]{0.25\textwidth}
		\centering
		\includegraphics[width=\textwidth,trim=0 0 0 10,clip]{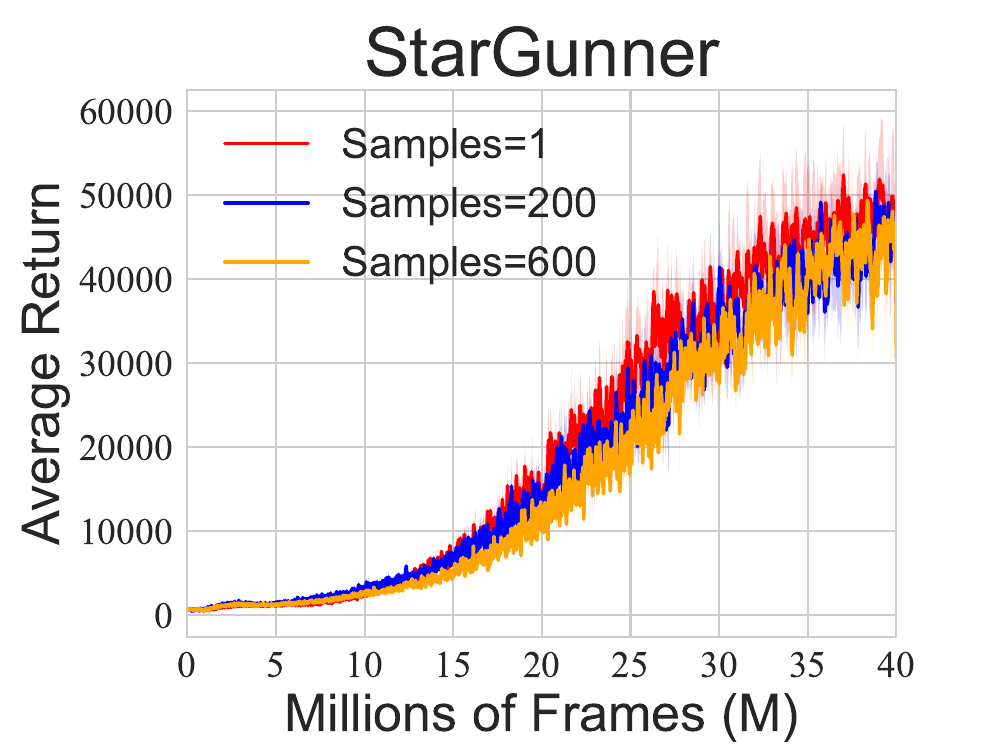}
		\caption{StarGunner: Samples}
	\end{subfigure}
	\begin{subfigure}[t]{0.23\textwidth}
		\centering
		\includegraphics[width=\textwidth,trim=0 0 0 10,clip]{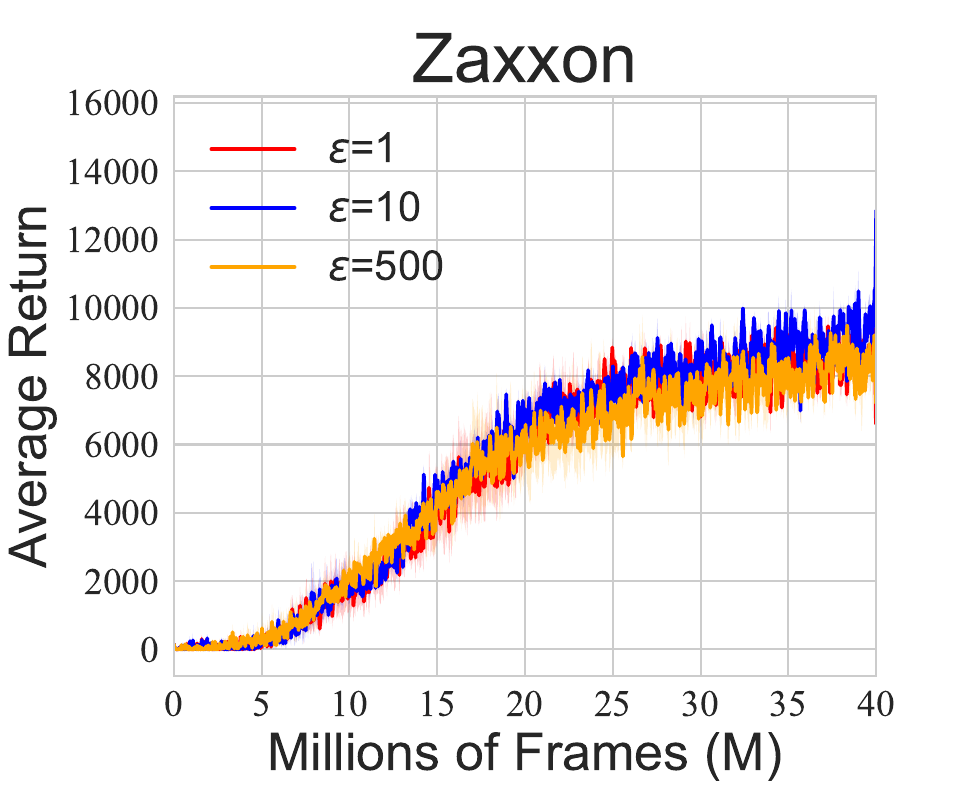}
		\caption{Zaxxon: $\varepsilon$}
	\end{subfigure}
	\begin{subfigure}[t]{0.25\textwidth}
		\centering
		\includegraphics[width=\textwidth,trim=0 0 0 10,clip]{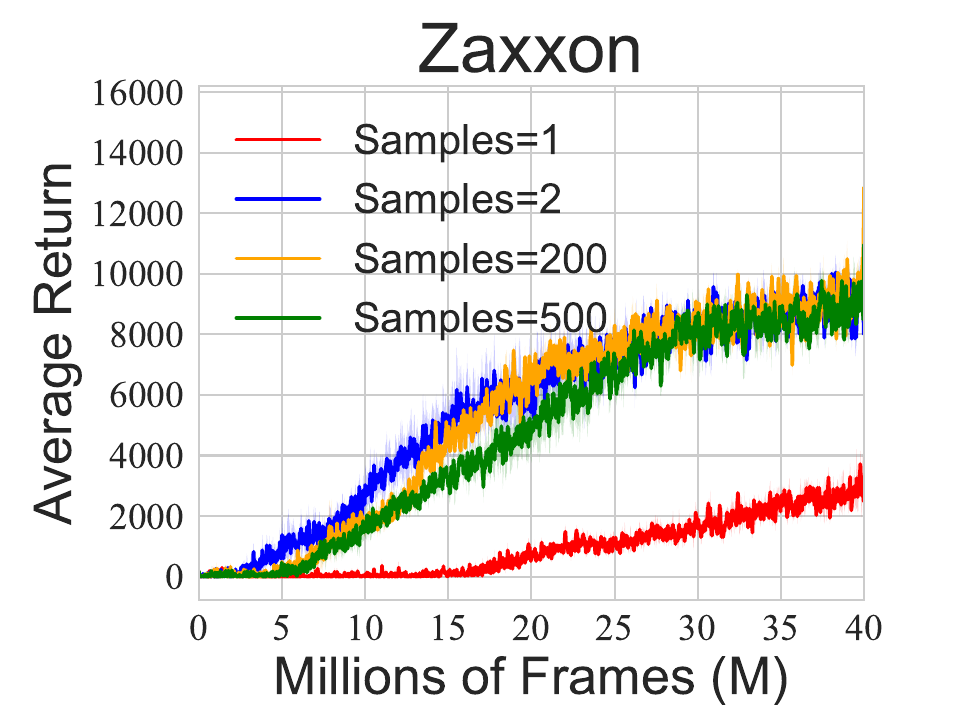}
		\caption{Zaxxon: Samples}
	\end{subfigure}
	\caption{Sensitivity analysis of SinkhornDRL on StarGunner and Zaxxon in terms of $\varepsilon$, and number of samples. Learning curves are reported over 3 seeds. }
	\label{fig:sensivity_stargunner_zaxxon}
\end{figure}

\paragraph{More Games on StarGunner and Zaxxon.} Beyond Breakout and Seaquest, we also provide sensitivity analysis on StarGunner and Zaxxon games in Figure~\ref{fig:sensivity_stargunner_zaxxon}. It suggests overly small samples, e.g., 1 and overall large samples tend to degrade the performance, especially on Zaxxon. Although the two games are robust to $\varepsilon$, and we find a small or large $\varepsilon$ hurts the performance in Seaquest. Thus, considering all games, we set samples 200, and $\varepsilon=10.0$ in a moderate range across all games, although a more careful tuning in each game will improve the performance further.

\subsection{Comparison with the Computational Cost}\label{appendix:cost}

\begin{figure}[htbp]
	\centering
	\includegraphics[width=0.6\textwidth,trim=0 0 0 0,clip]{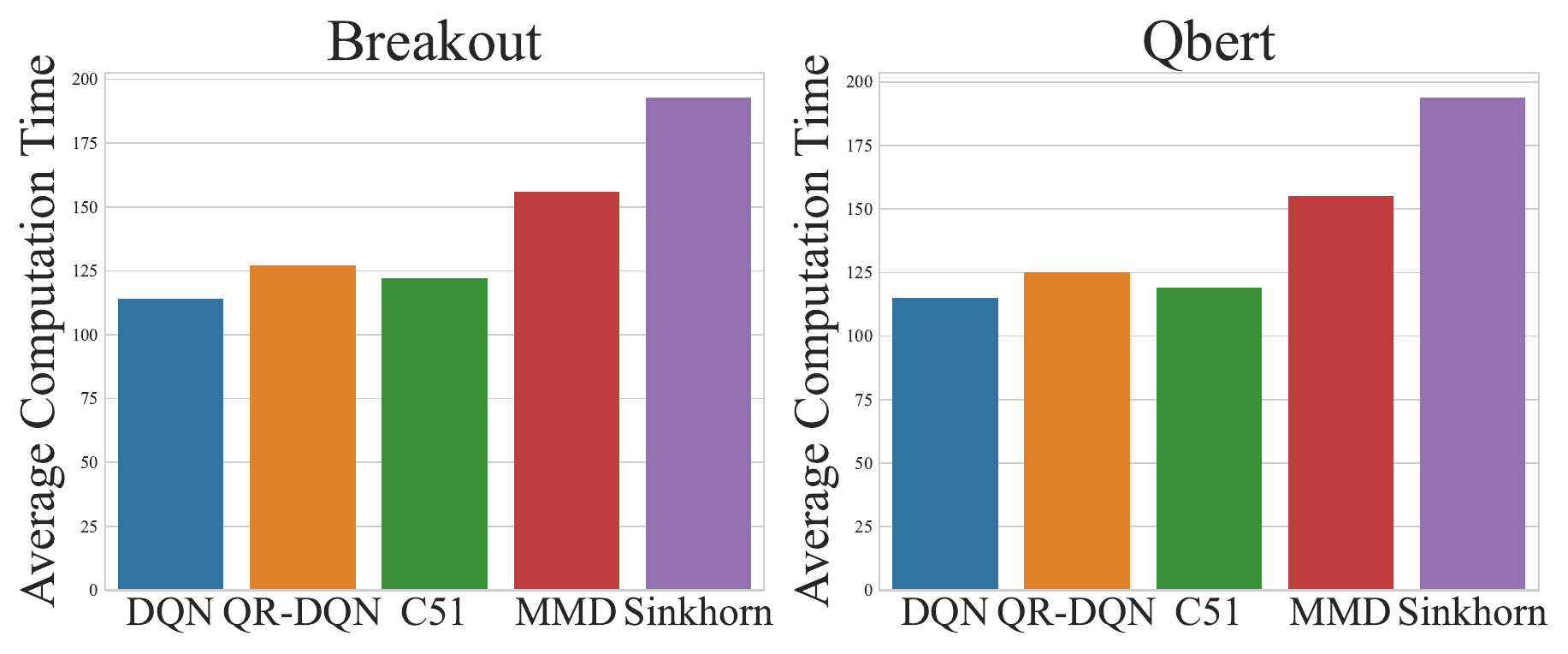}
	\caption{Average computational cost per 10,000 iterations of all considered distributional RL algorithm, where we select $\varepsilon=10$, $L=10$ and the number of samples $N=200$ in SinkhornDRL algorithm.}
	\label{fig:cost}
\end{figure}

We evaluate the computational time every 10,000 iterations across the whole training process of all considered distributional RL algorithms and make a comparison in Figure~\ref{fig:cost}. It suggests that SinkhornDRL indeed increases around 50$\%$ computation cost compared with QR-DQN and C51, but only slightly increases the cost in contrast to MMD-DQN on both Breakout and Qbert games. We argue that this additional computational burden can be tolerant given the significant outperformance of SinkhornDRL in a large number of environments.

\begin{figure}[htbp]
	\centering
	\includegraphics[width=0.6\textwidth,trim=0 0 0 0,clip]{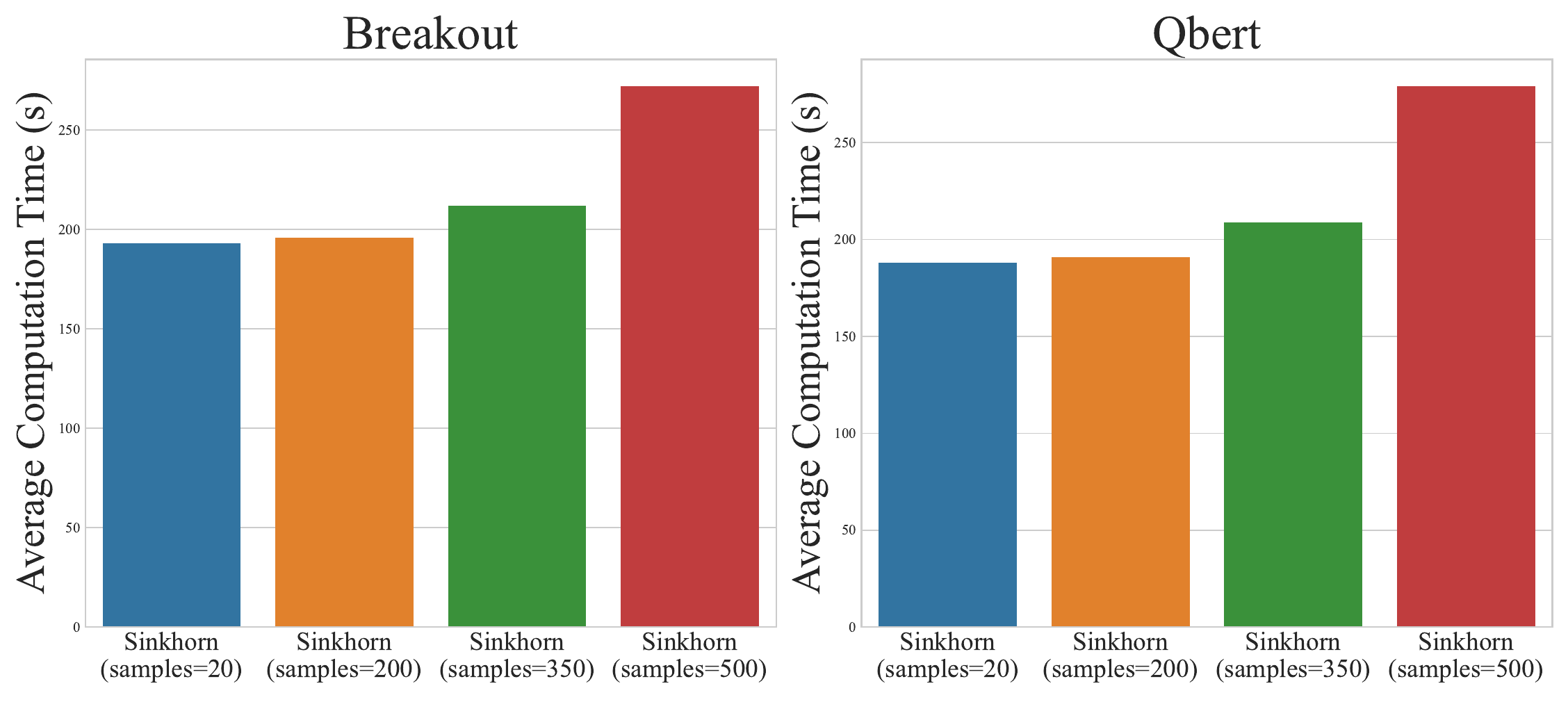}
	\caption{Average computational cost per 10,000 iterations of SinkhornDRL algorithm over different samples.}
	\label{fig:cost2}
\end{figure}

In addition, we also find that the number of Sinkhorn iterations $L$ is negligible to the computation cost, while an overly large sample $N$, e.g., 500, will lead to a large computational burden as illustrated in Figure~\ref{fig:cost2}. This can be intuitively explained as the computation complexity of the cost function $c_{i, j}$ is $\mathcal{O}(N^2)$ in SinkhornDRL, which is particularly heavy in the computation if $N$ is large enough.

\clearpage

\section{Experimental Setting in Multi-dimensional Return Distributions}\label{appendix:multi}

\textbf{Reward Structure and Decomposition.} In practice, the reward function can be multi-dimensional~\cite{van2017hybrid,lin2020rd,lizotte2010efficient,dann2023reinforcement,zhang2021distributional,lin2019distributional}, where distributional RL is aimed at modeling multivariate return distribution with multiple reward sources. We follow the multi-dimensional return distribution setting in \cite{zhang2021distributional}, which construct six Atari games with multiple sources of rewards by decomposing the scalar-valued primitive rewards into multi-dimension. For completeness, we introduce the respective reward structure and the decomposing method of the six considered Atari games, including AirRaid, Asteroids, Gopher, MsPacman, UpNDown, and Pong. The reward is decomposed while keeping the total reward unchanged.

\begin{itemize}
	\item \textbf{AirRaid.} For primitive rewards, the agent kills different kinds of monsters and then receive discrete values of the rewards. The scalar-based primitive rewards are decomposed into four dimensions. The agent will get multi-dimensional rewards [100, 0, 0, 0], [0, 75, 0, 0], [0, 0, 50, 0],[0, 0, 0, 25], [0, 0, 0, 0] respectively for the primitive reward 100, 75, 50, 25 and 0.
	
	\item \textbf{Asteroids.} For primitive rewards, the agent kills different kinds of monsters and then receive values of the rewards. We denote the primitive reward as $r$, and decompose it into the three-dimensional reward as $[r_1, r_2, r_3]$. If $(r - 20) \text{ mod } 50 = 0$, we let $r_1=20$, otherwise $r_1 = 0$. If $(r - r_1 - 50) \text{ mod } 100 = 0$, we let $r_2 = 50$, otherwise $r_2 = 0$. We let $r_3 = r - r_1 - r_2$.
	
	\item \textbf{Gopher.} For primitive rewards, the agent gets $+80$ reward for killing a monster and $+20$ reward after removing holes on the ground. We denote the primitive reward as $r$, and decompose it into the two-dimensions as $[r_1, r_2,]$. If $(r - 20) \text{ mod } 100 = 0$, we let $r_1=20$, otherwise $r_1 = 0$.  We let $r_2 = r - r_1$.
	
	\item \textbf{MsPacman.} For primitive rewards, the agent gets $\{+200, +400, +800, +1,600\}$ rewards after killing different monsters and $+10$ rewards after eating beans. In the reward decomposition, we decompose primitive reward denoted as $r$ into four dimensions $[r_1, r_2, r_3, r_4]$.  If $(r - 10) \text{ mod } 50 = 0$, we let $r_1=10$, otherwise $r_1 = 0$. If $(r - r_1 - 50) \text{ mod } 100 = 0$, we let $r_2 = 50$, otherwise $r_2 = 0$. If $(r - r_1 - r_2 - 100) \text{ mod } 200 = 0$, we let $r_3 = 100$, otherwise $r_3 = 0$. We let $r_4 = r - r_1 - r_2 - r_3$.
	
	\item \textbf{Pong.} For primitive rewards,  the agent gets $+1$ if it wins a round, and $-1$ for losing the round. We decompose the reward into two-dimension: the agent will get $[-1, 0]$ for a $-1$ reward, $[0, 1]$ for a $+1$ reward; otherwise, $[0, 0]$.
	
	\item \textbf{UpNDown.} For primitive rewards, the agent gets $+400$ reward for killing an energy car, $+100$ for reaching a flag, and $+10$ reward for being alive. We denote the primitive reward as $r$, and decompose it into the three-dimensional reward as $[r_1, r_2, r_3]$. If $(r - 10) \text{ mod } 100 = 0$, we let $r_1=10$, otherwise $r_1 = 0$. If $(r - r_1 - 100) \text{ mod } 200 = 0$, we let $r_2 = 100$, otherwise $r_2 = 0$. We let $r_3 = r - r_1 - r_2$.
	
\end{itemize}

\textbf{Detailed Experimental Setup.}  Our implementation extends our code in one-dimensional return setting to multi-dimensional return scenario and adopts the key aspects in  \cite{zhang2021distributional}. For instance, similar to \cite{zhang2021distributional}, we leverage a clipping reward normalizer to clip the multi-dimensional rewards into $[-1, 1]$ after applying the reward decomposition procedure mentioned above to the primitive rewards. We keep the same model architecture except only modifying the output of the last layer from $(B, |\mathcal{A}|, N)$ to $(B, |\mathcal{A}|, D, N)$, where $B$ is the batch size within each batch training, and $D$ is the dimension of the decomposed mutivariate reward function in each game.

\textbf{Baseline Algorithms.} Quantile regression can be used to approximate 1-Wasserstein distance in one-dimensional setting~\cite{dabney2017distributional} as the one-dimensional Wassertein distance has a closed-form expression via the quantile function. However, it remains elusive how to use quantile regression to approximate multi-dimensional Wasserstein distance. This is to say, it is still unclear how to extend the quantile regression distributional RL~(QR-DQN) into multi-dimensional return distribution setting, resulting in no proper baseline in our experiment. Despite that, 
we directly compare SinkrhornDRL with MMD-DQN~\cite{nguyen2020distributional} as MMD is applicable and computationally tractable in the multi-dimensional setting. Notably, we did not introduce other baselines, such as Hybrid Reward Architecture (HRA)~\cite{van2017hybrid}, or MD3QN~\cite{zhang2021distributional}. This is because 1) \cite{zhang2021distributional} shows that their proposed MD3QN and HRA do not outperform MMD-DQN in most of the six Atari games. By contrast, as suggested in Figure~\ref{fig:multi}, our SinkhornDRL has already surpassed MMD-DQN  across almost all the considered games, and thus excels over MD3QN and HRA, correspondingly. 2) The primary focus of our study is the comprehensive advantages of SinkhornDRL over other distributional RL classes, especially in the more common setting within one-dimensional return distributions. The extension capability of SinkhornDRL into the multi-dimensional reward setting is one of its merits, which is not the primary focus of our study.

\newpage
\section*{NeurIPS Paper Checklist}

\begin{enumerate}
	
	\item {\bf Claims}
	\item[] Question: Do the main claims made in the abstract and introduction accurately reflect the paper's contributions and scope?
	\item[] Answer: \answerYes{} 
	\item[] Justification: As the title states, we propose a new family of distributional reinforcement learning algorithm by leveraging Sinkhorn divergence, a regularized Wasserstein loss. The main content of our study accurately states our contributions from the perspectives of algorithm, theory and experiments.
	\item[] Guidelines:
	\begin{itemize}
		\item The answer NA means that the abstract and introduction do not include the claims made in the paper.
		\item The abstract and/or introduction should clearly state the claims made, including the contributions made in the paper and important assumptions and limitations. A No or NA answer to this question will not be perceived well by the reviewers. 
		\item The claims made should match theoretical and experimental results, and reflect how much the results can be expected to generalize to other settings. 
		\item It is fine to include aspirational goals as motivation as long as it is clear that these goals are not attained by the paper. 
	\end{itemize}
	
	\item {\bf Limitations}
	\item[] Question: Does the paper discuss the limitations of the work performed by the authors?
	\item[] Answer: \answerYes{} 
	\item[] Justification: We have explicitly discussed the limitations of our study in a separate paragraph of the last section. 
	\item[] Guidelines:
	\begin{itemize}
		\item The answer NA means that the paper has no limitation while the answer No means that the paper has limitations, but those are not discussed in the paper. 
		\item The authors are encouraged to create a separate "Limitations" section in their paper.
		\item The paper should point out any strong assumptions and how robust the results are to violations of these assumptions (e.g., independence assumptions, noiseless settings, model well-specification, asymptotic approximations only holding locally). The authors should reflect on how these assumptions might be violated in practice and what the implications would be.
		\item The authors should reflect on the scope of the claims made, e.g., if the approach was only tested on a few datasets or with a few runs. In general, empirical results often depend on implicit assumptions, which should be articulated.
		\item The authors should reflect on the factors that influence the performance of the approach. For example, a facial recognition algorithm may perform poorly when image resolution is low or images are taken in low lighting. Or a speech-to-text system might not be used reliably to provide closed captions for online lectures because it fails to handle technical jargon.
		\item The authors should discuss the computational efficiency of the proposed algorithms and how they scale with dataset size.
		\item If applicable, the authors should discuss possible limitations of their approach to address problems of privacy and fairness.
		\item While the authors might fear that complete honesty about limitations might be used by reviewers as grounds for rejection, a worse outcome might be that reviewers discover limitations that aren't acknowledged in the paper. The authors should use their best judgment and recognize that individual actions in favor of transparency play an important role in developing norms that preserve the integrity of the community. Reviewers will be specifically instructed to not penalize honesty concerning limitations.
	\end{itemize}
	
	\item {\bf Theory Assumptions and Proofs}
	\item[] Question: For each theoretical result, does the paper provide the full set of assumptions and a complete (and correct) proof?
	\item[] Answer: \answerYes{} 
	\item[] Justification: We have a few propositions and theorems in our study as our main theoretical contributions, and the complete proof of them are provided in Appendix, respectively. 
	\item[] Guidelines:
	\begin{itemize}
		\item The answer NA means that the paper does not include theoretical results. 
		\item All the theorems, formulas, and proofs in the paper should be numbered and cross-referenced.
		\item All assumptions should be clearly stated or referenced in the statement of any theorems.
		\item The proofs can either appear in the main paper or the supplemental material, but if they appear in the supplemental material, the authors are encouraged to provide a short proof sketch to provide intuition. 
		\item Inversely, any informal proof provided in the core of the paper should be complemented by formal proofs provided in appendix or supplemental material.
		\item Theorems and Lemmas that the proof relies upon should be properly referenced. 
	\end{itemize}
	
	\item {\bf Experimental Result Reproducibility}
	\item[] Question: Does the paper fully disclose all the information needed to reproduce the main experimental results of the paper to the extent that it affects the main claims and/or conclusions of the paper (regardless of whether the code and data are provided or not)?
	\item[] Answer: \answerYes{} 
	\item[] Justification: We have provided the detailed experimental settings in both the experimental part of the main content and the appendix, including the reference code, which our experiments adapt from.
	
	\item[] Guidelines:
	\begin{itemize}
		\item The answer NA means that the paper does not include experiments.
		\item If the paper includes experiments, a No answer to this question will not be perceived well by the reviewers: Making the paper reproducible is important, regardless of whether the code and data are provided or not.
		\item If the contribution is a dataset and/or model, the authors should describe the steps taken to make their results reproducible or verifiable. 
		\item Depending on the contribution, reproducibility can be accomplished in various ways. For example, if the contribution is a novel architecture, describing the architecture fully might suffice, or if the contribution is a specific model and empirical evaluation, it may be necessary to either make it possible for others to replicate the model with the same dataset, or provide access to the model. In general. releasing code and data is often one good way to accomplish this, but reproducibility can also be provided via detailed instructions for how to replicate the results, access to a hosted model (e.g., in the case of a large language model), releasing of a model checkpoint, or other means that are appropriate to the research performed.
		\item While NeurIPS does not require releasing code, the conference does require all submissions to provide some reasonable avenue for reproducibility, which may depend on the nature of the contribution. For example
		\begin{enumerate}
			\item If the contribution is primarily a new algorithm, the paper should make it clear how to reproduce that algorithm.
			\item If the contribution is primarily a new model architecture, the paper should describe the architecture clearly and fully.
			\item If the contribution is a new model (e.g., a large language model), then there should either be a way to access this model for reproducing the results or a way to reproduce the model (e.g., with an open-source dataset or instructions for how to construct the dataset).
			\item We recognize that reproducibility may be tricky in some cases, in which case authors are welcome to describe the particular way they provide for reproducibility. In the case of closed-source models, it may be that access to the model is limited in some way (e.g., to registered users), but it should be possible for other researchers to have some path to reproducing or verifying the results.
		\end{enumerate}
	\end{itemize}

	\item {\bf Open access to data and code}
	\item[] Question: Does the paper provide open access to the data and code, with sufficient instructions to faithfully reproduce the main experimental results, as described in supplemental material?
	\item[] Answer: \answerYes{} 
	\item[] Justification: We have provided the code in the supplementary files for faithful reproducibility. 
	\item[] Guidelines:
	\begin{itemize}
		\item The answer NA means that paper does not include experiments requiring code.
		\item Please see the NeurIPS code and data submission guidelines (\url{https://nips.cc/public/guides/CodeSubmissionPolicy}) for more details.
		\item While we encourage the release of code and data, we understand that this might not be possible, so “No” is an acceptable answer. Papers cannot be rejected simply for not including code, unless this is central to the contribution (e.g., for a new open-source benchmark).
		\item The instructions should contain the exact command and environment needed to run to reproduce the results. See the NeurIPS code and data submission guidelines (\url{https://nips.cc/public/guides/CodeSubmissionPolicy}) for more details.
		\item The authors should provide instructions on data access and preparation, including how to access the raw data, preprocessed data, intermediate data, and generated data, etc.
		\item The authors should provide scripts to reproduce all experimental results for the new proposed method and baselines. If only a subset of experiments are reproducible, they should state which ones are omitted from the script and why.
		\item At submission time, to preserve anonymity, the authors should release anonymized versions (if applicable).
		\item Providing as much information as possible in supplemental material (appended to the paper) is recommended, but including URLs to data and code is permitted.
	\end{itemize}

	\item {\bf Experimental Setting/Details}
	\item[] Question: Does the paper specify all the training and test details (e.g., data splits, hyperparameters, how they were chosen, type of optimizer, etc.) necessary to understand the results?
	\item[] Answer: \answerYes{} 
	\item[] Justification: We provide the links of the reference code, detailed experimental setting in the appendix, and the code in the supplementary files.
	\item[] Guidelines:
	\begin{itemize}
		\item The answer NA means that the paper does not include experiments.
		\item The experimental setting should be presented in the core of the paper to a level of detail that is necessary to appreciate the results and make sense of them.
		\item The full details can be provided either with the code, in appendix, or as supplemental material.
	\end{itemize}
	
	\item {\bf Experiment Statistical Significance}
	\item[] Question: Does the paper report error bars suitably and correctly defined or other appropriate information about the statistical significance of the experiments?
	\item[] Answer: \answerYes{} 
	\item[] Justification: We report learning curves on Atari games with the shading indicating the standard deviation. 
	\item[] Guidelines:
	\begin{itemize}
		\item The answer NA means that the paper does not include experiments.
		\item The authors should answer "Yes" if the results are accompanied by error bars, confidence intervals, or statistical significance tests, at least for the experiments that support the main claims of the paper.
		\item The factors of variability that the error bars are capturing should be clearly stated (for example, train/test split, initialization, random drawing of some parameter, or overall run with given experimental conditions).
		\item The method for calculating the error bars should be explained (closed form formula, call to a library function, bootstrap, etc.)
		\item The assumptions made should be given (e.g., Normally distributed errors).
		\item It should be clear whether the error bar is the standard deviation or the standard error of the mean.
		\item It is OK to report 1-sigma error bars, but one should state it. The authors should preferably report a 2-sigma error bar than state that they have a 96\% CI, if the hypothesis of Normality of errors is not verified.
		\item For asymmetric distributions, the authors should be careful not to show in tables or figures symmetric error bars that would yield results that are out of range (e.g. negative error rates).
		\item If error bars are reported in tables or plots, The authors should explain in the text how they were calculated and reference the corresponding figures or tables in the text.
	\end{itemize}
	
	\item {\bf Experiments Compute Resources}
	\item[] Question: For each experiment, does the paper provide sufficient information on the computer resources (type of compute workers, memory, time of execution) needed to reproduce the experiments?
	\item[] Answer: \answerYes{} 
	\item[] Justification: We run our experiments on multiple NVIDIA 3090 Ti GPUs and compare the computational costs of different algorithms we considered in Section \ref{sec:experiments_sensitivity} and Appendix \ref{appendix:sensitivity_cost}.  
	\item[] Guidelines:
	\begin{itemize}
		\item The answer NA means that the paper does not include experiments.
		\item The paper should indicate the type of compute workers CPU or GPU, internal cluster, or cloud provider, including relevant memory and storage.
		\item The paper should provide the amount of compute required for each of the individual experimental runs as well as estimate the total compute. 
		\item The paper should disclose whether the full research project required more compute than the experiments reported in the paper (e.g., preliminary or failed experiments that didn't make it into the paper). 
	\end{itemize}
	
	\item {\bf Code Of Ethics}
	\item[] Question: Does the research conducted in the paper conform, in every respect, with the NeurIPS Code of Ethics \url{https://neurips.cc/public/EthicsGuidelines}?
	\item[] Answer: \answerYes{} 
	\item[] Justification: We have preserved anonymity and conformed with the NeurIPS Code of Ethics.
	\item[] Guidelines:
	\begin{itemize}
		\item The answer NA means that the authors have not reviewed the NeurIPS Code of Ethics.
		\item If the authors answer No, they should explain the special circumstances that require a deviation from the Code of Ethics.
		\item The authors should make sure to preserve anonymity (e.g., if there is a special consideration due to laws or regulations in their jurisdiction).
	\end{itemize}

	\item {\bf Broader Impacts}
	\item[] Question: Does the paper discuss both potential positive societal impacts and negative societal impacts of the work performed?
	\item[] Answer: \answerNA{} 
	\item[] Justification: As our study is to propose a new family of reinforcement learning algorithm and we test the performance on widely used Atari environments, we do not think there is any societal impact of our work.
	\item[] Guidelines:
	\begin{itemize}
		\item The answer NA means that there is no societal impact of the work performed.
		\item If the authors answer NA or No, they should explain why their work has no societal impact or why the paper does not address societal impact.
		\item Examples of negative societal impacts include potential malicious or unintended uses (e.g., disinformation, generating fake profiles, surveillance), fairness considerations (e.g., deployment of technologies that could make decisions that unfairly impact specific groups), privacy considerations, and security considerations.
		\item The conference expects that many papers will be foundational research and not tied to particular applications, let alone deployments. However, if there is a direct path to any negative applications, the authors should point it out. For example, it is legitimate to point out that an improvement in the quality of generative models could be used to generate deepfakes for disinformation. On the other hand, it is not needed to point out that a generic algorithm for optimizing neural networks could enable people to train models that generate Deepfakes faster.
		\item The authors should consider possible harms that could arise when the technology is being used as intended and functioning correctly, harms that could arise when the technology is being used as intended but gives incorrect results, and harms following from (intentional or unintentional) misuse of the technology.
		\item If there are negative societal impacts, the authors could also discuss possible mitigation strategies (e.g., gated release of models, providing defenses in addition to attacks, mechanisms for monitoring misuse, mechanisms to monitor how a system learns from feedback over time, improving the efficiency and accessibility of ML).
	\end{itemize}
	
	\item {\bf Safeguards}
	\item[] Question: Does the paper describe safeguards that have been put in place for responsible release of data or models that have a high risk for misuse (e.g., pretrained language models, image generators, or scraped datasets)?
	\item[] Answer: \answerNA{} 
	\item[] Justification: We believe our paper poses no such risks.
	\item[] Guidelines:
	\begin{itemize}
		\item The answer NA means that the paper poses no such risks.
		\item Released models that have a high risk for misuse or dual-use should be released with necessary safeguards to allow for controlled use of the model, for example by requiring that users adhere to usage guidelines or restrictions to access the model or implementing safety filters. 
		\item Datasets that have been scraped from the Internet could pose safety risks. The authors should describe how they avoided releasing unsafe images.
		\item We recognize that providing effective safeguards is challenging, and many papers do not require this, but we encourage authors to take this into account and make a best faith effort.
	\end{itemize}
	
	\item {\bf Licenses for existing assets}
	\item[] Question: Are the creators or original owners of assets (e.g., code, data, models), used in the paper, properly credited and are the license and terms of use explicitly mentioned and properly respected?
	\item[] Answer: \answerNA{} 
	\item[] Justification: We believe our paper does not use existing assets.
	\item[] Guidelines:
	\begin{itemize}
		\item The answer NA means that the paper does not use existing assets.
		\item The authors should cite the original paper that produced the code package or dataset.
		\item The authors should state which version of the asset is used and, if possible, include a URL.
		\item The name of the license (e.g., CC-BY 4.0) should be included for each asset.
		\item For scraped data from a particular source (e.g., website), the copyright and terms of service of that source should be provided.
		\item If assets are released, the license, copyright information, and terms of use in the package should be provided. For popular datasets, \url{paperswithcode.com/datasets} has curated licenses for some datasets. Their licensing guide can help determine the license of a dataset.
		\item For existing datasets that are re-packaged, both the original license and the license of the derived asset (if it has changed) should be provided.
		\item If this information is not available online, the authors are encouraged to reach out to the asset's creators.
	\end{itemize}
	
	\item {\bf New Assets}
	\item[] Question: Are new assets introduced in the paper well documented and is the documentation provided alongside the assets?
	\item[] Answer: \answerNA{} 
	\item[] Justification:We believe our paper does not release new assets.
	\item[] Guidelines:
	\begin{itemize}
		\item The answer NA means that the paper does not release new assets.
		\item Researchers should communicate the details of the dataset/code/model as part of their submissions via structured templates. This includes details about training, license, limitations, etc. 
		\item The paper should discuss whether and how consent was obtained from people whose asset is used.
		\item At submission time, remember to anonymize your assets (if applicable). You can either create an anonymized URL or include an anonymized zip file.
	\end{itemize}
	
	\item {\bf Crowdsourcing and Research with Human Subjects}
	\item[] Question: For crowdsourcing experiments and research with human subjects, does the paper include the full text of instructions given to participants and screenshots, if applicable, as well as details about compensation (if any)? 
	\item[] Answer: \answerNA{} 
	\item[] Justification: The paper does not involve crowdsourcing nor research with human subjects.
	\item[] Guidelines:
	\begin{itemize}
		\item The answer NA means that the paper does not involve crowdsourcing nor research with human subjects.
		\item Including this information in the supplemental material is fine, but if the main contribution of the paper involves human subjects, then as much detail as possible should be included in the main paper. 
		\item According to the NeurIPS Code of Ethics, workers involved in data collection, curation, or other labor should be paid at least the minimum wage in the country of the data collector. 
	\end{itemize}
	
	\item {\bf Institutional Review Board (IRB) Approvals or Equivalent for Research with Human Subjects}
	\item[] Question: Does the paper describe potential risks incurred by study participants, whether such risks were disclosed to the subjects, and whether Institutional Review Board (IRB) approvals (or an equivalent approval/review based on the requirements of your country or institution) were obtained?
	\item[] Answer: \answerNA{} 
	\item[] Justification: The paper does not involve crowdsourcing nor research with human subjects.
	\item[] Guidelines:
	\begin{itemize}
		\item The answer NA means that the paper does not involve crowdsourcing nor research with human subjects.
		\item Depending on the country in which research is conducted, IRB approval (or equivalent) may be required for any human subjects research. If you obtained IRB approval, you should clearly state this in the paper. 
		\item We recognize that the procedures for this may vary significantly between institutions and locations, and we expect authors to adhere to the NeurIPS Code of Ethics and the guidelines for their institution. 
		\item For initial submissions, do not include any information that would break anonymity (if applicable), such as the institution conducting the review.
	\end{itemize}
	
\end{enumerate}

\end{document}